\documentclass[10pt,twocolumn,letterpaper]{article}

\usepackage{wacv}              

\usepackage{amsmath}
\usepackage{amssymb}
\usepackage{booktabs}
\usepackage{array}
\usepackage{multirow}
\usepackage{color}
\usepackage{colortbl}
\usepackage{framed}
\usepackage{bm}
\usepackage{bbm}
\usepackage{xspace}
\usepackage{enumitem}
\usepackage{cite}
\usepackage{xparse}
\usepackage{xcolor}
\usepackage{algorithm}
\usepackage{lipsum}
\usepackage{listings}
\usepackage{url}
\usepackage{algorithmic}
\usepackage{pifont}
\usepackage{bbding}
\usepackage{mathtools}

\definecolor{Gray}{gray}{0.93}
\definecolor{LightCyan}{rgb}{0.88,0.95,1}
\definecolor{blond}{rgb}{0.98, 0.94, 0.75}

\def \ie {\emph{i.e.}}
\def \eg {\emph{e.g.}}
\def \etal {\emph{et al.}}

\newcommand{\tit}[1]{\smallbreak\noindent\textbf{#1.}}
\newcommand{\tinytit}[1]{\noindent\textbf{#1.}}

\newcommand{\ours}{\texttt{\textbf{TPP-Gaze}}\xspace}
\newcommand{\tppgaze}{\texttt{TPP-Gaze}\xspace}

\DeclareMathOperator*{\argmax}{arg\,max}

\usepackage[pagebackref,breaklinks,colorlinks]{hyperref}

\usepackage[capitalize]{cleveref}
\crefname{section}{Sec.}{Secs.}
\Crefname{section}{Section}{Sections}
\Crefname{table}{Table}{Tables}
\crefname{table}{Tab.}{Tabs.}

\def\wacvPaperID{955} 
\def\confName{WACV}
\def\confYear{2025}

\begin{document}
\title{TPP-Gaze: Modelling Gaze Dynamics\\in Space and Time with Neural Temporal Point Processes}

\author{$^1$Alessandro D'Amelio, $^2$Giuseppe Cartella, $^2$Vittorio Cuculo,\\ $^1$Manuele Lucchi, $^2$Marcella Cornia, $^2$Rita Cucchiara, $^1$Giuseppe Boccignone
\\
$^1$University of Milan, Italy \quad $^2$University of Modena and Reggio Emilia, Italy\\
{\tt\small $^1$\{name.surname\}@unimi.it \quad $^2$\{name.surname\}@unimore.it}
}

\maketitle

\begin{abstract}
Attention guides our gaze to fixate the proper location of the scene and holds it in that location for the deserved amount of time given current processing demands, before shifting to the next one. As such, gaze deployment crucially is a temporal process. Existing computational models have made significant strides in predicting spatial aspects of observer's visual scanpaths (\emph{where} to look), while often putting on the background the temporal facet of attention dynamics (\emph{when}).
In this paper we present \tppgaze, a novel and principled approach to model scanpath dynamics based on Neural Temporal Point Process (TPP), that jointly learns the temporal dynamics of fixations position and duration, integrating deep learning methodologies with point process theory. We conduct extensive experiments across five publicly available datasets. Our results show the overall superior performance of the proposed model compared to state-of-the-art approaches. Source code and trained models are publicly available at: \small{\texttt{\href{https://github.com/phuselab/tppgaze}{https://github.com/phuselab/tppgaze}}}.
\end{abstract}

\section{Introduction}
\label{sec:intro}
Gaze, the act of directing the eyes toward a location in the visual world, is considered a good measure of overt attention and,  more generally,  a window to the observer's thoughts, intentions, and emotions.  It is no surprise that research spanning decades has struggled to produce several computational models aiming at effectively predicting attention towards regions or events within the landscape of visual and multimodal stimuli. With roots in psychology and neuroscience, these approaches have gained traction in the computer vision and pattern recognition fields since the seminal Itti~\etal~\cite{IttiKoch98} model; more recently,
state-of-the-art approaches rely on machine learning advancements,  typically employing deep neural architectures to the purpose (but see~\cite{kummerer2021state} or \cite{cartella2024trends} for an in-depth review). As a matter of fact, the vast majority of works in the field has focused on the computational modelling of spatial saliency in the shape of saliency maps,  namely, a topographic map representing  the likelihood of fixating a given location of the scrutinised stimulus, a fixation being defined as the period of time during which a part of the visual stimulus (the patch) on the screen is gazed at. 
Nevertheless, a growing number of models (\ie, scanpath models) are addressing the prediction of a sequence of fixations -- namely, the scanpath -- where the gaze shift from one fixation to the next represents a saccade. Beyond the salience representation, these models explicitly unfold the dynamics of overt attention allocation over a stimulus~\cite{kummerer2023predicting}.
\begin{figure}[t]
    \centering
    \includegraphics[width=\linewidth]{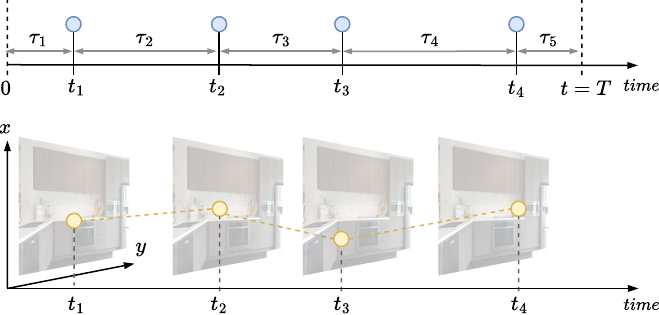}
    \vspace{-0.4cm}
    \caption{Scanpath dynamics as a marked TPP. Time is represented on the horizontal axis, and different scanpath fixations occurs at time $t_1, t_2, t_3$ and $t_4$.}
    \label{fig:ntpp}
    \vspace{-0.3cm}
\end{figure}
It is worth remarking, though, that barely predicting the spatial sequence of fixations, does not entail proper modelling of the temporal evolution of attention. By and large, most scanpath models predict an ordered sequence of events while neglecting their continuous timestamp information. As a result, these models are able to tell \textit{where} to look and in what order, but fail in answering \textit{when}. In many respects, this is not an innocent flaw: human actions often rely on visual information, therefore it is important to direct attention to the right place at the right time~\cite{tatler2017latest}. Practically, modelling \textit{when} to perform a saccade translates to devising scanpath models able to predict the sequence of both fixations position and corresponding duration. Albeit recently few approaches have successfully dealt with such problem~\cite{sun2019visual, chen2018scanpath,chen2021predicting,mondal2023gazeformer,chen2024beyond} via fully engineered approaches, only a marginal subset of them has tackled it in a mathematically principled way~\cite{tatler2017latest,boccignone2020gaze,d2021gazing}. This has typically resulted in a weaker generality of the methods which are tailored to specific contexts or applications. Under such circumstances,  the chief concern of the present work is to introduce a fresh, general and simple view on the problem of scanpath modelling: in brief, we consider a scanpath as the realisation of a point process in space and time, precisely that of a Neural Temporal Point Process. 

Temporal Point Processes (TPPs) are probabilistic generative models designed for continuous-time event sequences. Neural TPPs~\cite{mei2017neural,shchur2019intensity,zuo2020transformer, zhang2020self} integrate key concepts from point process literature with deep learning methodologies, facilitating the creation of adaptable and effective models.
Notably, the modelling assumptions of Neural TPPs align perfectly with the structure of scanpath data. A scanpath consists of a series of events (saccades) occurring at irregular intervals (fixation durations), which is \emph{exactly} what Neural TPPs are designed to model. While the psychological and neuroscience literature has used traditional point processes for eye movement analysis \cite{barthelme2013modeling,engbert2015spatial,ylitalo2017statistical}, these tools are not well-suited for scanpath prediction due to their inability to handle stimuli. In other words, traditional TPPs are effective for studying the observer but fall short when addressing Computer Vision tasks related to attention allocation prediction. In contrast, Neural TPP-based models offer the best of both worlds: they combine the robust theoretical framework of TPPs with the flexibility and power of modern neural networks. Nevertheless, this is the first attempt to adopt them for the scanpath modelling problem.

Our key contributions can be summarised as follows: 1) We propose a novel scanpath model able to jointly learn the temporal dynamics of both fixations position and duration. 2) We extend recent Neural TPP models to deal with visual data (\ie, images) and connect scanpath modelling and prediction to point process theory. To assess our proposal, which can be appreciated at a glance in~\cref{fig:ntpp}, we conduct experiments on five publicly available datasets, showing an overall superior performance of the proposed model when compared to state-of-the-art approaches.

\section{Background and Related Work}
\label{sec:related}
\subsection{Neural Temporal Point Processes (TPPs)}

Consider a sequence of generic events happening irregularly over time, TPPs model the next arrival time of an event by conditioning on the past events. Specifically, denote $\mathcal{H}_{t} = \{t_n \in \mathcal{T} : t_n < t\}$ (with $\mathcal{T}$ representing the sequence of strictly increasing arrival times of events) the history of arrival times of all events up to time $t$, the relation between the current arrival time $t$ and the history, is typically determined by the conditional intensity function $\lambda^*(t) = \lambda(t|\mathcal{H}_{t})$, whose functional form determines the properties of the TPP. Equivalently, the sequence of strictly positive inter-event times $\tau_n = t_n - t_{n-1}$ can be considered. Knowing the conditional intensity function allows to recover the conditional probability of the inter-arrival time of an event:
\begin{equation}
    \begin{split}
        p^*(\tau_n) &= p(\tau_n|\mathcal{H}_{t_n}) \\
        &=\lambda^*(t_{n-1}+\tau_n)\exp\left(-\int_{0}^{\tau_n}\lambda^*(t_{n-1} + s)ds\right).
    \end{split}
\label{eq:cond_prob}
\end{equation}

For instance, under the the assumptions of no dependence on the history and constancy over time (\ie, $\lambda^*(t) = k$, with $k\geq0$), the homogeneous Poisson process is recovered, with inter-event times distributed according to the exponential distribution. Choosing more complex functional forms for $\lambda^*(t)$ allows to recover many well known TPPs such as Hawkes or self-correcting processes~\cite{hawkes1971spectra,isham1979self}. Clearly, restricting $\lambda^*(t)$ to a specific parameterisation limits the general applicability of TPPs.  For this reason, most recent solutions resorted to neural approaches (Neural TPPs) implementing learnable parametric forms of the intensity function,  $\lambda_{\theta}^*(t)$~\cite{du2016recurrent,huang2019recurrent}.
As an example, early Neural TPPs, such as the Neural Hawkes Process~\cite{mei2017neural}, used RNNs to model the intensity function of the process. More recently, self-attention mechanisms have been employed to the same purpose~\cite{zuo2020transformer, zhang2020self}. The choice of the parametric form for the intensity function has to take into account the necessity of a closed form solution of the integral in \cref{eq:cond_prob}, thus practically restricting the expressiveness of the model. More complex parametric forms would require Monte Carlo approximation of the integral~\cite{mei2017neural}. To overcome such limitations, Shchur~\etal~\cite{shchur2019intensity} recently proposed to directly learn the parametric conditional distribution $p_{\theta}^*(\tau)$ of the inter-arrival times rather than the conditional intensity function $\lambda_{\theta}^*(t)$, thus recasting learning Neural TPPs as a density estimation problem. 

\tit{Marked TPPs} The basic mathematical formalism of TPPs allows to naturally handle the dynamics of arrival times of events. However, the distribution of time until the next event might depend on factors other than the history. Event data is often accompanied with some kind of covariate indicating the nature of the specific event being predicted. In the realm of TPPs, such covariate are called \textit{marks}. More formally, a marked TPP is a random process whose realisations consists of a sequence of discrete events localised in time, $\{\mathbf{r}_{F_n}, t_n\}$, with the timing $t_n \in \mathbb{R}^+$ and the mark $\mathbf{r}_{F_n} \in \mathcal{M}$. The mark $\mathbf{r}_{F_n}$ is typically modelled as an integer representing the type of event, however other kinds of marks (\eg, $\mathcal{M} = \mathbb{R}^2$) can be eventually adopted. Specifying a marked-TPP involves the definition of the joint conditional density function of the next event, with inter-event time $\tau_n$ and mark $\mathbf{r}_{F_n}$, given the history of past events: $p^*(\mathbf{r}_{F_n},\tau_n) = p(\mathbf{r}_{F_n},\tau_n | \mathcal{H}_{t_n})$. By assuming a conditional distribution parameterised by the weights of a neural model, $p_{\theta}^*(\mathbf{r}_{F_n},\tau_n)$, inference can be performed by maximising the joint likelihood of the $N$ observed events in a sequence:
\begin{equation}
\theta^*=\argmax_{\theta}\prod_{n=0}^Np_{\theta}(\mathbf{r}_{F_n},\tau_n|\mathcal{H}_t)=\prod_{n=0}^Np_{\theta}^*(\mathbf{r}_{F_n},\tau_n).
\end{equation}

Applications of Neural TPPs span a variety of fields of research~\cite{shchur2021neural}, such as healthcare~\cite{enguehard2020neural}, finance~\cite{bacry2015hawkes}, social network analysis~\cite{trivedi2019dyrep}, earthquake forecasting~\cite{bray2013assessment}, and recommender systems~\cite{kumar2019predicting}. Here we adopt the Neural TPP framework to model the dynamics of attention allocation on visual data.

\subsection{Scanpath Modelling}

Modelling scanpaths involves defining a mapping from visual data, $\mathbf{I}$ (raw image data representing either a static picture or a stream of images), to a sequence of time-stamped gaze locations $\mathcal{S} = \{(\mathbf{r}_{F_1}, t_1), (\mathbf{r}_{F_2}, t_2), \dots(\mathbf{r}_{F_N}, t_N)\}$. Here $\mathbf{r}_{F_n} \in \mathbb{R}^2$ represents the two-dimensional vector of spatial coordinates of the $n-$th fixation on the stimulus $\mathbf{I}$, while $t_n \in \mathbb{R}^+$ represents its arrival time. Eventually, a perceptual representation of the input stimuli, $\mathcal{Z}$, is computed, with the aim of locating the relevant objects inside the scene: 
\begin{equation}
	\mathbf{I} \rightarrow \mathcal{Z} \rightarrow \{(\mathbf{r}_{F_1}, t_1), (\mathbf{r}_{F_2}, t_2), \dots(\mathbf{r}_{F_N}, t_N)\}.
	\label{eq:mapping}
\end{equation}

Here we assume that no specific external task or goal is given to the observer (\ie, free-viewing condition). Notably, the dynamics of the attentive process, which unrolls as a sequence of fixations location with corresponding duration/arrival time, is characterised by an inherent randomness which likely stems from internal stochastic fluctuations affecting sensory and information processing, movement planning, and execution~\cite{vanBeers2007sources}, in both fixations location and corresponding duration.
Notably, many scanpath models proposed in the recent literature~\cite{assens2017saltinet,Assens2018pathgan, kummerer2022deepgaze,sui2023scandmm} get rid of fixations' timestamp information by rearranging the sequence $ \{(\mathbf{r}_{F_1}, t_1), (\mathbf{r}_{F_2}, t_2), \dots\}$ as $\left\{\mathbf{r}_{F}(1), \mathbf{r}_{F}(2), \cdots\right\}$, thus assuming  $\left(\mathbf{r}_{F_{n}}, t_{n}\right)=\mathbf{r}_{F}(n)$. 

A handful of solutions~\cite{tatler2017latest, boccignone2020gaze, d2021gazing} have dealt with this problem in its entirety by starting from specific theoretical frameworks. In~\cite{tatler2017latest} Tatler~\etal~modelled saccades timings as an evidence accumulation process with clear neurobiological significance. In a similar vein, in~\cite{d2021gazing} a Langevin-type SDE race model~\cite{bogacz2006physics} was adopted to predict fixations and their duration in socially relevant contexts, while in~\cite{boccignone2020gaze} fixation duration was equated to the patch residence time of a forager searching for nourishment. Conversely, the vast majority of recent methods~\cite{sun2019visual, chen2018scanpath,chen2021predicting,mondal2023gazeformer,chen2024beyond} simply model fixation duration by employing specific neural architectural choices that aim at associating each fixation to its corresponding duration.  

In a different vein, this work recasts the whole visual attention allocation process in the mathematical framework of point process theory~\cite{daley2007introduction}. This emphasises the central role of visual attention's spatio-temporal dynamics by explicitly modelling scanpaths as sequences of discrete events happening at irregular intervals. Specifically, we conceive a scanpath as a realisation of a random process whose events happen at strictly increasing arrival times $\mathcal{T} = \{t_1,\dots,t_N\}$. Fixations duration can be recovered by resorting to inter-event times $\tau_n = t_n - t_{n-1}$, while their locations can be represented as the two-dimensional continuous mark associated to the $n$-th event. Under this assumption, (Neural) Temporal Point Processes (TPPs) represent the natural choice for modelling this kind of data.

\section{Proposed Method}
\label{sec:method}
Given a stimulus (image) $\mathbf{I}_j$, an ensemble of $N_{obs}$ observers performs a sequence of fixations and saccades (scanpath) on it, thus obtaining a set of sequences $\mathcal{C}_j = \{S^1,\dots,S^{N_{obs}}\}$. Each scanpath $S^i$ is a sequence of pairs (events) $S^i_n = (\mathbf{r}_{F_n}, t_n)$ each composed by a fixation position (marker) $\mathbf{r}_{F_n} \in \mathbb{R}^2$, and a corresponding arrival time $t_n \in \mathbb{R}^+$. At the most general level, we are interested in modelling the stochastic generative process that given a semantic representation of the image $\mathcal{Z}_j$ and the history of past events $\mathcal{H}_t$, simulates the next fixation position and duration. More formally:
\begin{equation}
    S^i_{n+1} \sim p_{\theta}(\mathbf{r}_{F_{n+1}}, t_{n+1}|\mathcal{H}_t,\mathcal{Z}_j),
    \label{eq:model_Sin}
\end{equation}
\noindent where $p_{\theta}(\cdot)$ represents the parametric joint conditional distribution of a Neural TPP~\cite{shchur2019intensity}.

\begin{figure}[t]
    \centering
    \includegraphics[width=.99\linewidth]{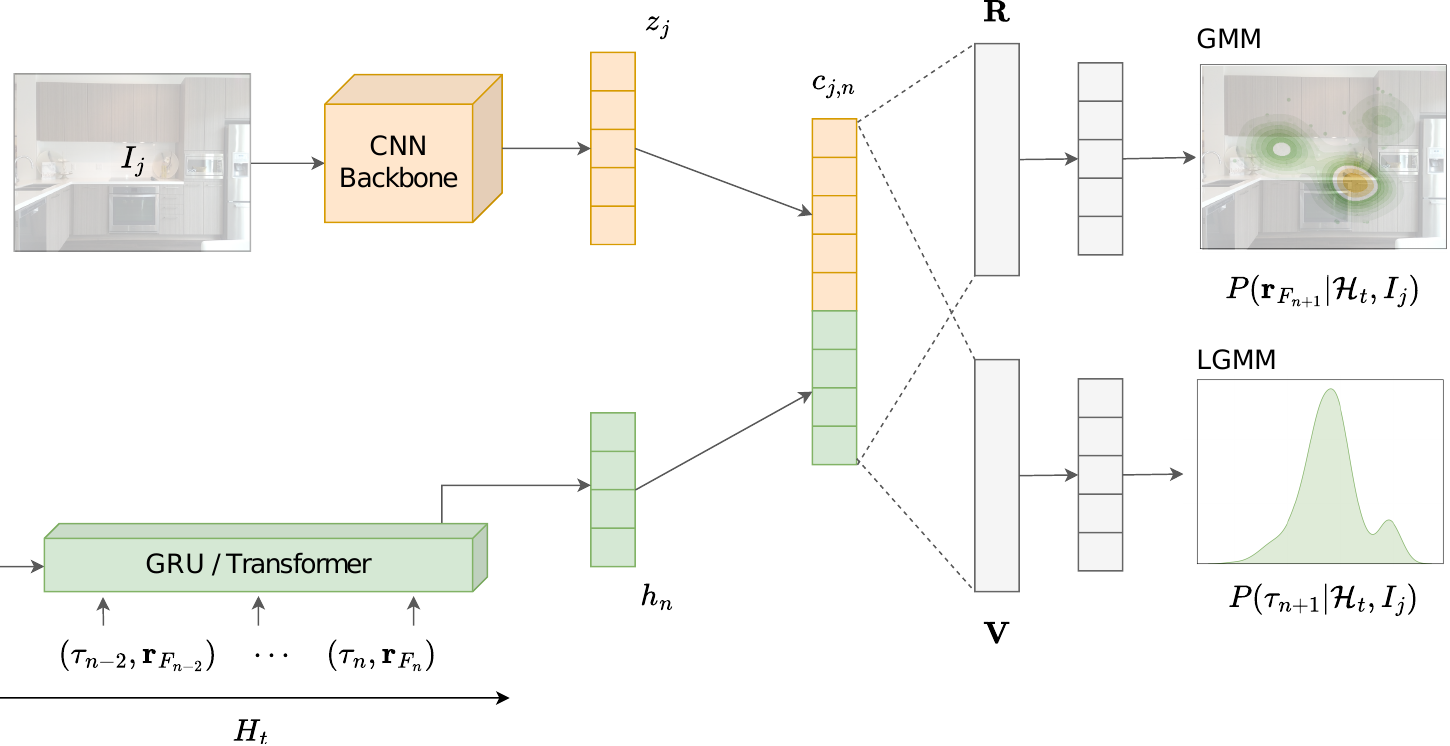}
    \vspace{-0.15cm}
    \caption{Overview of \tppgaze~model architecture. Given a semantic representation of the image ($z_j$) and the history of past events ($h_n$), the next fixation position and duration are simulated.}
    \label{fig:architecture}
    \vspace{-0.3cm}
\end{figure}

\subsection{Architecture}
\label{sec:architecture}
In the following, we present the architecture of \tppgaze, implementing a scanpath model on an image as a Neural TPP. 

\tit{Representing Scene Semantics} As outlined in \cref{eq:mapping}, the sequence of events composing a scanpath depends not only on the history of past events, but on a perceptual representation of the input stimulus $\mathbf{I}_j$, encoding scene semantics and relevant objects location. We extract the perceptual representation of the input image via a CNN architecture inspired by~\cite{kummerer2022deepgaze}. Specifically, the input image is first processed by a pre-trained DenseNet201 CNN~\cite{huang2017densely}. 
Activation maps from various convolutional layers (as reported in~\cite{kummerer2022deepgaze}) are extracted, thus obtaining a $2,048$ channels volume, each representing the location of semantic features inside the scene. It is worth noticing that learning to predict fixations location (\ie, marks) involves a mapping between coordinates in Cartesian space, a task in which standard convolutions have been reported to fail~\cite{liu2018intriguing}. In the vein of~\cite{martin2022scangan360,sui2023scandmm}, we adopt a CoordConv layer~\cite{liu2018intriguing} to give convolutions access to their own input coordinates. This results in a $2,051$ channels volume which is fed as input to $3$ layers of $1 \times 1$ convolutions, followed by a linear layer mapping to  $\mathbf{z}_j$ acting as our semantic representation.

\tit{Representing History} Neural TPPs employ either Recurrent Neural Networks (RNNs) and their variants (\eg, LSTM, GRU)~\cite{du2016recurrent,xiao2017modeling,shchur2019intensity} or Transformer encoders~\cite{zuo2020transformer,zhang2020self} to model the nonlinear dependency over both the markers and the timings from past events~\cite{shchur2021neural}. As shown in \cref{fig:architecture}, the pair $(\mathbf{r}_{F_n},\tau_n)$ representing the event occurring at the time $t_n$ with fixation position $\mathbf{r}_{F_n}$ and duration $\tau_n = t_n-t_{n-1}$, is fed as the input into either a GRU or a Transformer encoder as described in~\cite{zuo2020transformer}.
The Transformer/GRU state embedding $\mathbf{h}_{n}$ represents the influence of the history up to the $n-$th fixation. Hence, can be employed as a vector space representation of $\mathcal{H}_{t_n}$. Taking into account the semantic representation $\mathbf{z}_j$ and the history embedding $\mathbf{h}_{n}$, \cref{eq:model_Sin} can be rewritten as:
\begin{equation}
    S^i_{n+1} \sim p_{\theta}(\mathbf{r}_{F_{n+1}}, t_{n+1}|\mathbf{h}_{n},\mathbf{z}_j).
    \label{eq:model_Sin_vector}
\end{equation}

\tit{Fixation Duration Generation}
We model the conditional dependence of the distribution $p_{\theta}(\tau_{n+1}|\mathbf{h}_{n},\mathbf{z}_j)$ on both past events and stimulus by concatenating the history embedding and semantic vectors into a context vector $\mathbf{c}_{j,n} = [\mathbf{h}_{n} || \mathbf{z}_j]$. In the vein of~\cite{shchur2019intensity}, the latter is employed to learn the parameters of a Log-Gaussian Mixture Model (LGMM) via an affine transform:
\begin{equation}
\begin{split}
    \mathbf{w}=softmax(\mathbf{V}_{\mathbf{w}}&\mathbf{c}_{j,n})
    \quad  
    \mathbf{s}=\exp(\mathbf{V}_{\mathbf{s}}\mathbf{c}_{j,n}) \\ 
    &\boldsymbol{m}=\mathbf{V}_{\mathbf{m}}\mathbf{c}_{j,n}
\end{split}
\end{equation}
where $\boldsymbol{w} \in \mathbb{R}_+^K$ are the mixture weights, $\boldsymbol{m} \in \mathbb{R}^K$ are the mixture means, and $\boldsymbol{s} \in \mathbb{R}_+^K$ are the standard deviations. $K$ represents the number of mixture components. The fixation duration for the $n-$th event can be generated by sampling from the LGMM defined by:
\begin{equation}
    \begin{split}
        p_{\theta}^*(\tau_n|\boldsymbol{c}_{j,n}) &= p(\tau_n|\mathbf{w},\boldsymbol{m},\mathbf{s})\\
        &=\sum_{k=1}^Kw_k\frac1{\tau_n s_k\sqrt{2\pi}}\exp\left(-\frac{(\log\tau_n-m_k)^2}{2s_k^2}\right).
\end{split}
\label{eq:lik_dur}
\end{equation}

\tit{Fixation Position (Mark) Generation}
Similarly, given the context vector $\mathbf{c}_{j,n}$, we define the conditional probability of the next mark (fixation position), $p_{\theta}(\mathbf{r}_{F_{n+1}}|\mathbf{h}_{n},\mathbf{z}_j)$, as a 2D Gaussian Mixture Model (GMM) whose parameters are obtained via another affine projection:
\begin{equation}
\begin{split}
\boldsymbol{\omega}_g=softmax(\mathbf{R}^g_{\boldsymbol{\omega}}&\mathbf{c}_{j,n})\ 
    \quad \boldsymbol{\Sigma}_g=diag(\exp(\mathbf{R}^g_{\boldsymbol{\Sigma}}\mathbf{c}_{j,n})) \\ &\boldsymbol{\mu}_g=\mathbf{R}^g_{\boldsymbol{\mu}}\mathbf{c}_{j,n}
\end{split}
\end{equation}
where $\boldsymbol{\omega}_g \in \mathbb{R}_+^2$ are the mixture weights, $\boldsymbol{\mu}_g \in \mathbb{R}^2$ are the mixture means, and $\boldsymbol{\Sigma}_g \in \mathbb{R}^{2\times2}$ are the diagonal covariance matrices of $G$ bi-variate Gaussian distributions. The $x$ and $y$ coordinates of the $n-$th fixation can be generated by sampling from the GMM defined by:
\begin{equation}
    \begin{split}
        p_{\theta}^*(\mathbf{r}_{F_{n}}|&\boldsymbol{c}_{j,n}) = p(\mathbf{r}_{F_{n}}|\boldsymbol{\omega},\boldsymbol{\mu},\boldsymbol{\Sigma}) \\
        &= \sum_{g=1}^{G} \boldsymbol{\omega}_g \frac{\exp\left(-\frac12(\mathbf{r}_{F_{n}}-\boldsymbol{\mu}_g)^\mathrm{T}\boldsymbol{\Sigma}^{-1}\left(\mathbf{r}_{F_{n}}-\boldsymbol{\mu}_g\right)\right)}{\sqrt{(2\pi)^2|\boldsymbol{\Sigma}_g|}}.
    \label{eq:lik_fix}
    \end{split}
\end{equation}

\subsection{Model Inference} 

Consider a set of stimuli $\mathcal{I} = \{\mathbf{I}_1,\dots,\mathbf{I}_j,\dots,\mathbf{I}_J\}$ each gazed by $N_{obs}$ human observers. Each observer produces an ensemble of scanpaths $\mathcal{C}_j = \{S^1,\dots,S^{N_{obs}}\}$ with $S^i_n = (\mathbf{r}_{F_n}^i, \tau_n^i)$ representing an event (\ie, fixation position and duration). Model inference is performed by minimising a negative log-likelihood loss with respect to the parameters of the semantic network, the GRU/Transformer encoding history of events, and the affine transforms of the LGMM and GMM. Formally, the loss function is defined as follows: 
\begin{equation}
    \mathcal{L}(\boldsymbol{\theta})=-\sum_{j}\sum_{i}\sum_{n}\left[\log p_{\theta}^*(\tau_n^i|\boldsymbol{c}_{j,n})+\log p_{\theta}^*(\mathbf{r}_{F_{n}}^i|\boldsymbol{c}_{j,n})\right].
\end{equation}

\section{Experiments}
\label{sec:experiments}
\subsection{Experimental Setup} 
\label{sec:exper_setup}

\tinytit{Datasets} Regarding the stimuli and eye tracking data, we select five publicly available datasets of human recorded scanpaths comprising both fixation positions and durations: COCO-FreeView, MIT1003, OSIE, NUSEF, and FiFa.

COCO-FreeView~\cite{yang2023predicting} is a high-quality dataset capturing free viewing behaviour, featuring the same natural images adopted in COCO-Search18, annotated with $822,602$ eye fixations from a free-viewing task. Only train and validation splits are publicly released. Each image was presented for $5$ seconds.
The MIT1003 dataset~\cite{judd2009learning} comprises $1,003$ images primarily featuring natural scenes. It provides eye movement data from $15$ subjects, observing stimuli for 3 seconds. 
The OSIE dataset~\cite{xu2014predicting} comprises 700 images with eye-tracking data of 15 viewers. The dataset was explicitly devised to incorporate high-level semantic attributes.
The NUSEF (NUS Eye Fixation) dataset~\cite{ramanathan2010eye} features a diverse collection of images, representing a range of semantic concepts and capturing objects with varying scale, illumination, and orientation.
Each free-view experiment lasted $5$ seconds.
The Fixations In Faces (FiFa) database~\cite{cerf2008predicting} shares data related to observers' viewing of faces in natural settings.Each image was presented for $2$ seconds.

\tit{Implementation and Training Details} COCO-FreeView and MIT1003 datasets are used for model training. To this end, $70\%$ of the images from both datasets are used for training, while the remaining $30\%$ is equally partitioned between validation and test sets. We use AdamW as optimizer, with weight decay set to $10^{-1}$, and the learning rate set to $10^{-3}$. Batch size is equal to $128$. We employ early stopping after $20$ epochs with no improvement on the validation set.
Following previous literature~\cite{kummerer2022deepgaze,chen2018scanpath}, during training and evaluation, we discard the first fixation and removed all scanpaths containing less than four fixations.

\tit{Scanpath Evaluation Metrics}  A variety of scanpath evaluation metrics have been proposed to quantitatively assess the similarity between real and simulated eye-movements~\cite{anderson2015comparison,kummerer2021state}. Here we employ the MultiMatch, ScanMatch, and Sequence Score evaluation metrics since they explicitly consider fixation duration in the evaluation process. Moreover the String Edit Distance is adopted to further evaluate predicted scanpaths.

MultiMatch (MM)~\cite{jarodzka2010vector, dewhurst2012depends} assesses scanpaths based on five features: shape (Sh), length (Len), direction (Dir), position (Pos), and duration (Dur). Scanpaths are temporally aligned and compared using the Dijkstra algorithm. Similarity is determined by applying vector arithmetic to the aligned saccade pairs.
ScanMatch (SM)~\cite{cristino2010scanmatch} encodes scanpaths as letter sequences by segmenting them into spatial and temporal bins.
In our experiments, the longest dimension of the stimuli is divided into $14$ bins, while the shortest is split into $8$ bins. The temporal bin size is set to $50$ ms for scanpath models delivering fixation duration estimates. The encoded scanpaths are then aligned and compared, with higher scores reflecting greater spatial, temporal, and sequential similarity.
Sequence Score (SS)~\cite{Yang_2020_CVPR} transforms the human and predicted scanpaths into sequences of fixation cluster IDs and compares them using a string-matching algorithm.
String Edit Distance (SED)~\cite{brandt1997spontaneous}, 
first partitions the input stimulus into an $n\times n$ grid. Scanpaths are then transformed into strings and the string-edit algorithm calculates the distance between them.

\begin{table}[t]
  \centering
  \small
  \setlength{\tabcolsep}{.22em}
  \resizebox{0.98\linewidth}{!}{
    \begin{tabular}{lcc cc c cc c >{}c>{}c c >{}c>{}c c >{}c}
    \toprule
    & & & \multicolumn{2}{c}{\textbf{Dim}} & & \multicolumn{2}{c}{\textbf{GMM}} & & \multicolumn{2}{c}{MM (KL-Div) $\downarrow$} & & \multicolumn{2}{c}{SM (KL-Div) $\downarrow$} & & \multicolumn{1}{c}{SED $\downarrow$} \\
    \cmidrule{4-5} \cmidrule{7-8} \cmidrule{10-11} \cmidrule{13-14} \cmidrule{16-16}
    \hspace{0.4cm} & \textbf{CNN} & & Img & TPP & & $K$ & $G$ & & Dur & Avg & & w/ Dur & w/o Dur & & Avg \\
    \midrule
    \rowcolor{Gray}
    \multicolumn{16}{l}{\textit{Image Backbone}} \\
     & RN & & 256 & 256 & & 4 & 16 & & \textbf{0.011}	& 0.037 & & 0.113 & 0.101 & & 17.575 \\
    & DN & & 256 & 256 & & 4 & 16 & & 0.012	& \textbf{0.028} & & \textbf{0.078} & \textbf{0.060} & & \textbf{17.032} \\
    \midrule
    \rowcolor{Gray}
    \multicolumn{16}{l}{\textit{Image and TPP Dimensionalities}} \\
    & DN & & 128 & 128 & & 4 & 16 & & 0.010 & 0.031 & & 0.094 & 0.069 & & 16.959 \\
    & DN & & 128 & 256 & & 4 & 16 & & 0.012 & 0.030 & & 0.084 & 0.063 & & \textbf{16.887} \\
    & DN & & 256 & 128 & & 4 & 16 & & 0.009 & 0.037 & & 0.105 & 0.082 & & 17.413 \\
    & DN & & 256 & 256 & & 4 & 16 & & 0.012	& 0.028 & & \textbf{0.078} & \textbf{0.060} & & 17.032 \\
    & DN & & 256 & 512 & & 4 & 16 & & 0.010 & 0.031 & & 0.101 & 0.095 & & 17.462 \\
    & DN & & 512 & 256 & & 4 & 16 & & \textbf{0.008} & 0.032 & & 0.110 & 0.093 & & 17.497 \\
    & DN & & 512 & 512 & & 4 & 16 & & 0.009 & \textbf{0.027} & & 0.104 & 0.077 & & 17.154 \\
    \midrule
    \rowcolor{Gray}
    \multicolumn{16}{l}{\textit{Mixture Components}} \\
    & DN & & 256 & 256 & & 2 & 16 & & 0.014 & \textbf{0.027} & & 0.092 & 0.071 & & \textbf{16.944} \\
    & DN & & 256 & 256 & & 4 & 16 & & 0.012	& 0.028 & & \textbf{0.078} & \textbf{0.060} & & 17.032 \\
    & DN & & 256 & 256 & & 2 & 32 & & \textbf{0.009} & 0.030 & & 0.109 & 0.098 & & 17.216\\
    & DN & & 256 & 256 & & 4 & 32 & & \textbf{0.009} & 0.031 & & 0.103 & 0.076 & & 17.252 \\
    \bottomrule
  \end{tabular}
  }
  \vspace{-0.2cm}
  \caption{Ablation study results comparing different model configurations and hyperparameters. We report the results for ResNet50 (RN) and DenseNet201 (DN) visual backbones, various embedding vector dimensions for the image representation and the TPP history, and different numbers of Gaussian mixture components.
  }
  \label{tab:ablation}
  \vspace{-0.45cm}
\end{table}

\begin{table*}[t]
  \centering
  \small
  \setlength{\tabcolsep}{.2em}
  \resizebox{\linewidth}{!}{
  \begin{tabular}{lc >{}c>{}c>{}c>{}c>{}c>{}c c >{}c>{}c c >{}c>{}c c >{}c cc cccccc c cc c >{}c>{}c c c c}
    \toprule
    & & \multicolumn{14}{c}{\textbf{COCO-FreeView}} & & & \multicolumn{14}{c}{\textbf{MIT1003}} \\ 
    \midrule
    & & \multicolumn{6}{c}{MM (KL-Div) $\downarrow$} & & \multicolumn{2}{c}{SM (KL-Div) $\downarrow$} & & \multicolumn{2}{c}{SS (KL-Div) $\downarrow$} & & \multicolumn{1}{c}{SED $\downarrow$} & & & \multicolumn{6}{c}{MM (KL-Div)$\downarrow$} & & \multicolumn{2}{c}{SM (KL-Div) $\downarrow$} & & \multicolumn{2}{c}{SS (KL-Div) $\downarrow$} & & \multicolumn{1}{c}{SED $\downarrow$} \\
    \cmidrule{3-8} \cmidrule{10-11} \cmidrule{13-14} \cmidrule{16-16} \cmidrule{19-24} \cmidrule{26-27} \cmidrule{29-30} \cmidrule{32-32}
    & & Sh & Len & Dir & Pos & Dur & Avg & & w/ Dur & w/o Dur & & w/ Dur & w/o Dur & & Avg & & & Sh & Len & Dir & Pos & Dur & Avg & & w/ Dur & w/o Dur & & w/ Dur & w/o Dur & & Avg \\
    \midrule
    Itti-Koch~\cite{IttiKoch98} & & 0.42 & 0.40 & 0.21 & 1.02 & - & 0.51 & & - & 2.54 & & - & 1.01 & & 14.00 & & & 0.91 & 0.64 & 0.71 & 1.53 & - & 0.95 &  & - & 2.27 & & - & 6.30 & & 8.86 \\
    CLE (Itti)~\cite{bfpha04,IttiKoch98} & & 0.07 & 0.30 & 0.35 & 1.43 & - & 0.54 & & - & 2.50 & & -  & 1.27 & & 14.37 & & & 0.10 & 0.10 & 0.32 & 1.04 & - & 0.39 & & - & 2.16 & & - & 6.25 & & 9.09 \\
    CLE (DG)~\cite{bfpha04,kummerer2014deep} & & 0.06 & 0.18 & 0.23 & 1.31 & - & 0.44 & & - & 2.37 & & - & 1.22 & & 14.31 & & & - & - & - & - & - & - & & - & - & & - & - & & - \\
    G-Eymol~\cite{zanca2019gravitational} & & 0.37 & 0.73 & 0.93 & 1.22 & 1.99 & 1.05 & & 9.00 & 6.67 & & 8.75 & 6.30 & & 14.20 & & & 0.68 & 0.68 & 0.46 & 1.54 & 1.03 & 0.88 & & 15.90 & 4.89 & & 3.32 & 6.96 & & \underline{6.96} \\
    IOR-ROI-LSTM~\cite{chen2018scanpath} & & 1.15 & 0.47 & 0.03 & 0.19 & 0.05 & 0.38 & & 1.54 & 0.76 & & 0.56 & 0.64 & & 13.55 & & & 0.59 & 0.27 & \underline{0.07} & 0.57 & 0.05 & 0.31 & & 0.69 & 0.45 & & 5.08 & 8.61 & & 8.61 \\
    DeepGazeIII~\cite{kummerer2022deepgaze} & & 0.04 & 0.02 & 0.03 & \underline{\textbf{0.03}} & - & \underline{\textbf{0.03}} & & - & 0.33 & & - & 0.33 & & 13.15 & & & - & - & - & - & - & - & & - & - & & - & - & & - \\
    Scanpath-VQA~\cite{chen2021predicting} & & 0.05 & 0.16 & 0.10 & 0.06 & 0.25 & 0.12 & & 1.07 & 0.34 & & 0.43 & 0.28 & & \underline{12.76} & & & 0.04 & 0.05 & 0.08 & 0.05 & 0.14 & 0.07 & & \underline{0.06} & \underline{0.05} & & \underline{0.05} & \underline{0.11} & & 7.26 \\
    \midrule
    \rowcolor{Gray}
    DeepGazeIII~\cite{kummerer2022deepgaze} & & \underline{\textbf{0.01}} & 0.03 & 0.05 & 0.05 & - & 0.04 & & - & 0.34 & & - & 0.36 & & \textbf{13.15} & & & 0.05 & \underline{\textbf{0.01}} & 0.20 & 0.05 & - & 0.08 & & - & 0.19 & & - & 5.06 & & 8.28 & \\
    \rowcolor{Gray}
    Scanpath-VQA~\cite{chen2021predicting} & & 0.62 & 0.41 & \underline{\textbf{0.02}} & 0.05 & 0.03 & 0.23 & & \underline{\textbf{0.08}} & \underline{\textbf{0.03}} & & \underline{\textbf{0.03}} & 0.31 & & 14.34 & & & 0.20 & 0.14 & 0.15 & 0.08 & 0.02 & 0.12 & & 0.23 & 0.19 & & 0.14 & 0.25 & & 9.27 \\
    \rowcolor{Gray} 
    \ours (GRU) & & 0.06 & 0.02 & \underline{\textbf{0.02}} & \underline{\textbf{0.03}} & \underline{\textbf{0.01}} & \underline{\textbf{0.03}} & & \underline{\textbf{0.08}} & 0.06 & & 0.05 & \underline{\textbf{0.11}} & & 17.03 & & & \underline{\textbf{0.01}} & 0.03 & \textbf{0.09} & \underline{\textbf{0.04}} & \underline{\textbf{0.01}} & \underline{\textbf{0.04}} & & \textbf{0.15} & \textbf{0.11} & & \textbf{0.12} & \underline{\textbf{0.11}} & & \textbf{7.21} \\
    \rowcolor{Gray}
    \ours (Trans.) & & 0.05 & \underline{\textbf{0.01}} & \underline{\textbf{0.02}} & \underline{\textbf{0.03}} & \underline{\textbf{0.01}} & \underline{\textbf{0.03}} & & 0.10 & 0.07 & & 0.06 & 0.12 & & 16.93 & & &  \underline{\textbf{0.01}} & 0.02 & \textbf{0.09} & 0.07 & 0.02 & \underline{\textbf{0.04}} & & 0.22 & 0.16 & & 0.14 & 0.14 & & 7.33 \\
    \bottomrule
  \end{tabular}
  }
  \vspace{-0.2cm}
  \caption{Comparison of various models on COCO-FreeView and MIT1003.
  \colorbox{Gray}{\textbf{Gray color}} indicates models trained under the same settings and datasets. Within this group, \textbf{bold} values represent the best performance for each metric. \underline{Underline} values indicate the overall best performance across all models and metrics.}
  \label{tab:cross1}
  \vspace{-0.25cm}
\end{table*}

\begin{table*}[ht]
  \centering
  \small
  \setlength{\tabcolsep}{.2em}
  \resizebox{\linewidth}{!}{
  \begin{tabular}{lc cccccc c cc cc cccccc c cc cc cccccc c cc}
    \toprule
    & & \multicolumn{9}{c}{\textbf{OSIE}} & & & \multicolumn{9}{c}{\textbf{NUSEF}} & & & \multicolumn{9}{c}{\textbf{FiFa}} \\ 
    \midrule
    & & \multicolumn{6}{c}{MM (KL-Div) $\downarrow$} & & \multicolumn{2}{c}{SM (KL-Div) $\downarrow$} & & & \multicolumn{6}{c}{MM (KL-Div) $\downarrow$} & & \multicolumn{2}{c}{SM (KL-Div) $\downarrow$} & & & \multicolumn{6}{c}{MM (KL-Div) $\downarrow$} & & \multicolumn{2}{c}{SM (KL-Div) $\downarrow$} \\
    \cmidrule{3-8} \cmidrule{10-11} \cmidrule{14-19} \cmidrule{21-22} \cmidrule{24-30} \cmidrule{32-33}
    & & Sh & Len & Dir & Pos & Dur & Avg & & w/ Dur & w/o Dur & & & Sh & Len & Dir & Pos & Dur & Avg & & w/ Dur & w/o Dur & & & Sh & Len & Dir & Pos & Dur & Avg & & w/ Dur & w/o Dur \\
    \midrule
    Itti-Koch~\cite{IttiKoch98} & & 1.62 & 0.89 & 0.45 & 3.69 & - & 1.66 & & - & 2.22 & & & 0.63 & 0.44 & 0.17 & 0.56 & - & 0.45 &  & - & 0.61 & & & 1.51 & 0.51 & 1.08 & 3.46 & - & 1.64 &  & - & 6.08 \\
    CLE (Itti)~\cite{bfpha04,IttiKoch98} & & 0.13 & \underline{0.03} & 0.20 & 0.75 & - & 0.28 & & - & 1.98 & & & 0.26 & 0.03 & 0.09 & 0.42 & - & 0.20 &  & - & 0.79 &  &  & 0.38 & 0.10 & 0.29 & 1.14 & - & 0.48 &  & - & 3.97 \\
    CLE (DG)~\cite{bfpha04,kummerer2014deep} & & 0.17 & \underline{0.03} & 0.15 & 0.60 & - & 0.24 & & - & 1.43 & & & 0.28 & 0.06 & 0.06 & 0.18 & - & 0.15 &  & - & 0.50 & & & 0.40 & 0.14 & 0.36 & 0.97 & - & 0.46 &  & - & 3.10 \\
    G-Eymol~\cite{zanca2019gravitational} & & 1.18 & 1.08 & 0.25 & 2.12 & 1.18 & 1.16 & & 16.17 & 7.29 & & & 0.38 & 0.30 & 0.05 & 0.29 & 3.02 & 0.81 &  & 1.76 & 0.55 &  &  & 0.34 & 0.57 & 0.59 & 2.48 & 2.40 & 1.28 &  & 17.36 & 11.71 \\
    IOR-ROI-LSTM~\cite{chen2018scanpath} & & 1.72 & 0.73 & \underline{0.03} & 0.96 & 0.03 & 0.69 & & 0.75 & 0.76 & & & 0.90 & 0.36 & 0.12 & 0.23 & 0.17 & 0.36 &  & 0.11 & 0.13 & &  & 1.24 & 0.51 & \underline{0.10} & 1.71 & \underline{0.05} & 0.72 &  & 1.25 & 1.56 \\
    DeepGazeIII~\cite{kummerer2022deepgaze} & & 0.14 & 0.08 & 0.06 & 0.15 & - & 0.11 & & - & 0.12 & & & 0.10 & 0.06 & 0.08 & 0.05 & - & 0.07 &  & - & 0.07 & & & 0.28 & 0.12 & 0.21 & 0.34 & - & 0.24 &  & - & 0.60 \\
    Scanpath-VQA~\cite{chen2021predicting} & & 0.07 & 0.07 & 0.04 & \underline{0.04} & 0.16 & 0.08 & & \underline{0.03} & \underline{0.03} & & & 0.11 & 0.04 & 0.02 & 0.05 & \underline{0.08} & 0.06 &  & \underline{0.02} & 0.03 & & & 0.14 & \underline{0.04} & 0.13 & \underline{0.07} & 0.12 & \underline{0.10} &  & \underline{0.03} & \underline{0.13} \\
    \midrule
    \rowcolor{Gray}
    DeepGazeIII~\cite{kummerer2022deepgaze} & & 0.04 & \underline{\textbf{0.03}} & 0.09 & 0.14 & - & 0.08 &  & - & \textbf{0.22} & & & 0.11 & 0.07 & 0.09 & 0.04 & - & 0.08 &  & - & 0.06 &  &  & 0.25 & 0.13 & 0.40 & \textbf{0.18} & - & 0.24 &  & - & 0.69  \\
    \rowcolor{Gray}
    Scanpath-VQA~\cite{chen2021predicting} & & 0.49 & 0.35 & 0.09 & 0.20 & \underline{\textbf{0.02}} & 0.23 & & 0.40 & 0.28 & & & 0.11 & 0.07 & 0.06 & 0.03 & 0.16 & 0.09 &  & 0.06 & 0.06  & & & 0.44 & 0.26 & 0.33 & 0.31 & 0.08 & 0.28 &  & 0.47 & 0.79  \\
    \rowcolor{Gray}
    \ours (GRU) & & 0.03 & 0.04 & \textbf{0.05} & \textbf{0.12} & 0.03 & \underline{\textbf{0.05}} & & \textbf{0.20} & 0.30 & & & \underline{\textbf{0.03}} & 0.02 & \underline{\textbf{0.01}} & 0.02 & \textbf{0.10} & \underline{\textbf{0.04}} & & \textbf{0.04} & 0.04 & & & \underline{\textbf{0.05}} & \textbf{0.05} & \textbf{0.12} & 0.25 & \underline{\textbf{0.05}} & \underline{\textbf{0.10}} & & \textbf{0.23} & \textbf{0.47} \\
    \rowcolor{Gray}
    \ours (Trans.) & & \underline{\textbf{0.02}} & 0.04 & 0.06 & 0.14 & 0.05 & 0.06 & & 0.25 & 0.44 & & & \underline{\textbf{0.03}} & \underline{\textbf{0.01}} & 0.02 & \underline{\textbf{0.01}} & 0.13 & \underline{\textbf{0.04}} & & \textbf{0.04} & \underline{\textbf{0.01}} & & & 0.06 & \textbf{0.05} & \textbf{0.12} & 0.30 & \underline{\textbf{0.05}} & 0.12 & & 0.32 & 0.52 \\
    \bottomrule
  \end{tabular}
  }
  \vspace{-0.2cm}
  \caption{Comparison of various models on OSIE, NUSEF, and FiFa datasets.
  \colorbox{Gray}{\textbf{Gray color}} indicates models trained under the same settings and datasets. Within this group, \textbf{bold} values represent the best performance for each metric. \underline{Underline} values indicate the overall best performance across all models and metrics.}
  \label{tab:cross2}
  \vspace{-0.4cm}
\end{table*}

\tit{Evaluation Protocol}
We compare the scanpaths synthesised from various models with those recorded from human observers. The objective is to evaluate whether the simulated behaviours exhibited statistical properties closely resembling those exhibited by human subjects who are eye-tracked while viewing a given stimulus.
The evaluation protocol unfolds as follows. Suppose there are $N_{obs}$ human observers. For each stimulus, we first compute the evaluation scores for every possible pair of the $N_{obs}$ observers (Real vs. Real). Then, for each model, (i) we generate gaze trajectories from artificial observers and (ii) calculate the evaluation scores for every possible pair of real and artificial scanpaths (Real vs. Simulated).

For a given metric this procedure yields a target distribution $P$ of similarity scores between observers (Real vs. Real) and a distribution $Q$ of similarity scores for the given model \wrt humans (Real vs. Simulated). As reported in~\cite{kummerer2023predicting}, MM, SM, and SS average values may deliver inconsistent results: models exhibiting less variability \wrt humans, can score systematically better than the ground truth model. This issue can be tackled by considering a good model as the one that minimises the discrepancy between the target and model-derived score distributions. We quantified such discrepancy using the Kullback-Leibler Divergence (KL-Div): $D_{\mathrm{KL}}(P\parallel Q)=\sum_{x\in\mathcal{X}}P(x)\log(P(x)/Q(x))$. Conversely, as SED is an evaluation metric not requiring alignment, it is not susceptible to the inconsistency issues associated with MM, SM, and SS. Consequently, its values are directly reported without any further processing.

\subsection{Scanpath Prediction}

\begin{figure*}[t]
    \centering
    \footnotesize
    \setlength{\tabcolsep}{.24em}
    \resizebox{\linewidth}{!}{
    \begin{tabular}{cccccc}
         G-Eymol~\cite{zanca2019gravitational} & IOR-ROI-LSTM~\cite{chen2018scanpath} & DeepGazeIII~\cite{kummerer2022deepgaze} & Scanpath-VQA~\cite{chen2021predicting} & \textbf{\ours (Ours)} & Humans \\
         \addlinespace[0.08cm]
         \includegraphics[width=0.16\linewidth]{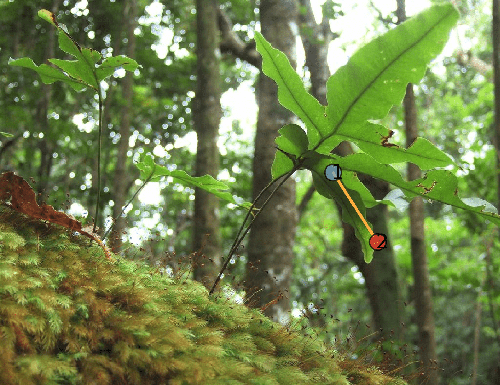} &
         \includegraphics[width=0.16\linewidth]{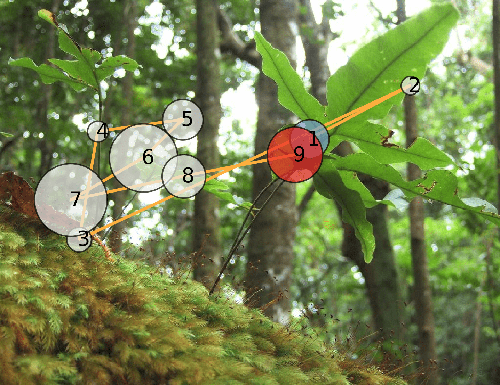} &
         \includegraphics[width=0.16\linewidth]{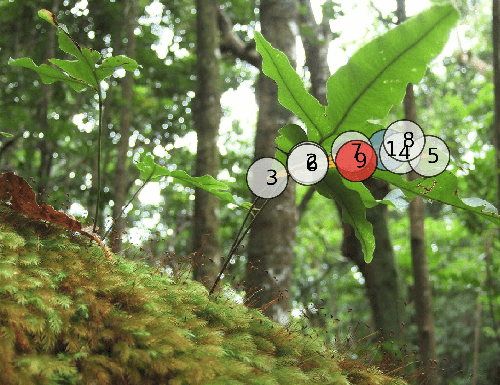} & 
         \includegraphics[width=0.16\linewidth]{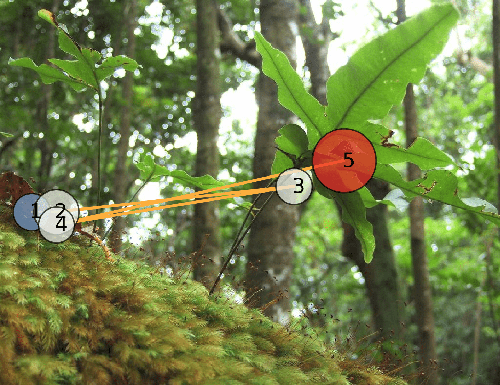} & 
         \includegraphics[width=0.16\linewidth]{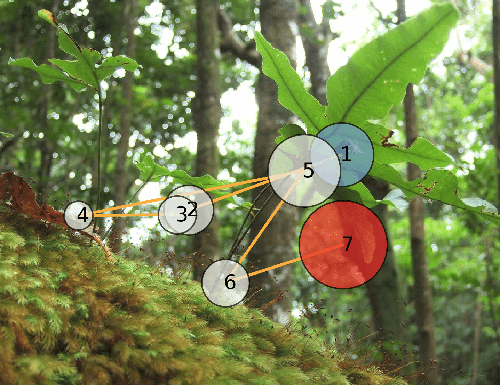} & 
         \includegraphics[width=0.16\linewidth]{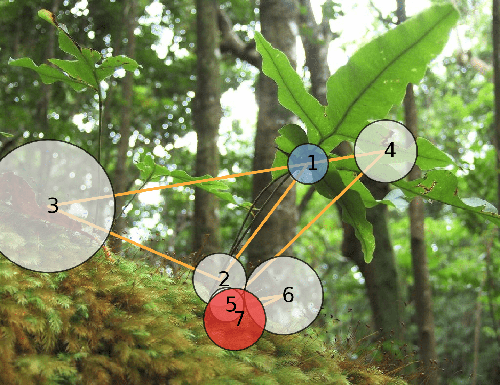} \\
         \includegraphics[width=0.16\linewidth]{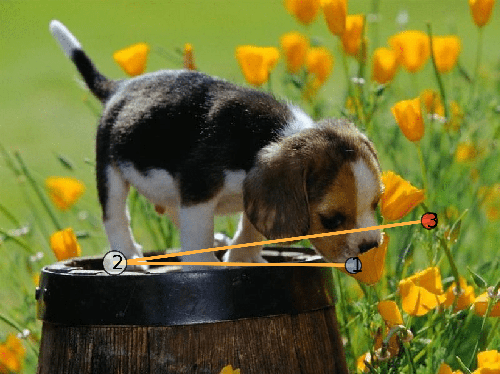} &
         \includegraphics[width=0.16\linewidth]{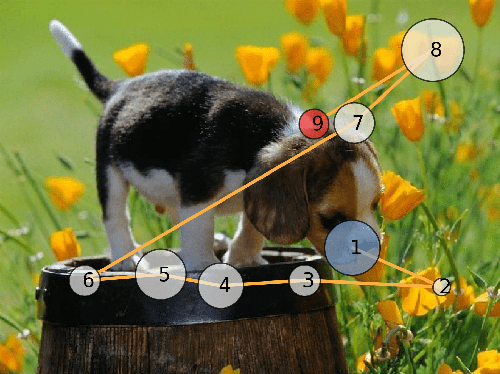} &
         \includegraphics[width=0.16\linewidth]{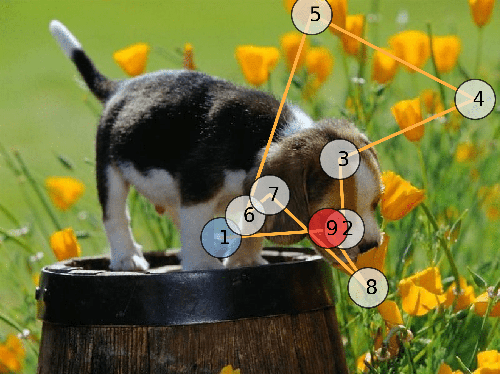} & 
         \includegraphics[width=0.16\linewidth]{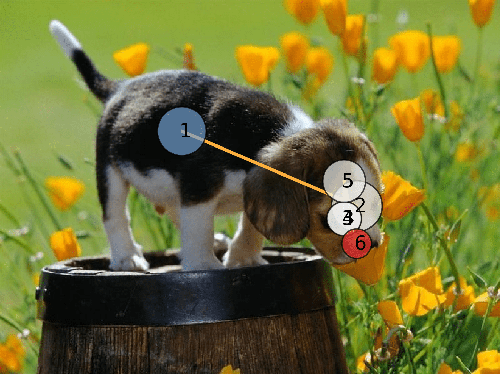} & 
         \includegraphics[width=0.16\linewidth]{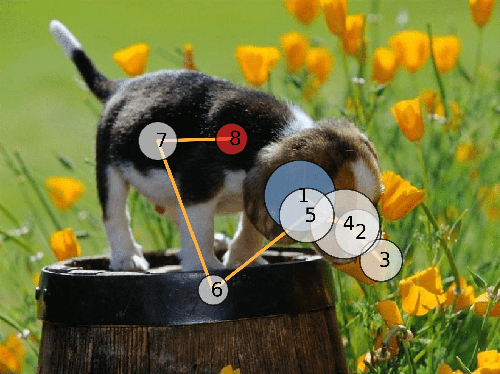} & 
         \includegraphics[width=0.16\linewidth]{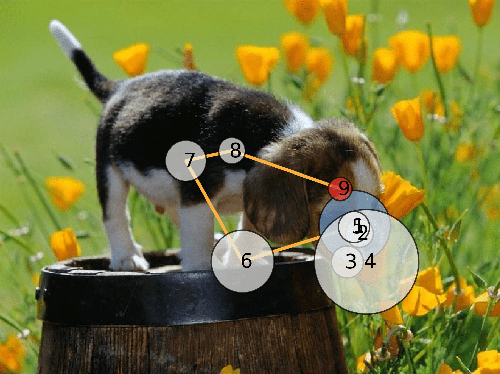} \\
    \end{tabular}
    }
    \vspace{-0.3cm}
    \caption{Comparison of simulated and human scanpaths. Each circle represents a fixation point, with its diameter proportional to the fixation duration. For methods that do not model fixation duration, circles are shown with a uniform size.
    }
    \label{fig:qualitatives}
    \vspace{-0.5cm}
\end{figure*}

\begin{figure}[t]
    \centering
    \footnotesize
    \resizebox{\linewidth}{!}{
    \setlength{\tabcolsep}{.2em}
    \begin{tabular}{cc}
    \includegraphics[height=0.49\linewidth]{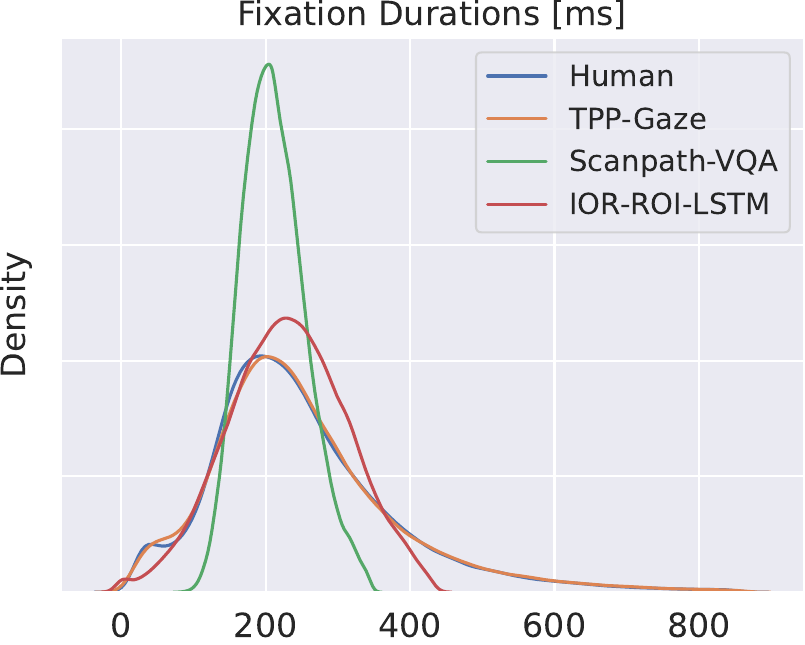} &
    \includegraphics[height=0.49\linewidth]{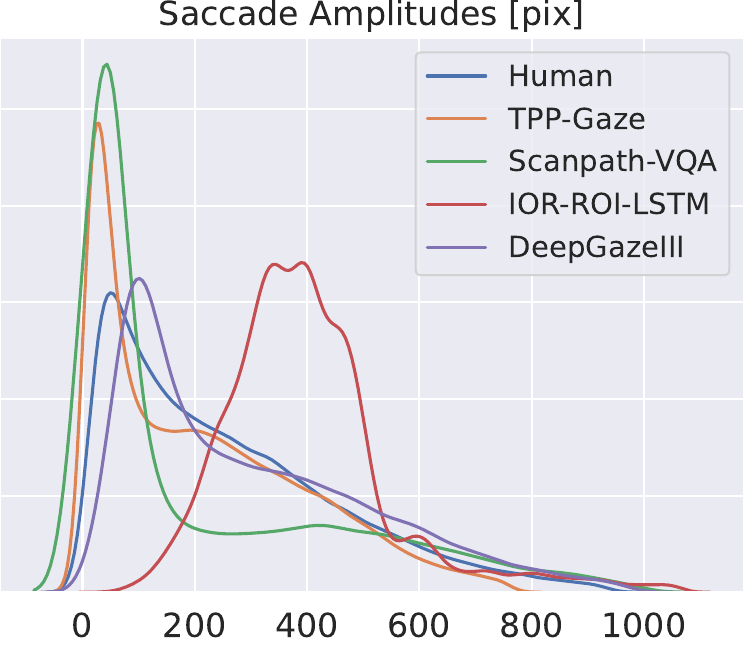} \\
    \addlinespace[0.05cm]
    \includegraphics[height=0.49\linewidth]{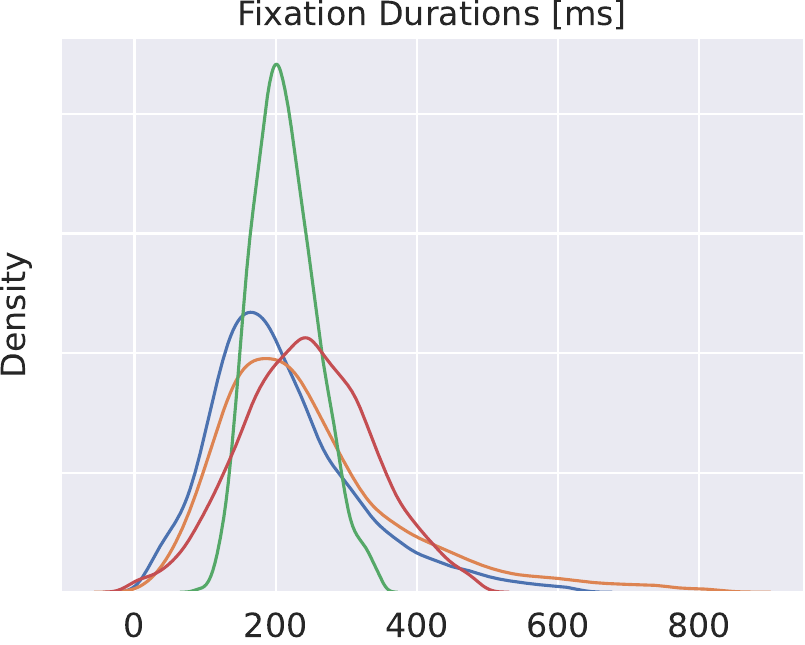} &
    \includegraphics[height=0.49\linewidth]{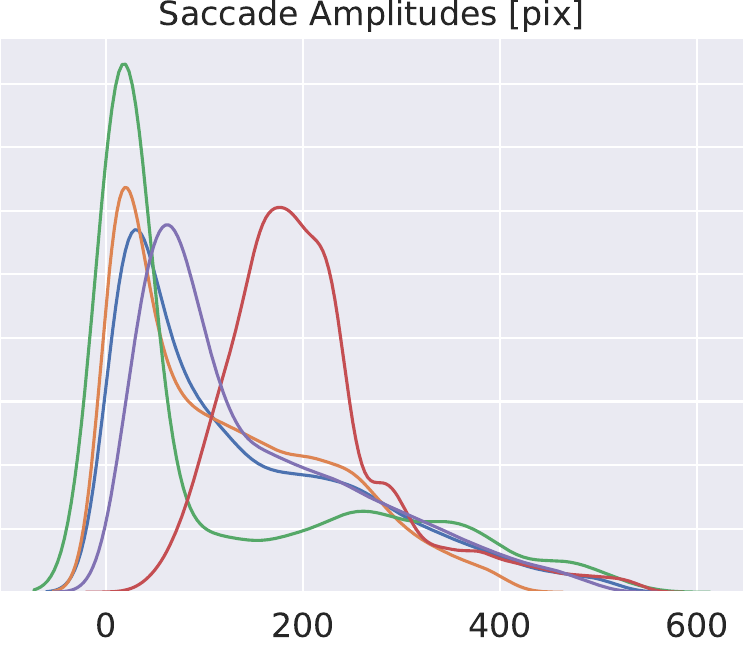} \\
    \end{tabular}
    }
    \vspace{-0.3cm}
    \caption{Statistical properties exhibited by \tppgaze and other methods relative to those of human observers, in terms of empirical fixation durations and saccade amplitudes on the COCO-FreeView (top row) and OSIE (bottom row) datasets.}
    \label{fig:distributions}
    \vspace{-0.5cm}
\end{figure}

\tinytit{Ablation Studies} The \tppgaze architecture consists of three main blocks: image encoding (CNN backbone), history encoding (RNN/Transformer), and fixation/inter-time prediction (GMM/LGMM). To break down these components and make the adopted design choices explicit, we perform extensive ablation studies. Specifically, we evaluate two different CNN backbones for image encoding (\ie, a ResNet50~\cite{he2016deep} and a DenseNet201~\cite{huang2017densely}) as well as three embedding vector dimensions for the image semantic representation ($\mathbf{z}_j$) and the history embedding ($\mathbf{h}_n$, TPP dimensionality). Moreover, different numbers of components for the GMM/LGMM are considered. Table~\ref{tab:ablation} reports the results of the ablation studies conducted on the COCO-FreeView dataset. In our experiments, we select the hyperparameters yielding the best trade-off according to the considered evaluation metrics, resulting in a DesNet201 backbone and a dimensionality equal to 256 for all embedding vectors. Moreover, the parameters $K$ and $G$ representing mixture components are respectively set to 4 and 16.

\tit{Comparison with the State of the Art} To compare the proposed approach with others, we include state-of-the-art approaches that either reach high performance in recent scanpath benchmarks~\cite{kummerer2021state}, offer source code availability, and are representative of different approaches and architectures. As to the latter criteria, following the taxonomy proposed in~\cite{kummerer2021state}, scanpath models can be aggregated into the following categories: biologically inspired (\eg~Itti-Koch model~\cite{IttiKoch98} and G-Eymol~\cite{zanca2019gravitational}); statistically inspired (\eg~CLE model~\cite{bfpha04}); cognitively inspired (\eg~IOR-ROI-LSTM~\cite{chen2018scanpath}); engineered models (\eg~DeepGazeIII~\cite{kummerer2022deepgaze} and Scanpath-VQA~\cite{chen2021predicting}); but see~\cite{kummerer2021state,kummerer2022deepgaze,kummerer2023predicting} for an in-depth review. Under such circumstances, we assess the performance of \tppgaze against the aforementioned models.

Table~\ref{tab:cross1} reports quantitative results on the COCO-FreeView and MIT1003 datasets in terms of all considered metrics, while model performance on OSIE, NUSEF, and FiFa are shown in Table~\ref{tab:cross2} in terms of MM and SM\footnote{We refer to the supplementary material for the results in terms of SS and SED on OSIE, NUSEF, and FiFa datasets.}. In all experiments, we compare the aforementioned approaches using the pre-trained model weights released by the authors. As DeepGaze models were trained on the entire MIT1003 dataset, the results from DeepGazeIII and CLE (DG) have not been included in this comparison. Additionally, to explicitly measure the effect of the proposed architecture and mathematical framework, we retrain and test the two most recent models (DeepGazeIII and Scanpath-VQA) under the same conditions adopted for \tppgaze (see Sec.~\ref{sec:exper_setup}). Specifically, beyond training on the same data, the large-scale pre-training of DeepGazeIII as well as the fine-tuning stage of Scanpath-VQA based on reinforcement learning~\cite{rennie2017self} have been inhibited.
These results are reported in gray color at the bottom of the tables.

\begin{figure}[t]
    \centering
    \footnotesize
    \resizebox{\linewidth}{!}{
    \setlength{\tabcolsep}{.2em}
    \begin{tabular}{cc}
    \includegraphics[height=0.49\linewidth]{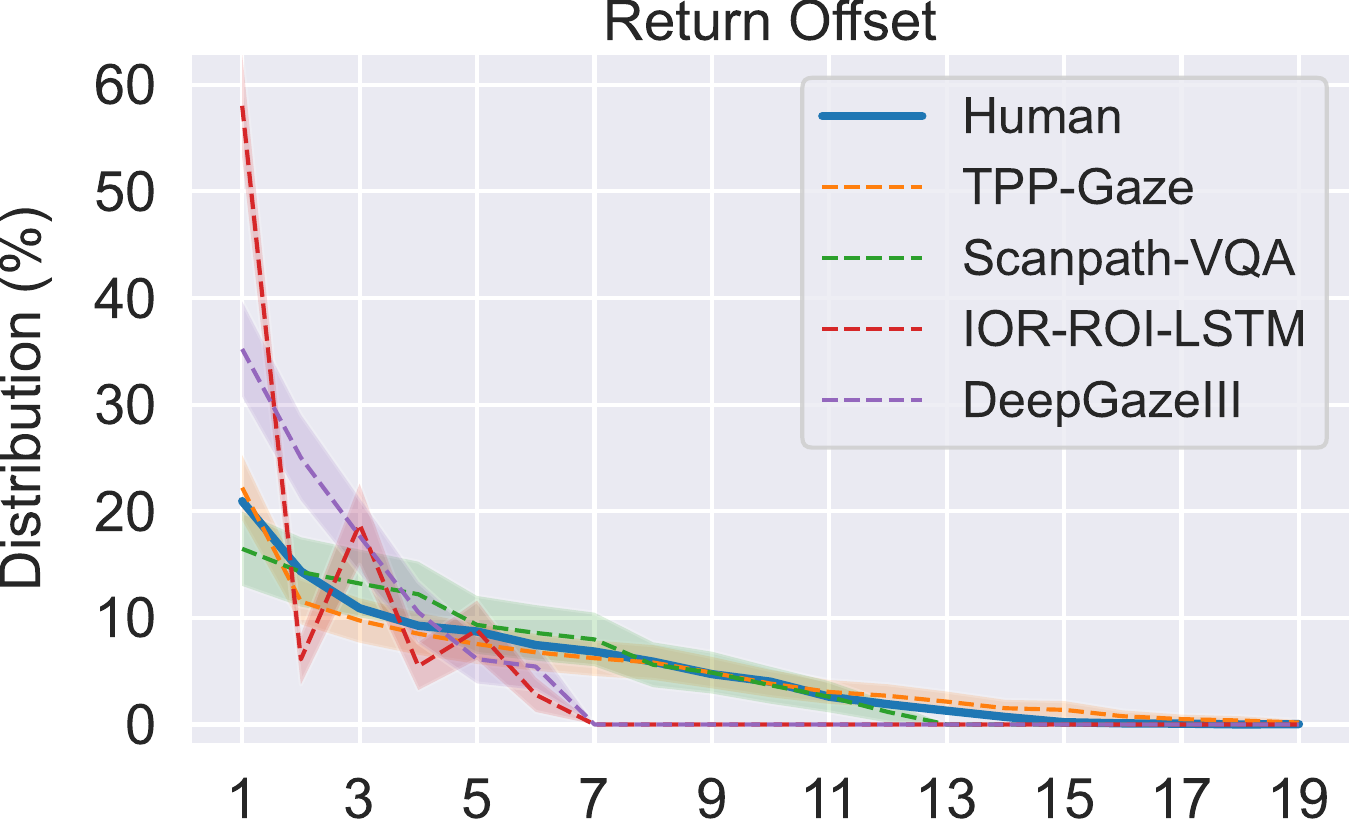} &
    \includegraphics[height=0.49\linewidth]{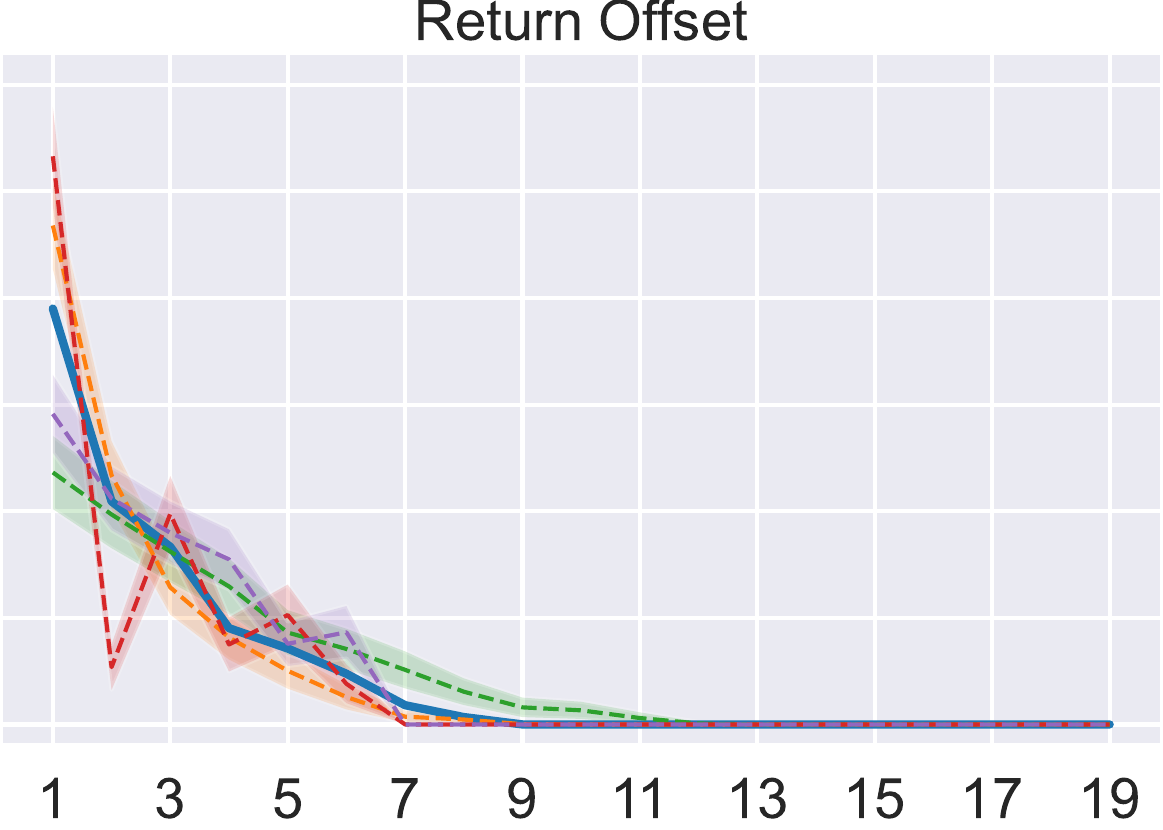} \\
    \end{tabular}
    }
    \vspace{-0.3cm}
    \caption{Return fixations analysis comparing \tppgaze with other methods and human observers. Results are shown on  COCO-FreeView (left plot) and OSIE (right plot) datasets.}
    \label{fig:return_fixations}
    \vspace{-0.5cm}
\end{figure}

\begin{table*}
  \centering
  \small
  \setlength{\tabcolsep}{.3em}
  \resizebox{\linewidth}{!}{
  \begin{tabular}{lc ccc c ccc c ccc c ccc c ccc}
    \toprule
    & & \multicolumn{3}{c}{\textbf{COCO-FreeView}} & & \multicolumn{3}{c}{\textbf{MIT1003}} & & \multicolumn{3}{c}{\textbf{OSIE}} & & \multicolumn{3}{c}{\textbf{NUSEF}} & & \multicolumn{3}{c}{\textbf{FiFa}} \\
    \cmidrule{3-5} \cmidrule{7-9} \cmidrule{11-13} \cmidrule{15-17} \cmidrule{19-21}
    & & KL-Div $\downarrow$ & AUC $ \uparrow$ & NSS $ \uparrow$ & & KL-Div $\downarrow$ & AUC $ \uparrow$ & NSS $ \uparrow$ & & KL-Div $\downarrow$ & AUC $ \uparrow$ & NSS $ \uparrow$ & & KL-Div $\downarrow$ & AUC $ \uparrow$ & NSS $ \uparrow$ & & KL-Div $\downarrow$ & AUC $ \uparrow$ & NSS $ \uparrow$ \\
    \midrule
    \multicolumn{16}{l}{\textit{Saliency-based}} \\
    \hspace{0.4cm}CLE (DG)~\cite{bfpha04,kummerer2014deep} & & 8.65 & 0.55 & 0.09 & & - & - & - & & 5.08 & 0.59 & 0.28 & & 4.99 & 0.63 & 0.38 & & 6.39 & 0.59 & 0.25 \\    
    \hspace{0.4cm}DeepGazeIII~\cite{kummerer2022deepgaze} & & 0.85 & 0.84 & 1.75 & & - & - & - & & 0.32 & 0.87 & 2.01 & & 0.49 & 0.85 & 1.89 & & 0.62 & 0.88 & 2.52 \\
    \midrule
    \multicolumn{16}{l}{\textit{Saliency-free}} \\
    \hspace{0.4cm}Itti-Koch~\cite{IttiKoch98} & & 8.94 & 0.56 & 0.24 & & 5.01 & 0.64 & 0.47 & & 3.35 & 0.65 & 0.51 & & 4.84 & 0.63 & 0.40 & & 5.47 & 0.64 & 0.42 \\
    \hspace{0.4cm}CLE (Itti)~\cite{bfpha04,IttiKoch98} & & 7.45 & 0.54 & 0.07 & & 4.15 & 0.61 & 0.23 & & 3.45 & 0.61 & 0.23 & & 3.36 & 0.63 & 0.31 & & 4.84 & 0.60 & 0.23 \\    
    \hspace{0.4cm}G-Eymol~\cite{zanca2019gravitational} & & 10.98 & 0.56 & 0.26 & & 7.64 & 0.62 & 0.35 & & 4.58 & 0.67 & 0.60 & & 5.09 & 0.66 & 0.55 & & 9.04 & 0.62 & 0.47 \\
    \hspace{0.4cm}IOR-ROI-LSTM~\cite{chen2018scanpath} & & 1.30 & 0.77 & 0.99 & & \textbf{0.78} & 0.81 & 1.40 & & \textbf{0.50} & \underline{0.83} & 1.46 & & \textbf{0.74} & \underline{0.80} & 1.32 & & \textbf{0.83} & \underline{0.85} & 1.72 \\
    \hspace{0.4cm}Scanpath-VQA~\cite{chen2021predicting} & & 3.53 & 0.77 & \underline{1.56} & & 2.12 & 0.82 & \underline{2.01} & & 1.26 & \textbf{0.84} & \textbf{2.12} & & 2.45 & \underline{0.80} & \textbf{1.76} & & 1.88 & \textbf{0.86} & \textbf{2.89} \\
    \rowcolor{Gray}
    \hspace{0.4cm}\ours (GRU) & & \textbf{1.01} & \textbf{0.84} & \textbf{1.65} & & \textbf{0.78} & \textbf{0.86} & \textbf{2.06} & & \underline{0.67} & \textbf{0.84} & \underline{1.72} & & 0.84 & \textbf{0.84} & \underline{1.71}& & \underline{1.07} & \textbf{0.86} & \underline{2.06} \\
    \rowcolor{Gray} 
    \hspace{0.4cm}\ours (Transformer) & & \underline{1.11} & \underline{0.83} & 1.54 & & \underline{0.83} & \underline{0.85} & 1.93 & & 0.68 & \textbf{0.84} & 1.68 & & \underline{0.79} & \textbf{0.84} & 1.70 & & 1.11 & \underline{0.85} & 1.91 \\
    \bottomrule
  \end{tabular}
  }
  \vspace{-0.2cm}
  \caption{Saliency prediction results on COCO-FreeView, MIT1003, OSIE, NUSEF, and FiFa datasets. 
  Models are grouped into saliency-based and saliency-free methods, where the former (\ie, CLE (DG) and DeepGazeIII) incorporate components trained to predict saliency maps. \textbf{Bold} values represent the best performance within each metric, while \underline{underline} values indicate the second-best results.
  }
  \label{tab:saliency}
  \vspace{-0.4cm}
\end{table*}

As can be observed, when trained under the same settings and datasets, \tppgaze (with either GRU or Transformer-based history encoding) outperforms all the other approaches on most of the adopted metrics. Interestingly, in many cases the proposed approach offers the best overall performance, even when considering the pre-trained models released by the authors, except for ScanMatch where Scanpath-VQA, which is directly optimized via reinforcement learning on this metric, understandably proves to be the best. Some qualitative results are shown in \cref{fig:qualitatives}, where we report sampled scanpaths from five models alongside those from humans. Notably, \tppgaze can predict fixations that better align with those recorded from human subjects, confirming the advantages of the proposed approach for predicting scanpaths during free-viewing.

Additional analyses are reported in \cref{fig:distributions} that shows empirical distributions summarizing \tppgaze's scanpath statistics compared to those yielded by human observers and other methods. Beyond common scanpath statistics, we further evaluate the proposed approach using a return fixations (RF) analysis~\cite{zhang2022look}. RF analysis describes the tendency of observers (either human or simulated) to revisit previously foveated locations. The frequency of RFs and the temporal offset (\ie, the number of intervening fixations before returning to a location) at which they occur, provide a more nuanced description of the cognitive processes underlying attention allocation~\cite{zhang2022look}. \cref{fig:return_fixations} reports the results of this analysis in comparison with existing methods across two datasets. Notably, although \tppgaze was not explicitly trained for this objective, it produces the most accurate RF patterns with respect to human behavior when compared to state-of-the-art approaches\footnote{Results of the RF analysis on OSIE, NUSEF and FiFa are reported in the supplementary material.}.

\subsection{Applications}
\tinytit{Saliency Prediction} The performance of \tppgaze are further evaluated by comparing the saliency maps ``backward'' generated from fixations with those of human observers across all evaluated scanpath models. The results, presented in Table~\ref{tab:saliency}, are measured using three commonly adopted saliency metrics~\cite{mit-saliency-benchmark,8315047}: Kullback-Leibler Divergence (KL-Div), Judd's Area Under the Curve (AUC), and Normalised Scanpath Saliency (NSS). DeepGazeIII and CLE (DG) are reported here only as references for the performance of a saliency prediction model, given their adoption of an extensive pre-training phase designed expressly for saliency generation (DeepGazeIII), or explicit adoption of a saliency model (CLE (DG)). Overall, \tppgaze obtains the best or second-best performance across all metrics and datasets. It yields results that are comparable to or surpass those of IOR-ROI-LSTM~\cite{chen2018scanpath} and Scanpath-VQA~\cite{chen2021predicting}, which are significantly better than all other approaches. This further demonstrates the effectiveness of our approach in predicting fixation points that better resemble human scanpaths than those predicted by existing methods.

\begin{figure}[t]
    \centering
    \includegraphics[width=\linewidth]{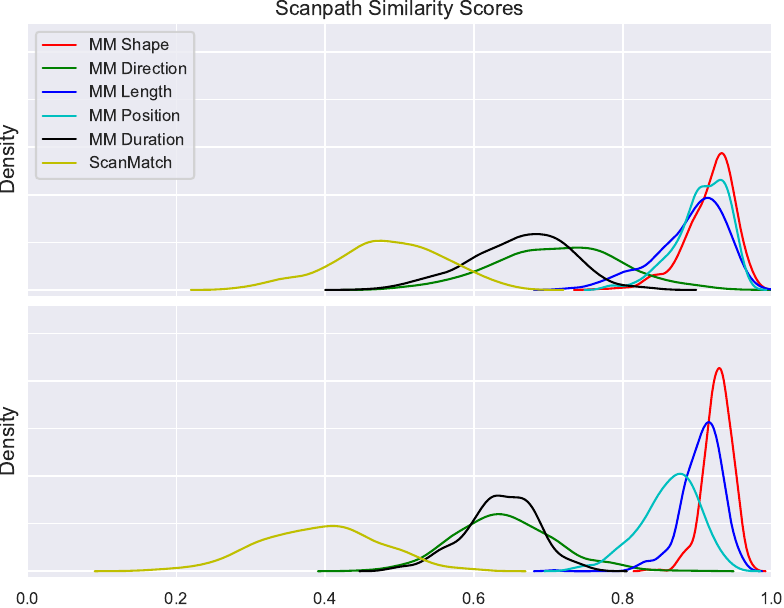}
    \vspace{-0.65cm}
    \caption{Empirical distributions of the adopted metrics quantifying inter-humans (top) and human \textit{vs.} \tppgaze (bottom) scanpath similarity for the visual search task on COCO-Search18.}
    \label{fig:visual_search}
    \vspace{-0.5cm}
\end{figure}

\begin{figure}[t]
    \centering
    \footnotesize
    \resizebox{\linewidth}{!}{
    \setlength{\tabcolsep}{.2em}
    \begin{tabular}{cc}
    \includegraphics[height=0.49\linewidth]{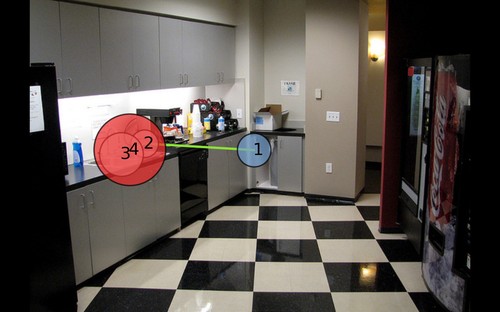} &
    \includegraphics[height=0.49\linewidth]{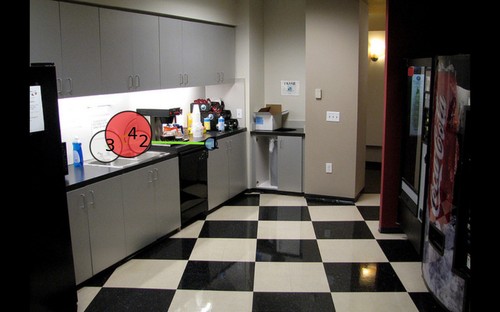} \\
    \end{tabular}
    }
    \vspace{-0.3cm}
    \caption{Human (left) and simulated (right) scanpaths for the visual search task. Search objective is ``Sink".}
    \label{fig:search_qualitative}
    \vspace{-0.5cm}
\end{figure}

\tit{Extending the Model to Visual Search Tasks} Recently, several works~\cite{zhang2018finding,chen2021predicting,yang2022target,mondal2023gazeformer,ding2022efficient} have focused on predicting attention allocation on specific targets (visual search tasks). Although \tppgaze was originally devised and evaluated for the free-viewing scenario, it can be extended to tackle the visual search problem in various ways. Here, we propose a proof-of-concept model featuring a simple architectural variation that enables goal-directed attention prediction with \tppgaze. In a nutshell, we use RoBERTa \cite{liu2019roberta} to perform a linguistic embedding of the search target and learn a target-oriented image semantic representation\footnote{More details and simulations are shown in the supplementary material}. (\textit{cf.} \cref{sec:architecture}). Preliminary results show encouraging trends on the COCO-Search18 dataset \cite{Yang_2020_CVPR,chen2021coco}, as illustrated in \cref{fig:visual_search}, where visual search patterns produced by \tppgaze are compared to human patterns using the MM and SM metrics. A qualitative example is depicted in \cref{fig:search_qualitative}.

\section{Conclusion}
\label{sec:conclusion}
We presented \tppgaze, a novel approach that explicitly models via Neural Temporal Point Processes the temporal evolution of visual attention as instantiated through a scanpath. Cogently,  \tppgaze~allows for a principled modelling of both fixations position and corresponding duration. Extensive experiments on five publicly available datasets have proven the effectiveness of the proposed approach in modelling gaze spatio-temporal dynamics, as witnessed by the overall best performances in scanpath similarity and fixation duration prediction. Also, it exhibits competitive results in terms of saliency prediction. 

\section*{Acknowledgments}
This work was supported by a grant from Università degli Studi di Milano (Bando Linea 3 My First SEED -- DM 737/2021 MUR) and by the PNRR project “Italian Strengthening of Esfri RI Resilience (ITSERR)” funded by the European Union - NextGenerationEU (CUP B53C22001770006).

{\small
\bibliographystyle{ieee_fullname}
\bibliography{bibliography}
}

\appendix
\section*{Supplementary Material}
We introduced \tppgaze, a scanpath prediction method that models gaze dynamics as a neural temporal point process.
In the following sections, we provide additional results showing evidence of the superiority of our proposed approach compared to the state-of-the-art. Additionally, we describe how the proposed approach can be extended for the visual search task.
\section{Additional Quantitative Results}
\tinytit{Additional Metrics on OSIE, NUSEF, and FiFa}
As a complement of Table~3 of the main paper, we report in Table~\ref{tab:cross2_supp} the results on OSIE, NUSEF, and FiFa datasets in terms of SS and SED. Also for these metrics, \tppgaze~achieves the best results when compared with models trained under the same settings and datasets. It is also worth noting that, especially for the NUSEF and FiFa datasets, our approach can achieve the best overall results in terms of SS with and without duration.

\tit{Scanpath Statistics on MIT1003, NUSEF, and FiFa}
As discussed in the main paper, \tppgaze~features scanpath statistics that better align with human behavior when compared with IOR-ROI-LSTM~\cite{chen2018scanpath}, DeepGazeIII~\cite{kummerer2022deepgaze} and Scanpath-VQA~\cite{chen2021predicting}. The same trend is appreciable from \cref{fig:additionaldistributions}. Notably, even when tested on MIT1003, NUSEF, and FiFa, \tppgaze~effectively models the long-tail distribution of both fixation durations and saccade amplitudes.  In contrast, other methods tend to capture only the average human gaze dynamics. An exception is DeepGazeIII on NUSEF, which achieves comparable results for saccade amplitudes but does not model fixation duration. 

\tit{Return Fixations Analysis on MIT1003, NUSEF, and FiFa}
\cref{fig:return_fixations_supp} complements the analysis reported in the main paper by showing the distribution of return fixations (RFs) for the MIT1003, NUSEF, and FiFa datasets. In these settings as well, \tppgaze~demonstrates its ability to model RF patterns effectively, generally presenting an RF distribution that aligns better with human observers compared to other methods.

\begin{table}[t]
  \centering
  \small
  \setlength{\tabcolsep}{.2em}
  \resizebox{\linewidth}{!}{
  \begin{tabular}{lc cc c c cc cc c c cc cc c c}
    \toprule
    & & \multicolumn{4}{c}{\textbf{OSIE}} & & & \multicolumn{4}{c}{\textbf{NUSEF}} & & & \multicolumn{4}{c}{\textbf{FiFa}} \\ 
    \midrule
    & & \multicolumn{2}{c}{SS (KL-Div) $\downarrow$} & & \multicolumn{1}{c}{SED $\downarrow$} & & & \multicolumn{2}{c}{SS (KL-Div) $\downarrow$} & & \multicolumn{1}{c}{SED $\downarrow$} & & & \multicolumn{2}{c}{SS (KL-Div) $\downarrow$} & & \multicolumn{1}{c}{SED $\downarrow$} \\
    \cmidrule{3-4} \cmidrule{6-6} \cmidrule{9-10} \cmidrule{12-12} \cmidrule{15-16} \cmidrule{18-18}
    & & w/ Dur & w/o Dur & & Avg & & & w/ Dur & w/o Dur & & Avg & & & w/ Dur & w/o Dur & & Avg \\
    \midrule
    Itti-Koch~\cite{IttiKoch98} & & - & 3.93 & & 9.07 & & & - & 1.89 & & 9.97 & & & - & 14.70 & & 8.65 \\
    CLE (Itti)~\cite{bfpha04,IttiKoch98} & & - & 3.24 & & 9.29 & & & - & 1.40 & & 10.16 & & & - & 12.62 & & 8.86 \\
    CLE (DG)~\cite{bfpha04,kummerer2014deep} & & - & 3.65 & & 9.23 & & & -  & 1.35 & & 10.07 & & & - & 14.38 & & 8.83 \\
    G-Eymol~\cite{zanca2019gravitational} & & 12.28& 2.95 & & 8.00 & & & 1.99 & 0.53 & & \underline{8.02} & & & 13.17 & 5.00 & & \underline{6.13} \\
    IOR-ROI-LSTM~\cite{chen2018scanpath} & & 0.20 & 2.84 & & 8.82 & & & 0.06 & 1.10 & & 9.69 & & & 0.30 & 12.44 & & 8.27 \\
    DeepGazeIII~\cite{kummerer2022deepgaze} & & - & 2.51 & & 8.47 & & & - & 1.04 & & 9.38 & & & - & 12.08 & & 7.97 \\
    Scanpath-VQA~\cite{chen2021predicting} & & \underline{0.02} & \underline{0.09} & & \underline{7.55} & & & \underline{0.02} & 0.10 & & 8.39 & & & \underline{0.03} & 0.44 & & 6.81 \\
    \midrule
    \rowcolor{Gray}
    DeepGazeIII~\cite{kummerer2022deepgaze} & & - & 2.52 & & \textbf{8.57} & & & - & 1.04 & & 9.42 & & & - & 12.25 & & 8.00 \\
    \rowcolor{Gray}
    Scanpath-VQA~\cite{chen2021predicting} & & 0.29 & 0.31 & & 9.70 & & & 0.06 & 0.18 & & 10.61 & & & 0.35 & 0.90 & & 9.73 \\
    \rowcolor{Gray}
    \ours (GRU) & & \textbf{0.25} & \textbf{0.30} & & 8.05 & & & \underline{\textbf{0.02}} & \underline{\textbf{0.03}} & & 8.41 & & & \textbf{0.15} & \underline{\textbf{0.24}} & & \textbf{7.00} \\
    \rowcolor{Gray}
    \ours (Transf.) & & 0.29 & 0.35 & & 8.10 & & & \underline{\textbf{0.02}} & 0.04 & & \textbf{8.40} & & & 0.21 & 0.31 & & 7.05 \\
    \bottomrule
  \end{tabular}
  }
  \vspace{-0.15cm}
  \caption{Additional results on OSIE, NUSEF, and FiFa datasets.
  \colorbox{Gray}{\textbf{Gray color}} indicates models trained under the same settings and datasets. Within this group, \textbf{bold} values represent the best performance for each metric. \underline{Underline} values indicate the overall best performance across all models and metrics.}
  \label{tab:cross2_supp}
  \vspace{-0.3cm}
\end{table}

\begin{figure}[t]
    \centering
    \footnotesize
    \resizebox{\linewidth}{!}{
    \setlength{\tabcolsep}{.2em}
    \begin{tabular}{cc}
    \includegraphics[height=0.49\linewidth]{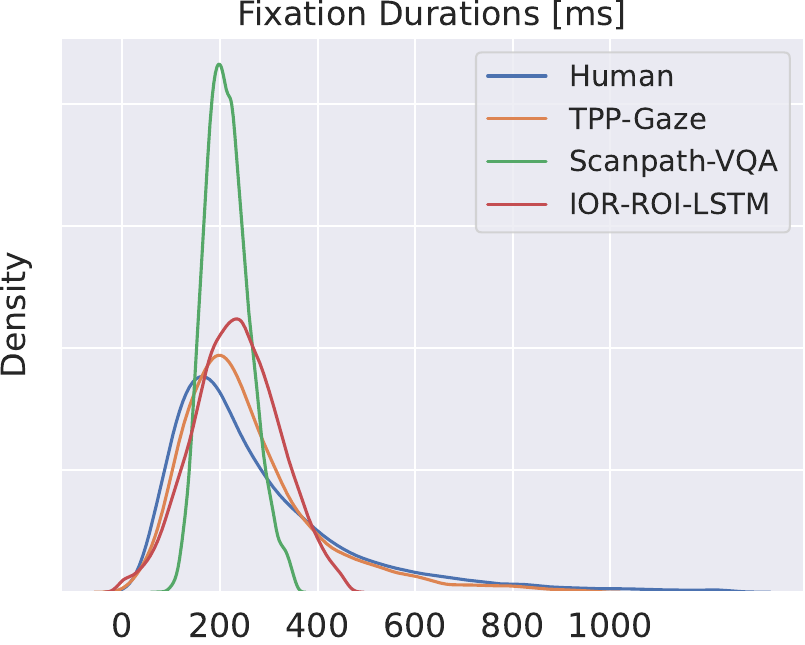} &
    \includegraphics[height=0.49\linewidth]{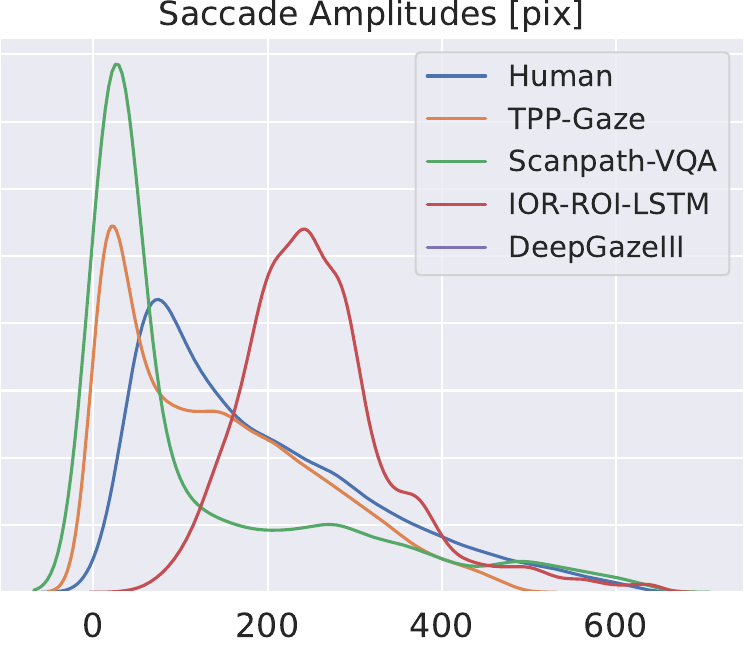} \\
    \addlinespace[0.05cm]
    \includegraphics[height=0.49\linewidth]{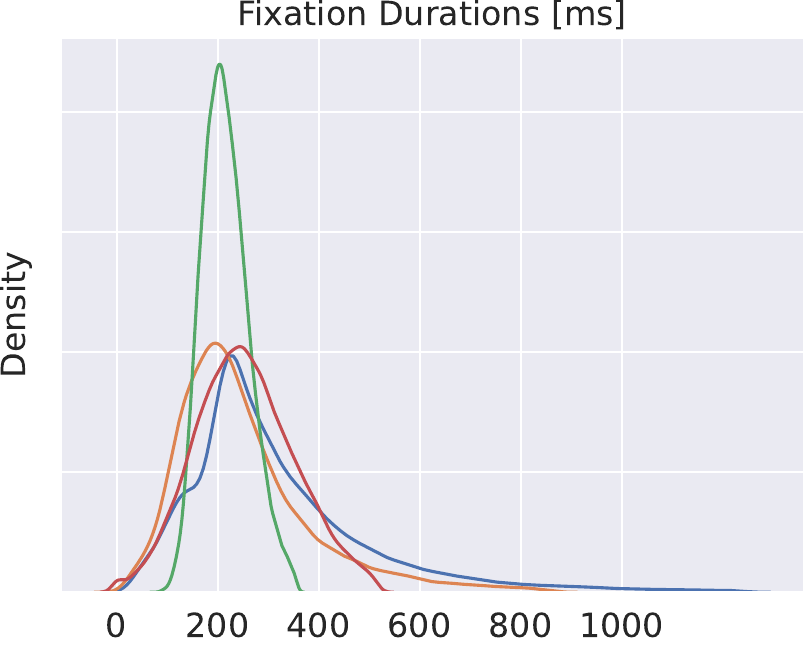} &
    \includegraphics[height=0.49\linewidth]{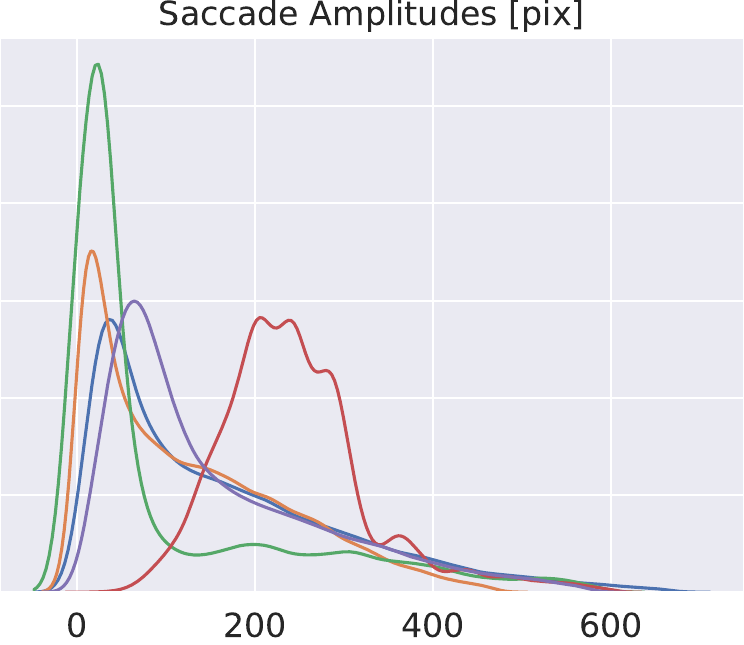} \\
    \addlinespace[0.05cm]
    \includegraphics[height=0.49\linewidth]{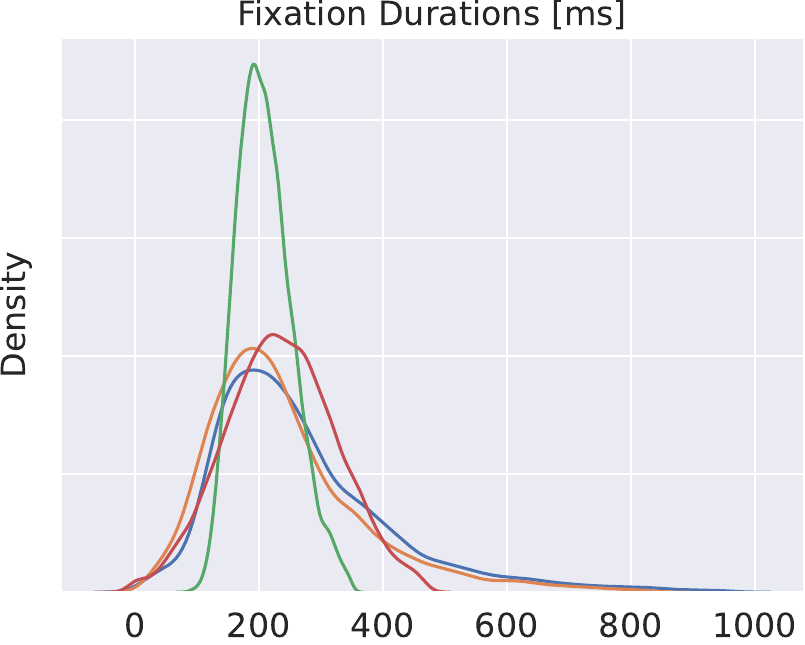} &
    \includegraphics[height=0.49\linewidth]{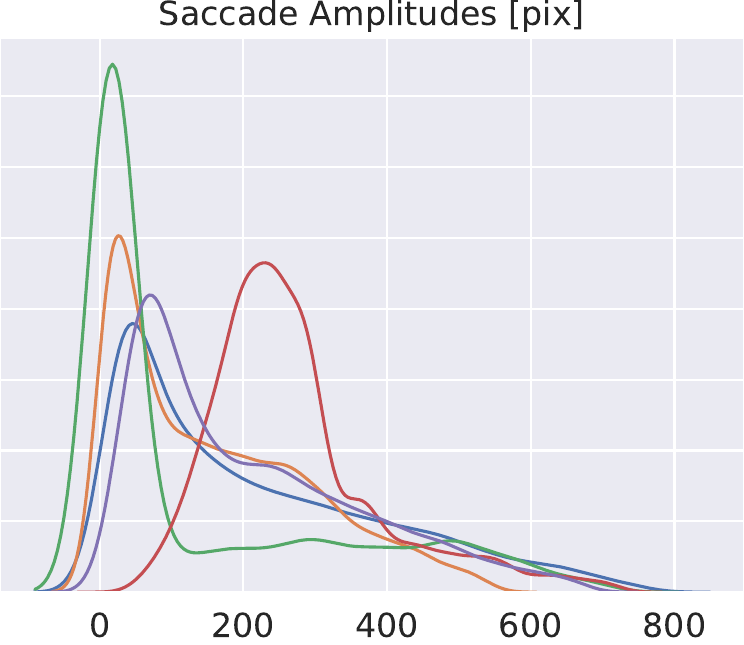}

    \end{tabular}
    }
    \vspace{-0.25cm}
    \caption{Statistical properties exhibited by \tppgaze~and other methods relative to those of human observers, in terms of empirical fixation durations and saccade amplitudes on MIT1003 (top row), NUSEF (middle row) and FiFa (bottom row) datasets. For consistency with the main paper, comparison against DeepGazeIII on MIT1003 is omitted.}
    \label{fig:additionaldistributions}
    \vspace{-0.3cm}
\end{figure}

\begin{figure}[t]
    \centering
    \footnotesize
    \resizebox{\linewidth}{!}{
    \setlength{\tabcolsep}{.25em}
    \begin{tabular}{cc}
    \includegraphics[height=0.49\linewidth]{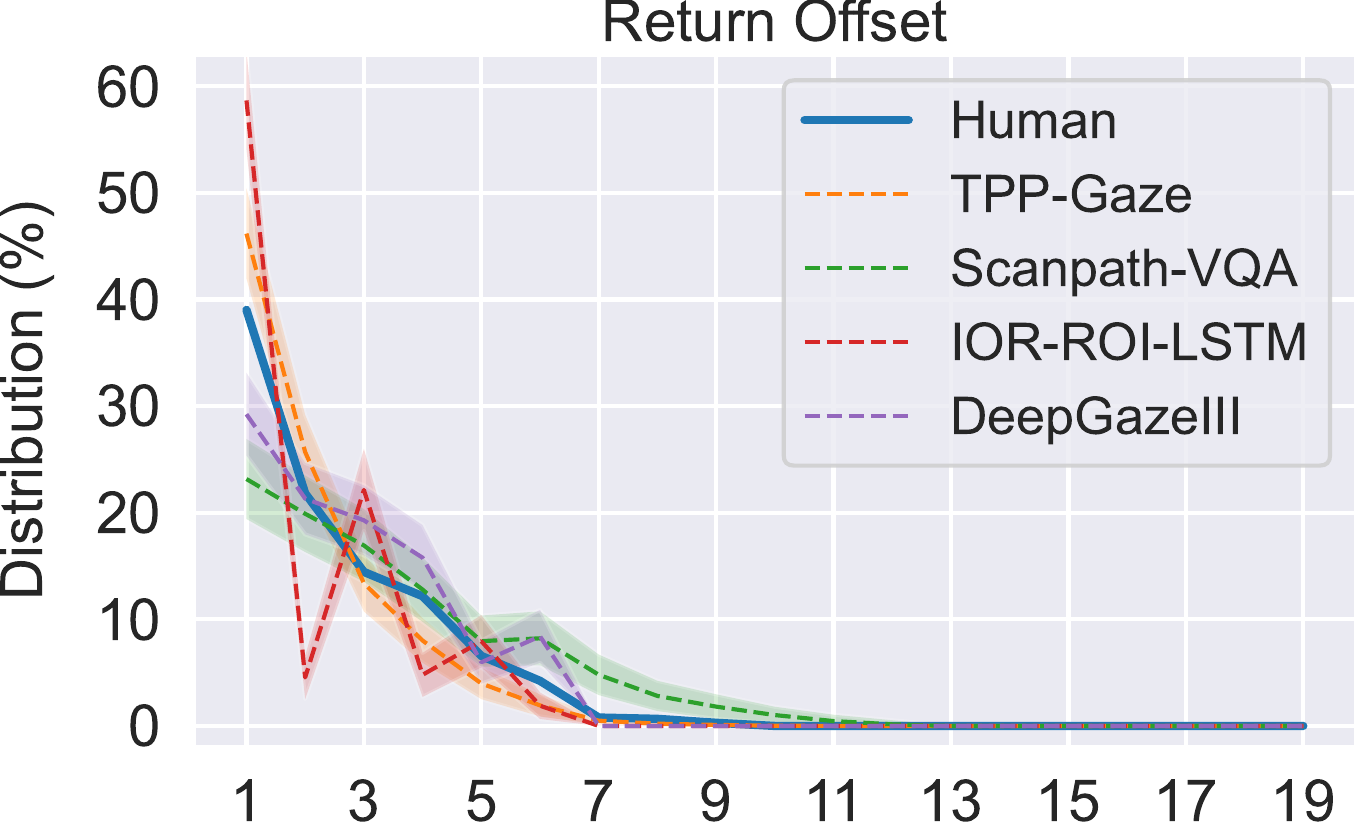} &
    \includegraphics[height=0.49\linewidth]{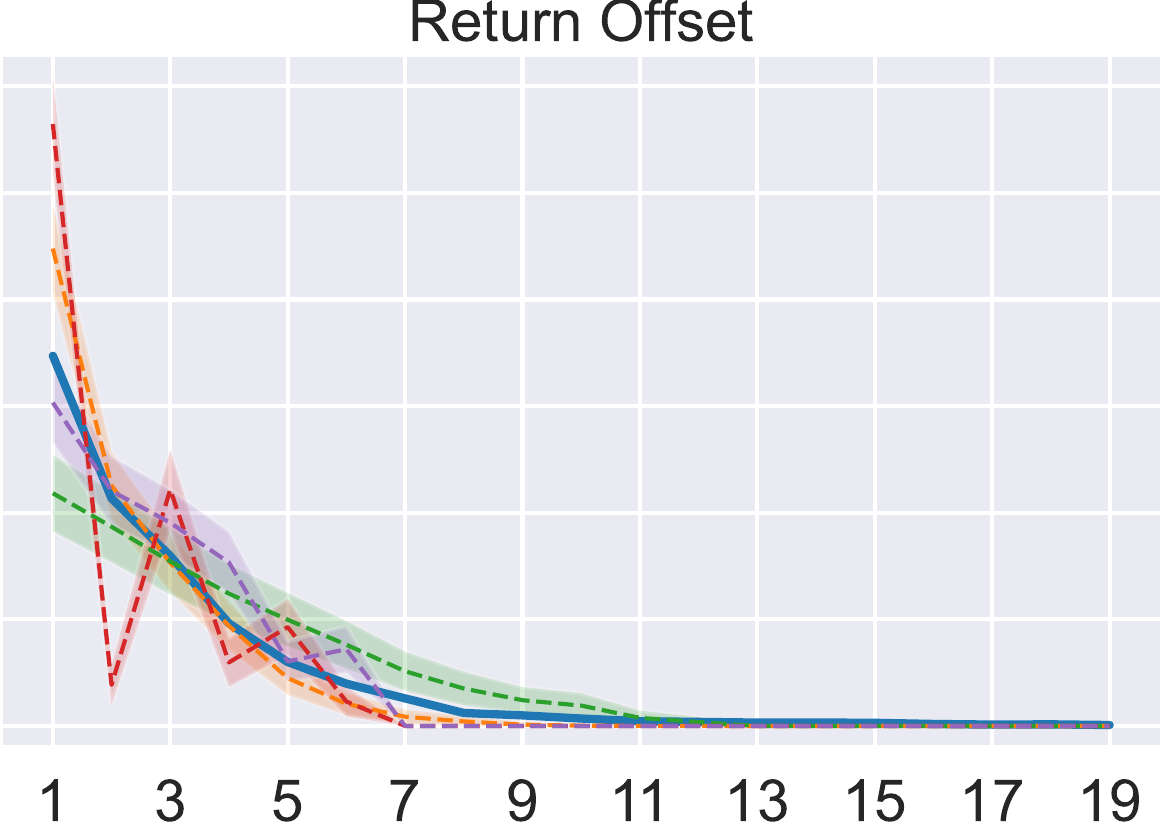} \\
    \addlinespace[0.05cm]
    \multicolumn{2}{c}{\includegraphics[height=0.49\linewidth]{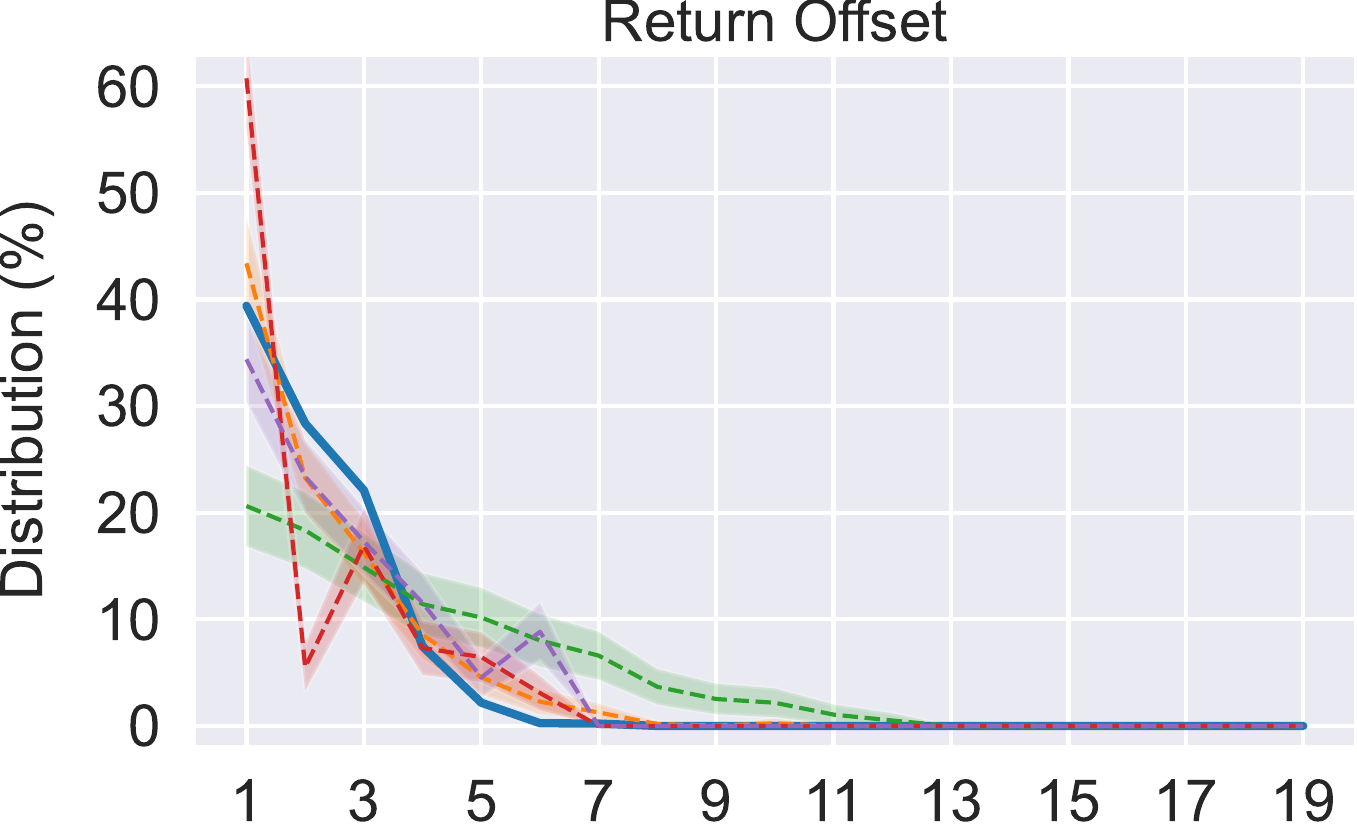}} \\
    \end{tabular}
    }
    \vspace{-0.25cm}
    \caption{Return fixations analysis comparing \tppgaze~with other methods and human observers. Results are shown on MIT1003 (top-left plot), NUSEF (top-right plot), and FiFa (bottom plot) datasets.}
    \label{fig:return_fixations_supp}
    \vspace{-0.5cm}
\end{figure}

\section{Extending the Model to Visual Search}

\begin{figure}
    \centering
    \includegraphics[width=.99\linewidth]{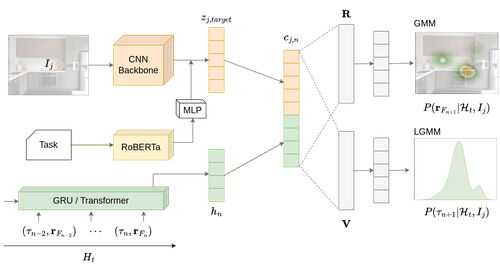}
    \vspace{-0.25cm}
    \caption{Overview of \tppgaze~model architecture extended to handle the visual search task. A linguistic embedding (RoBERTa) of the search target is employed to learn a task-drive semantic representation ($z_j$). The latter, together with the history of past events ($h_n$), is used to simulated the next fixation position and duration.}
    \label{fig:architecture_vs}
    \vspace{-0.5cm}
\end{figure}

We extend the \tppgaze~architecture to handle the visual search task by forcing the model to learn a task-specific semantic representation of the input image (see \cref{fig:architecture_vs}). Recall that the \tppgaze's semantic representation module consists of a DenseNet201 CNN backbone and a learnable readout network composed of three $1\times1$ convolutional layers with 8, 16, and 1 channels, respectively. The obtained spatial priority map is then projected to a fixed-dimensional vector, $\mathbf{z}_j$, to obtain the $j$-th image semantic representation. Specifically, the last layer performing a $1\times1$ convolution is responsible for learning a (non-linear) combination of the feature maps from the previous layers.

To guide the model toward a specific search objective, we redefine the architecture to enable \tppgaze~to learn such a combination conditioned on a given text string. To this end, we first obtain a linguistic embedding of the search target using the RoBERTa language model. Let $\mathbf{F}_{target}$ be the embedding vector representing the search objective. The readout network for the visual search model consists of three $1\times1$ convolutional layers with 16, 64, and 256 channels, respectively. Thus, it is modified to output $M=256$ feature maps. Let $\mathbf{X} = [\mathbf{x}_0; \cdots; \mathbf{x}_M] \in \mathbb{R}^{M\times d}$ represent the matrix of flattened image features. The task-specific semantic representation for the $j$-th image, $\mathbf{z}_{j,target}$, is then obtained as follows:
\begin{equation}
    \begin{aligned}&w=\mathrm{softplus}(\mathrm{MLP}(\mathbf{F}_{target}))\\&
    \mathbf{z}_{j,target}=\sum_{i=1}^{M}w_{i}\mathbf{x}_{i}.\end{aligned}
\end{equation}
\section{Additional Qualitative Results}
\begin{figure*}[t]
    \footnotesize
    \setlength{\tabcolsep}{.1em}
    \resizebox{\linewidth}{!}{
    \begin{tabular}{ccc}
         \tiny G-Eymol~\cite{zanca2019gravitational} & \tiny IOR-ROI-LSTM~\cite{chen2018scanpath} & \tiny DeepGazeIII~\cite{kummerer2022deepgaze} \\
         \addlinespace[0.08cm]
         \includegraphics[width=0.16\linewidth]{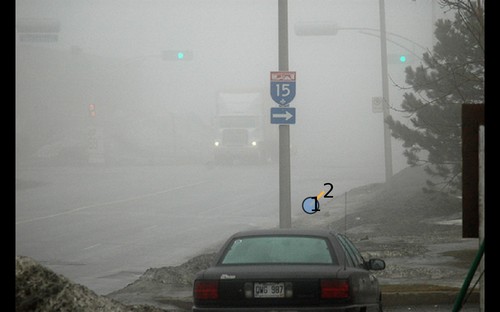} &
         \includegraphics[width=0.16\linewidth]{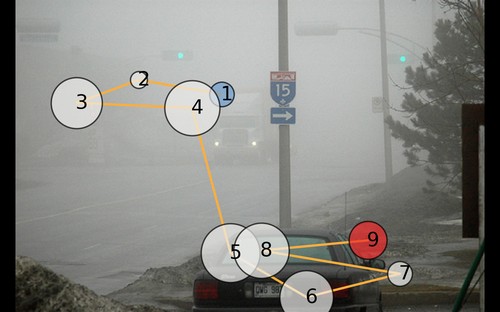} &
         \includegraphics[width=0.16\linewidth]{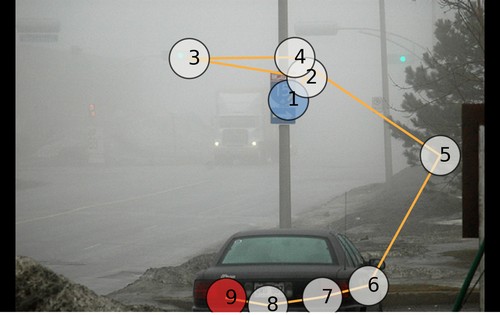} \\ 
         
         \tiny Scanpath-VQA~\cite{chen2021predicting} & \tiny \textbf{\ours (Ours)} & \tiny Humans \\
         \includegraphics[width=0.16\linewidth]{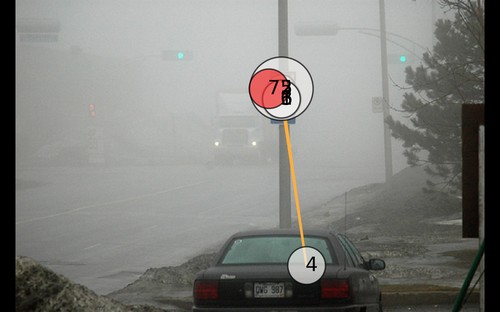} & 
         \includegraphics[width=0.16\linewidth]{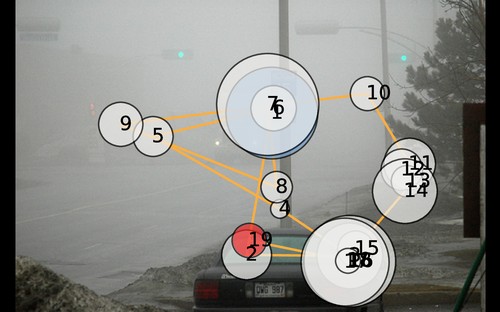} & 
         \includegraphics[width=0.16\linewidth]{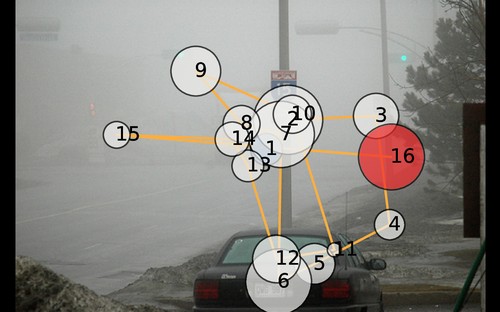} \\

         \addlinespace[0.6cm]
         
         \tiny G-Eymol~\cite{zanca2019gravitational} & \tiny IOR-ROI-LSTM~\cite{chen2018scanpath} & \tiny DeepGazeIII~\cite{kummerer2022deepgaze} \\
         \addlinespace[0.08cm]
         \includegraphics[width=0.16\linewidth]{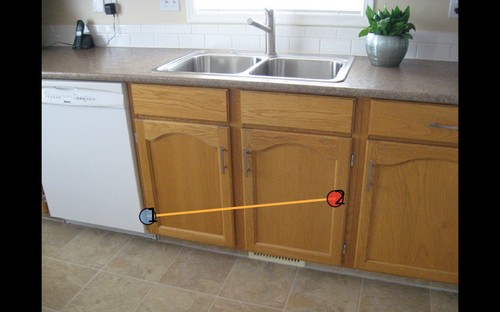} &
         \includegraphics[width=0.16\linewidth]{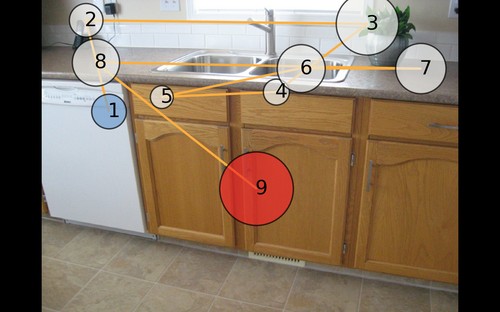} &
         \includegraphics[width=0.16\linewidth]{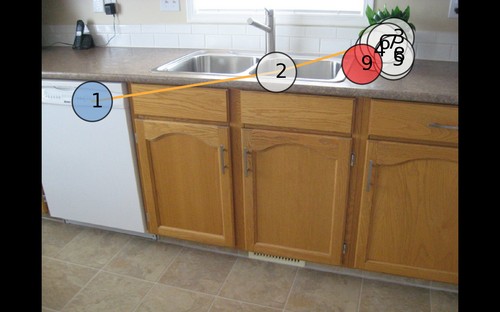} \\ 
         
          \tiny Scanpath-VQA~\cite{chen2021predicting} & \tiny \textbf{\ours (Ours)} & \tiny  \\
         \includegraphics[width=0.16\linewidth]{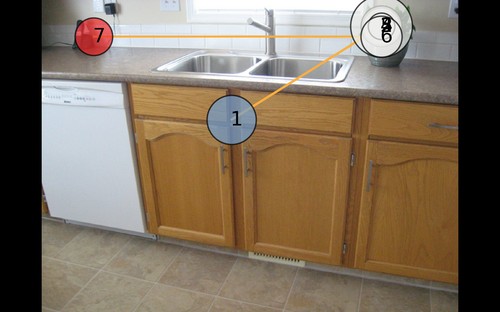} & 
         \includegraphics[width=0.16\linewidth]{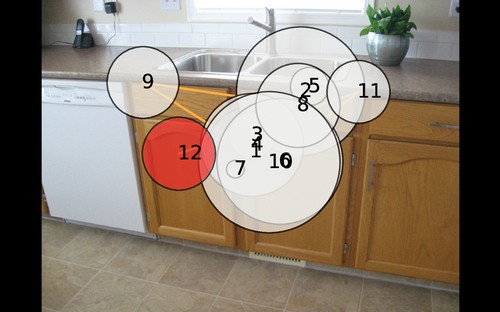} & 
         \includegraphics[width=0.16\linewidth]{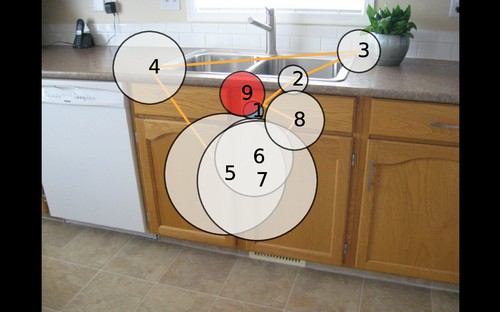}
    \end{tabular}
    }
    \vspace{-0.15cm}
    \caption{Qualitative comparison of simulated and human scanpaths on the COCO-FreeView dataset.}
    \label{fig:qualitatives_coco}
    \vspace{-0.4cm}
\end{figure*}

\begin{figure*}[t]
    \footnotesize
    \setlength{\tabcolsep}{.1em}
    \resizebox{\linewidth}{!}{
    \begin{tabular}{ccc}
         \tiny G-Eymol~\cite{zanca2019gravitational} & \tiny IOR-ROI-LSTM~\cite{chen2018scanpath} & \tiny Scanpath-VQA~\cite{chen2021predicting} \\
         \addlinespace[0.08cm]
         \includegraphics[width=0.16\linewidth]{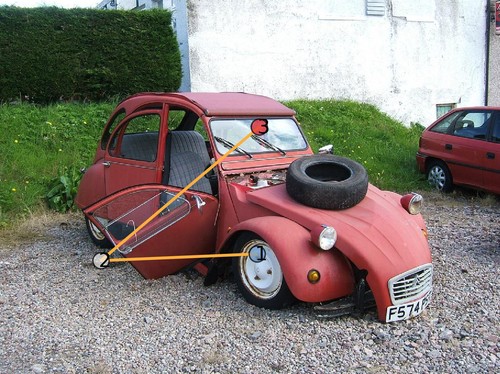} &
         \includegraphics[width=0.16\linewidth]{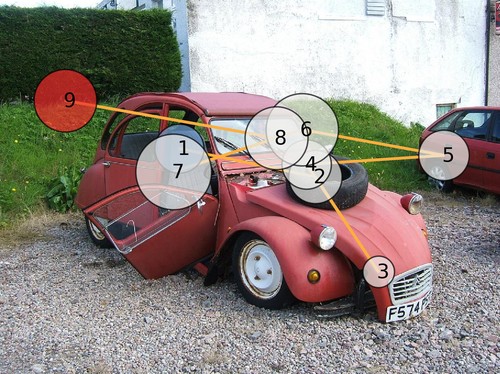} &
         \includegraphics[width=0.16\linewidth]{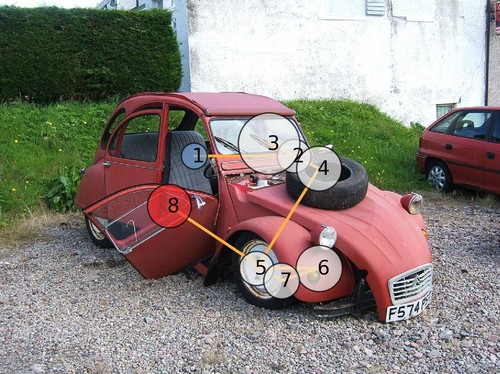} \\ 
         
         \tiny \textbf{\ours (Ours)} & \tiny Humans \\
         \includegraphics[width=0.16\linewidth]{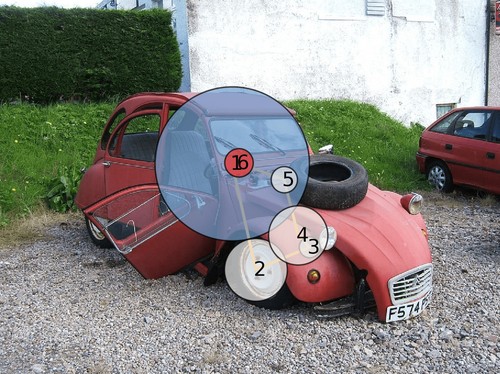} & 
         \includegraphics[width=0.16\linewidth]{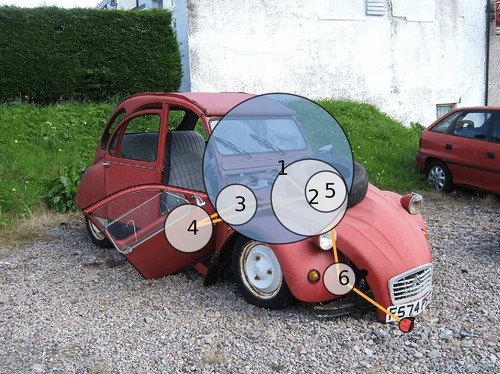} \\

         \addlinespace[0.4cm]
         
         \tiny G-Eymol~\cite{zanca2019gravitational} & \tiny IOR-ROI-LSTM~\cite{chen2018scanpath} & \tiny Scanpath-VQA~\cite{chen2021predicting}\\
         \addlinespace[0.08cm]
         \includegraphics[width=0.16\linewidth]{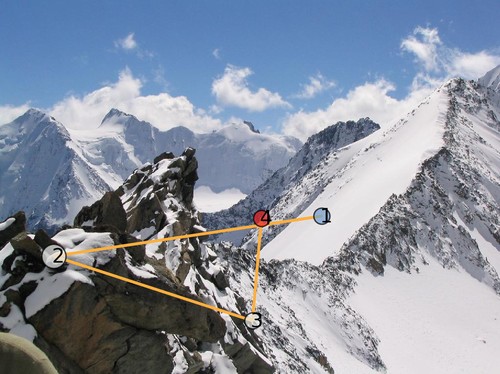} &
         \includegraphics[width=0.16\linewidth]{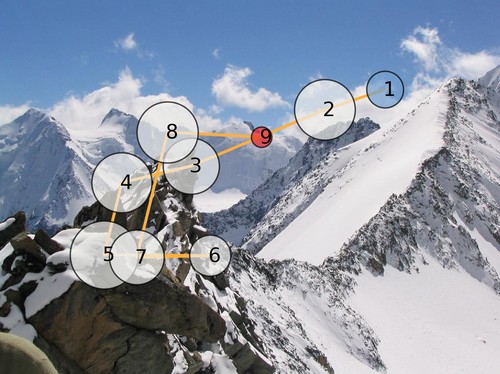} & 
         \includegraphics[width=0.16\linewidth]{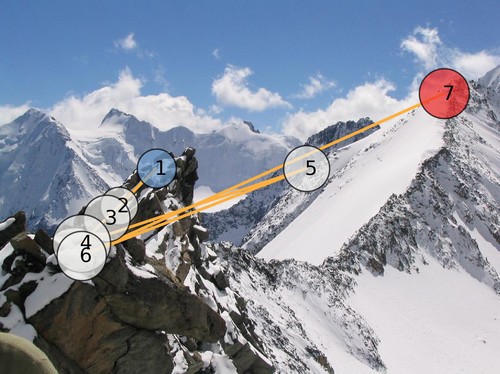} \\ 
         
         \tiny \textbf{\ours (Ours)} & \tiny Humans \\
         \includegraphics[width=0.16\linewidth]{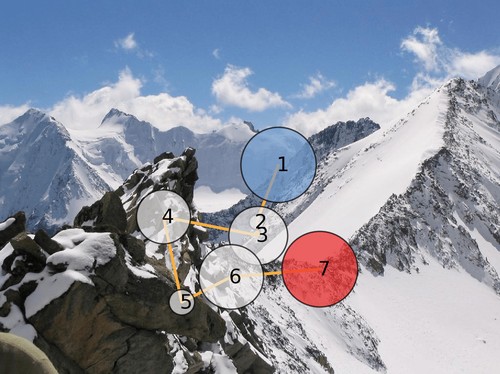} & 
         \includegraphics[width=0.16\linewidth]{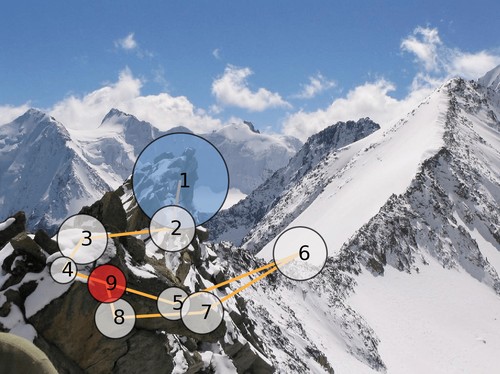}
    \end{tabular}
    }
    \vspace{-0.15cm}
    \caption{Qualitative comparison of simulated and human scanpaths on the MIT1003 dataset. We omit DeepGazeIII for consistency with the experimental settings described in the main paper.}
    \label{fig:qualitatives_mit1003}
    \vspace{-0.4cm}
\end{figure*}

\begin{figure*}[t]
    \footnotesize
    \setlength{\tabcolsep}{.1em}
    \resizebox{\linewidth}{!}{
    \begin{tabular}{ccc}
         \tiny G-Eymol~\cite{zanca2019gravitational} & \tiny IOR-ROI-LSTM~\cite{chen2018scanpath} & \tiny DeepGazeIII~\cite{kummerer2022deepgaze} \\
         \addlinespace[0.08cm]
         \includegraphics[width=0.16\linewidth]{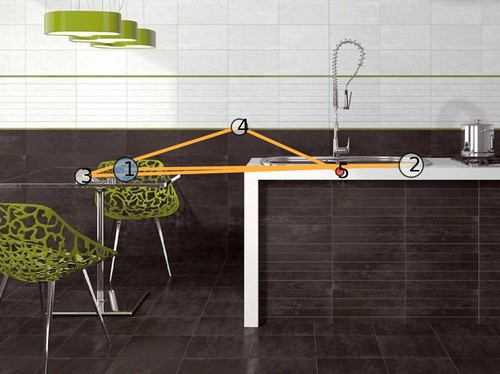} &
         \includegraphics[width=0.16\linewidth]{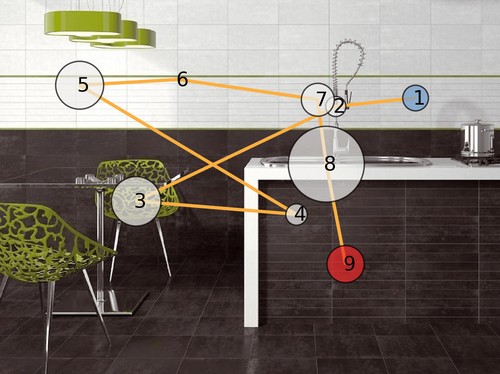} &
         \includegraphics[width=0.16\linewidth]{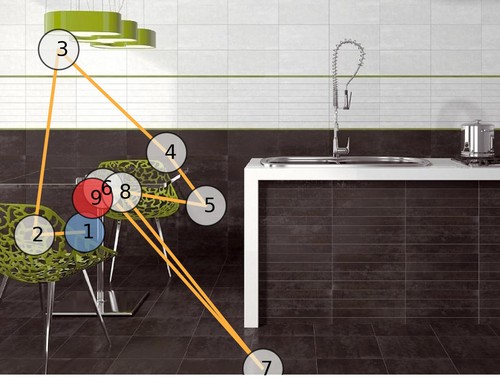} \\ 
         
         \tiny Scanpath-VQA~\cite{chen2021predicting} & \tiny \textbf{\ours (Ours)} & \tiny Humans \\
         \includegraphics[width=0.16\linewidth]{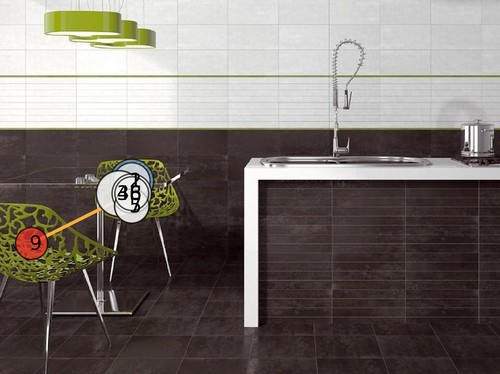} & 
         \includegraphics[width=0.16\linewidth]{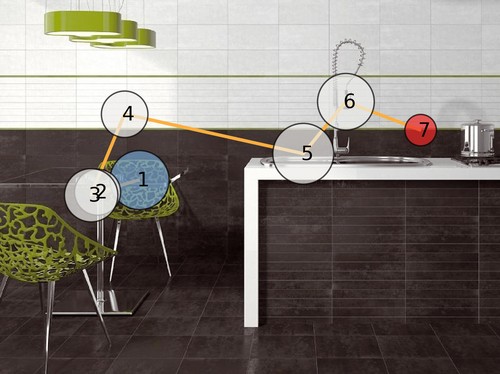} & 
         \includegraphics[width=0.16\linewidth]{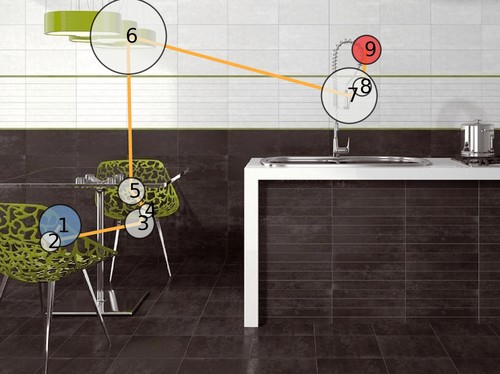} \\

         \addlinespace[0.6cm]
         
         \tiny G-Eymol~\cite{zanca2019gravitational} & \tiny IOR-ROI-LSTM~\cite{chen2018scanpath} & \tiny DeepGazeIII~\cite{kummerer2022deepgaze} \\
         \addlinespace[0.08cm]
         \includegraphics[width=0.16\linewidth]{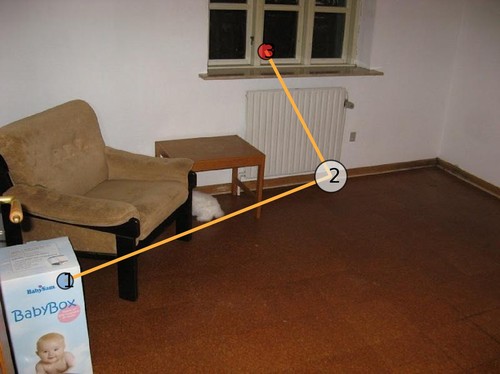} &
         \includegraphics[width=0.16\linewidth]{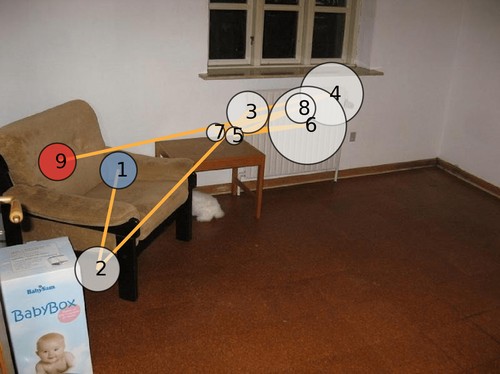} &
         \includegraphics[width=0.16\linewidth]{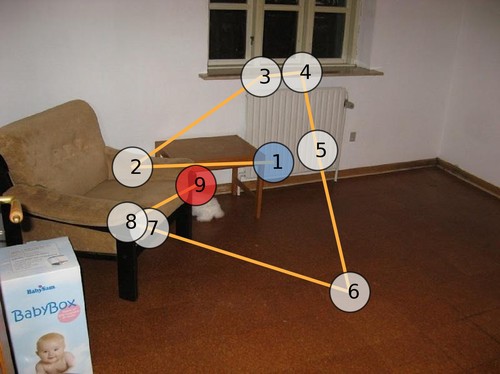} \\ 
         
         \tiny Scanpath-VQA~\cite{chen2021predicting} & \tiny \textbf{\ours (Ours)} & \tiny Humans \\
         \includegraphics[width=0.16\linewidth]{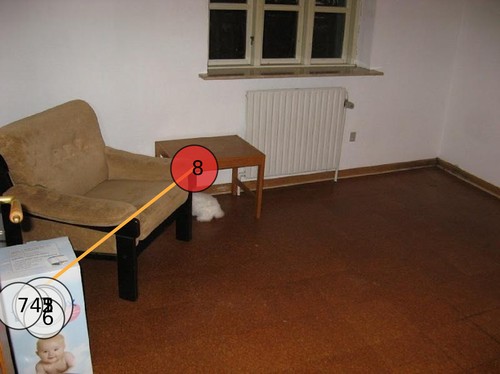} & 
         \includegraphics[width=0.16\linewidth]{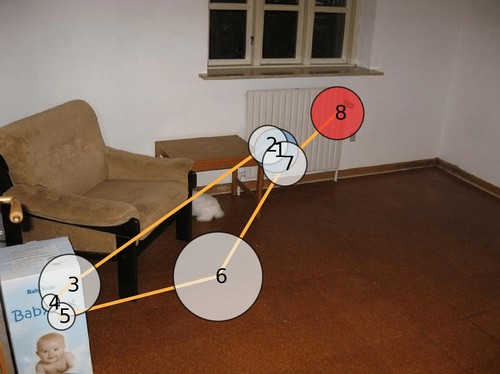} & 
         \includegraphics[width=0.16\linewidth]{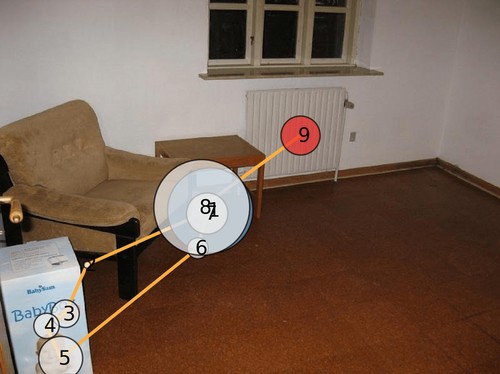}
    \end{tabular}
    }
    \vspace{-0.15cm}
    \caption{Qualitative comparison of simulated and human scanpaths on the OSIE dataset.}
    \label{fig:qualitatives_osie}
    \vspace{-0.4cm}
\end{figure*}

\begin{figure*}[t]
    \footnotesize
    \setlength{\tabcolsep}{.1em}
    \resizebox{\linewidth}{!}{
    \begin{tabular}{ccc}
         \tiny G-Eymol~\cite{zanca2019gravitational} & \tiny IOR-ROI-LSTM~\cite{chen2018scanpath} & \tiny DeepGazeIII~\cite{kummerer2022deepgaze} \\
         \addlinespace[0.08cm]
         \includegraphics[width=0.16\linewidth]{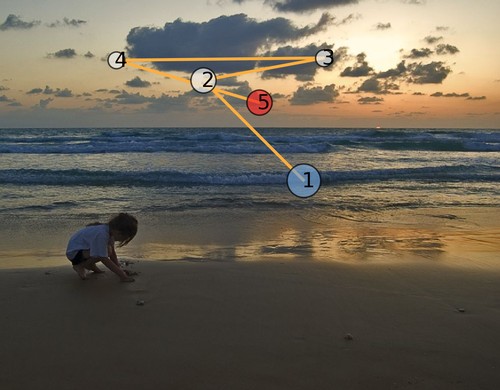} &
         \includegraphics[width=0.16\linewidth]{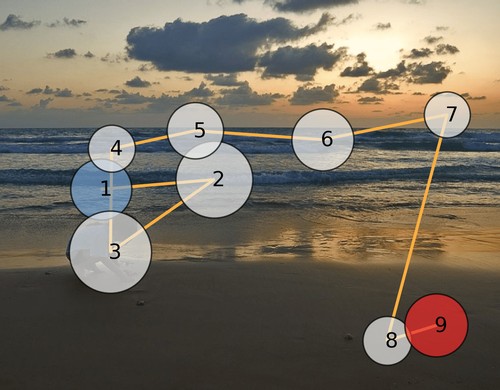} &
         \includegraphics[width=0.16\linewidth]{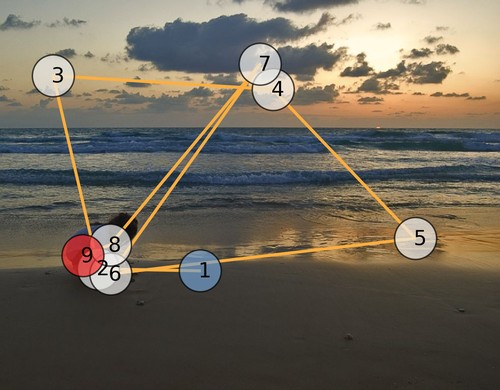} \\ 
         
         \tiny Scanpath-VQA~\cite{chen2021predicting} & \tiny \textbf{\ours (Ours)} & \tiny Humans \\
         \includegraphics[width=0.16\linewidth]{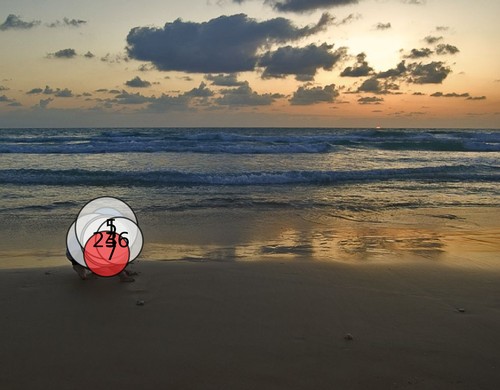} & 
         \includegraphics[width=0.16\linewidth]{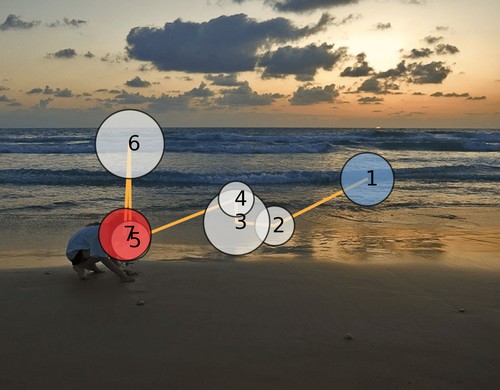} & 
         \includegraphics[width=0.16\linewidth]{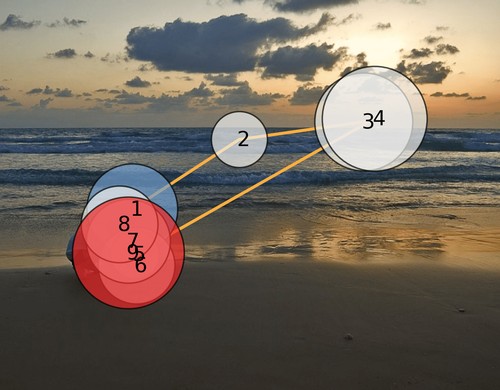} \\

         \addlinespace[0.5cm]
         
         \tiny G-Eymol~\cite{zanca2019gravitational} & \tiny IOR-ROI-LSTM~\cite{chen2018scanpath} & \tiny DeepGazeIII~\cite{kummerer2022deepgaze} \\
         \addlinespace[0.08cm]
         \includegraphics[width=0.16\linewidth]{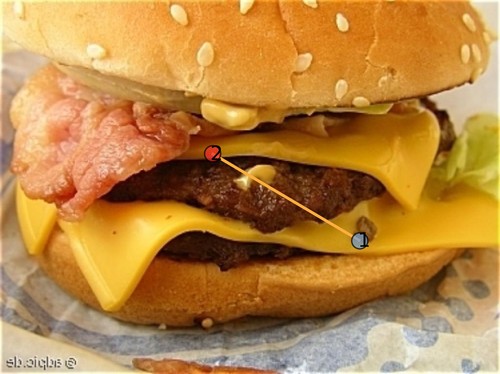} &
         \includegraphics[width=0.16\linewidth]{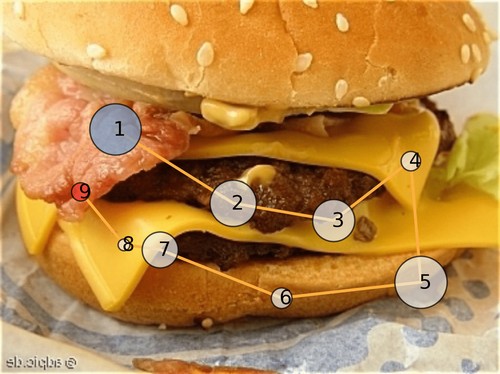} &
         \includegraphics[width=0.16\linewidth]{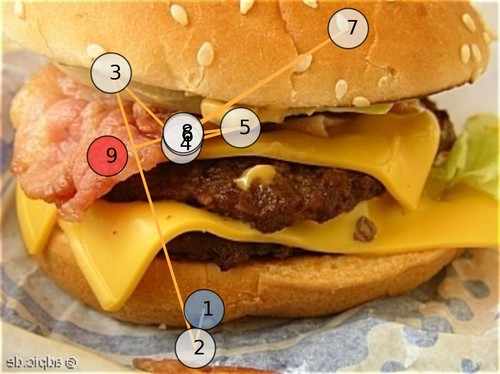} \\ 
         
         \tiny Scanpath-VQA~\cite{chen2021predicting} & \tiny \textbf{\ours (Ours)} & \tiny Humans \\
         \includegraphics[width=0.16\linewidth]{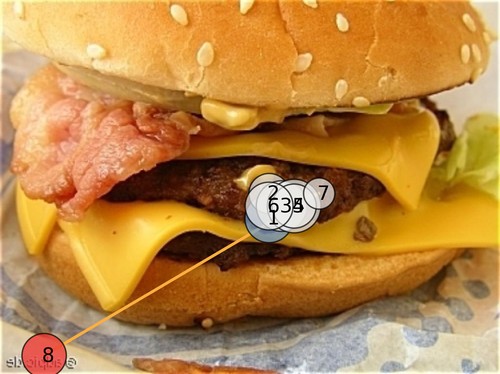} & 
         \includegraphics[width=0.16\linewidth]{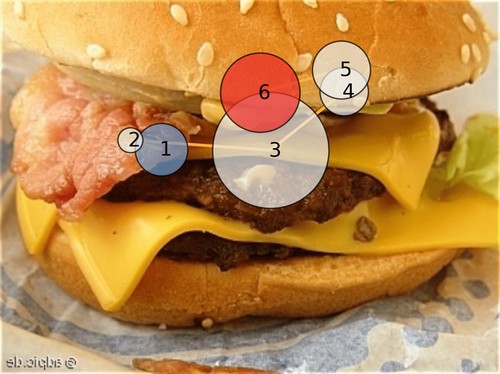} & 
         \includegraphics[width=0.16\linewidth]{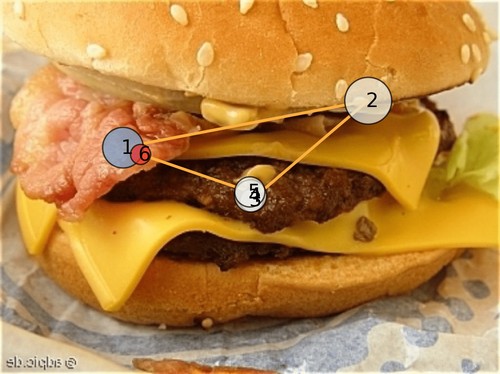}
    \end{tabular}
    }
    \vspace{-0.15cm}
    \caption{Qualitative comparison of simulated and human scanpaths on the NUSEF dataset.}
    \label{fig:qualitatives_nusef}
    \vspace{-0.4cm}
\end{figure*}

\begin{figure*}[t]
    \footnotesize
    \setlength{\tabcolsep}{.1em}
    \resizebox{\linewidth}{!}{
    \begin{tabular}{ccc}
         \tiny G-Eymol~\cite{zanca2019gravitational} & \tiny IOR-ROI-LSTM~\cite{chen2018scanpath} & \tiny DeepGazeIII~\cite{kummerer2022deepgaze} \\
         \addlinespace[0.08cm]
         \includegraphics[width=0.16\linewidth]{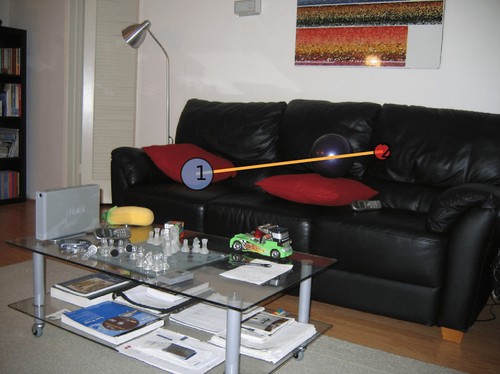} &
         \includegraphics[width=0.16\linewidth]{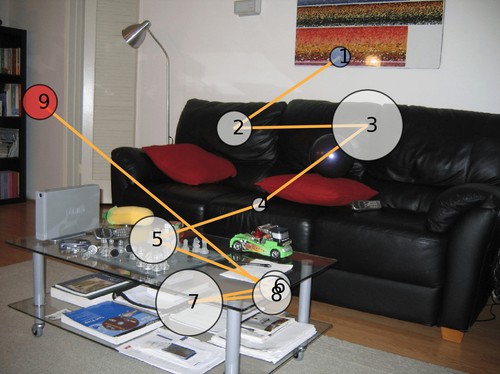} &
         \includegraphics[width=0.16\linewidth]{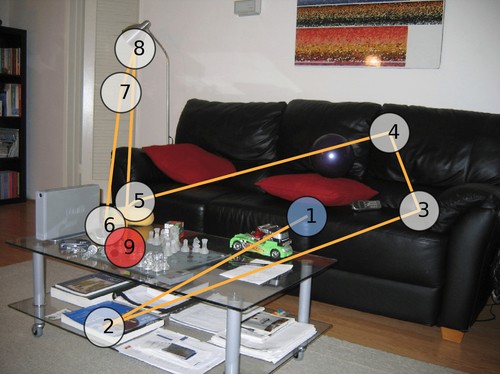} \\ 
         
         \tiny Scanpath-VQA~\cite{chen2021predicting} & \tiny \textbf{\ours (Ours)} & \tiny Humans \\
         \includegraphics[width=0.16\linewidth]{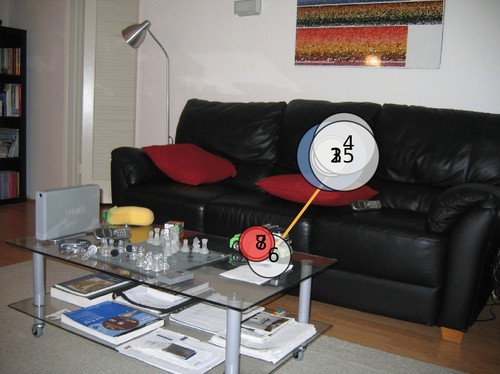} & 
         \includegraphics[width=0.16\linewidth]{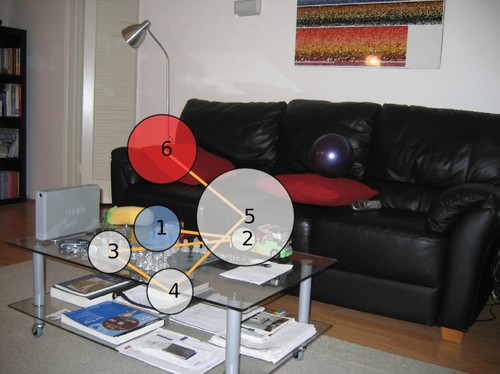} & 
         \includegraphics[width=0.16\linewidth]{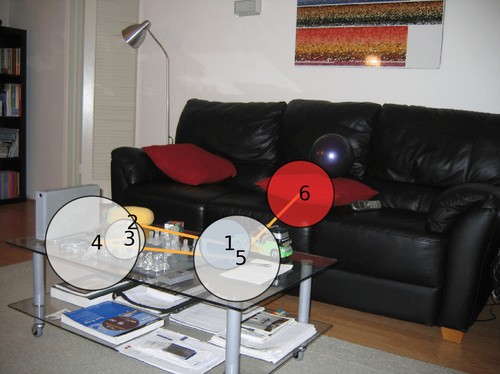} \\

         \addlinespace[0.6cm]
         
         \tiny G-Eymol~\cite{zanca2019gravitational} & \tiny IOR-ROI-LSTM~\cite{chen2018scanpath} & \tiny DeepGazeIII~\cite{kummerer2022deepgaze} \\
         \addlinespace[0.08cm]
         \includegraphics[width=0.16\linewidth]{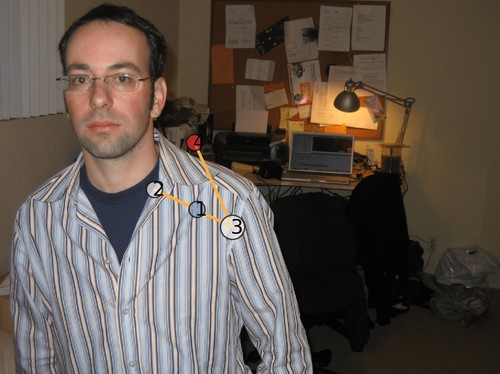} &
         \includegraphics[width=0.16\linewidth]{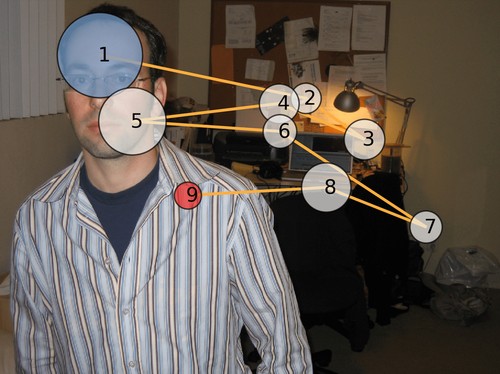} &
         \includegraphics[width=0.16\linewidth]{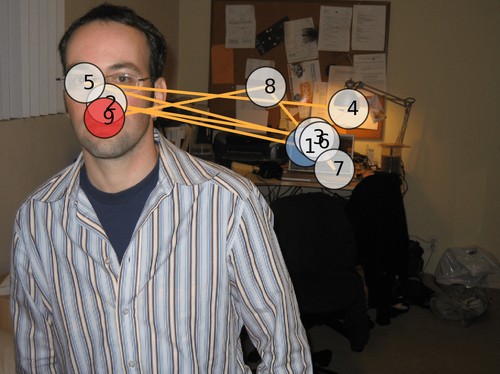} \\ 
         
         \tiny Scanpath-VQA~\cite{chen2021predicting} & \tiny \textbf{\ours (Ours)} & \tiny Humans \\
         \includegraphics[width=0.16\linewidth]{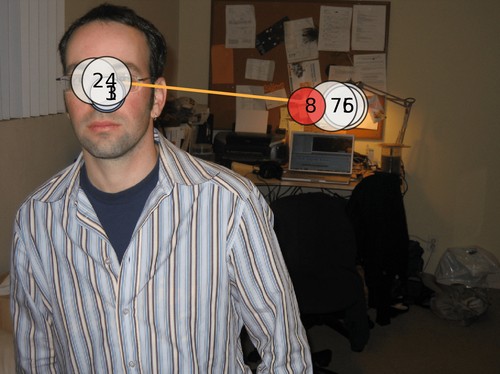} & 
         \includegraphics[width=0.16\linewidth]{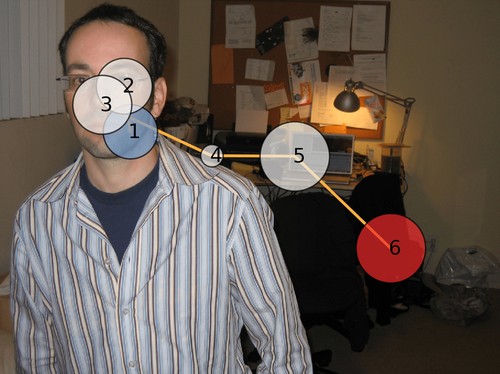} & 
         \includegraphics[width=0.16\linewidth]{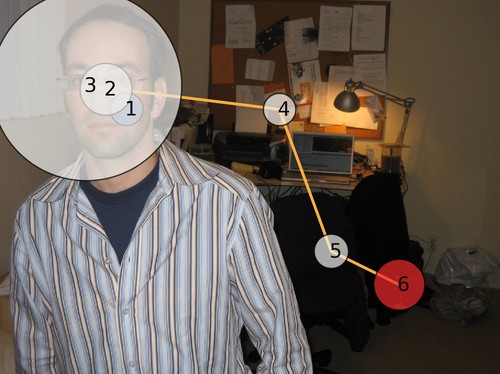}
    \end{tabular}
    }
    \vspace{-0.15cm}
    \caption{Qualitative comparison of simulated and human scanpaths on the FiFa dataset.}
    \label{fig:qualitatives_fifa}
    \vspace{-0.4cm}
\end{figure*}

Additional qualitative results are depicted from \cref{fig:qualitatives_coco} to \cref{fig:qualitatives_fifa} on COCO-FreeView, MIT1003, OSIE, NUSEF, and FiFa datasets, respectively. Each fixation is represented by a circle, with its diameter proportional to the fixation duration. For methods that do not model fixation duration, circles are shown with a uniform size. The first fixation of each scanpath is omitted. The qualitative results support the findings of the main paper, highlighting the accuracy of \tppgaze~in predicting human-like scanpaths. Other methods, instead, either overfit on a few salient objects, especially people and faces in the case of Scanpath-VQA, or predict scanpath trajectories containing fixations on unlikely locations (see the bottom sample in \cref{fig:qualitatives_osie} or the top sample in \cref{fig:qualitatives_nusef}).

\begin{figure*}[t]
\centering
    \footnotesize
    \setlength{\tabcolsep}{.12em}
    \resizebox{0.95\linewidth}{!}{
    \begin{tabular}{cc ccc}
         & & \tiny G-Eymol~\cite{zanca2019gravitational} & \tiny IOR-ROI-LSTM~\cite{chen2018scanpath} & \tiny DeepGazeIII~\cite{kummerer2022deepgaze} \\
         \multirow{3}{*}[1.8em]{\includegraphics[width=0.15\linewidth]{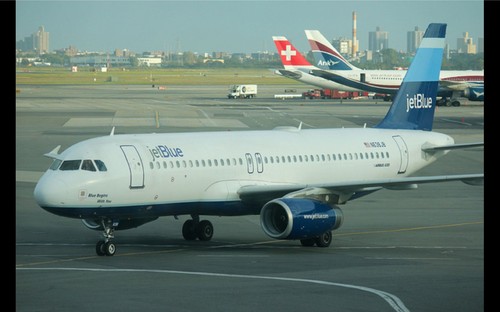}} 
         & & \includegraphics[width=0.14\linewidth]{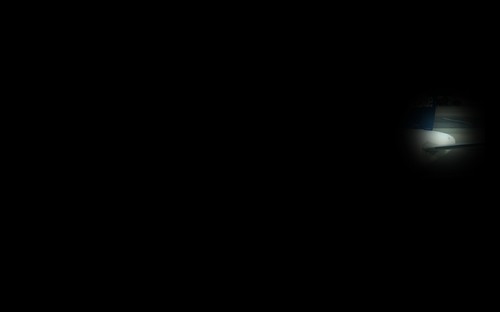}
         & \includegraphics[width=0.14\linewidth]{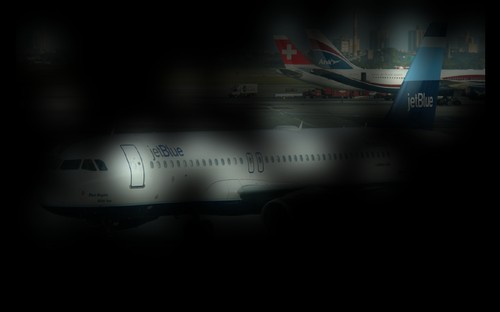}
         & \includegraphics[width=0.14\linewidth]{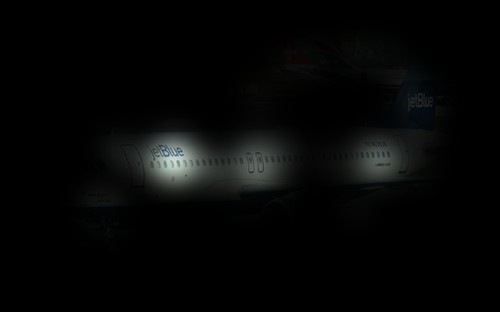} \\ 
         &  & \tiny Scanpath-VQA~\cite{chen2021predicting} & \tiny \textbf{\ours (Ours)}& \tiny Humans \\
         & & \includegraphics[width=0.14\linewidth]{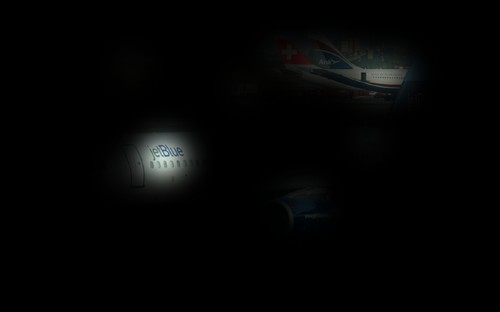}
         & \includegraphics[width=0.14\linewidth]{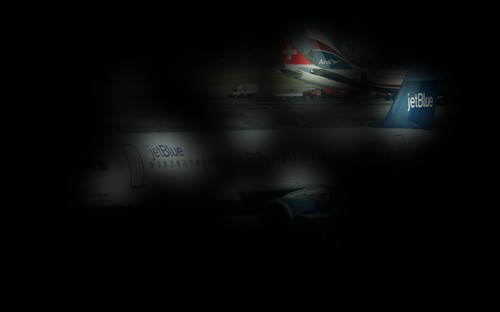}
         & \includegraphics[width=0.14\linewidth]{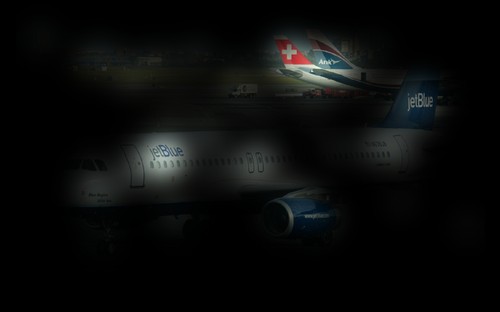} \\
         & & \tiny G-Eymol~\cite{zanca2019gravitational} & \tiny IOR-ROI-LSTM~\cite{chen2018scanpath} & \tiny DeepGazeIII~\cite{kummerer2022deepgaze} \\
         \multirow{3}{*}[1.8em]{\includegraphics[width=0.15\linewidth]{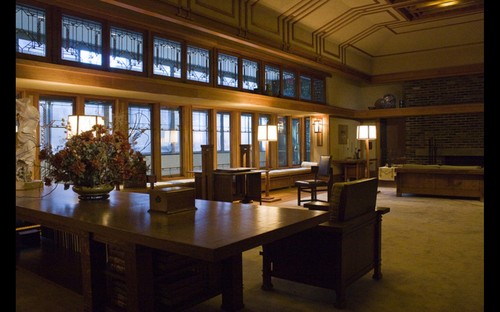}} 
         & & \includegraphics[width=0.14\linewidth]{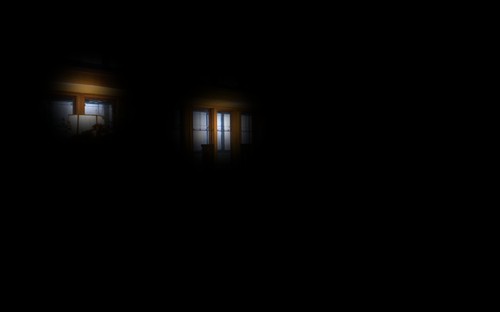}
         & \includegraphics[width=0.14\linewidth]{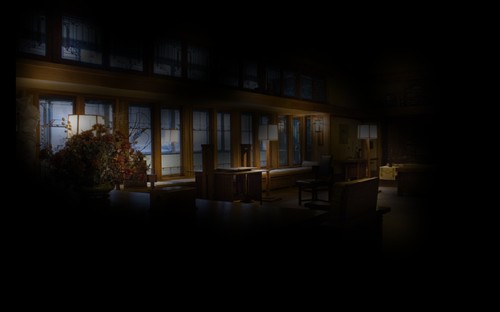}
         & \includegraphics[width=0.14\linewidth]{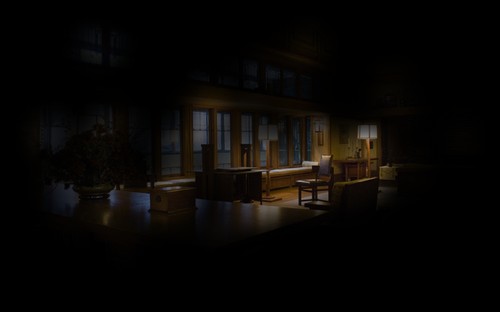} \\ 
         &  & \tiny Scanpath-VQA~\cite{chen2021predicting} & \tiny \textbf{\ours (Ours)}& \tiny Humans \\
         & & \includegraphics[width=0.14\linewidth]{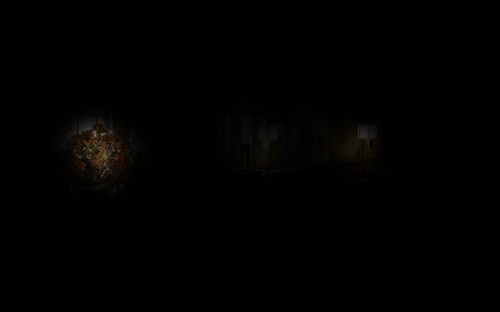}
         & \includegraphics[width=0.14\linewidth]{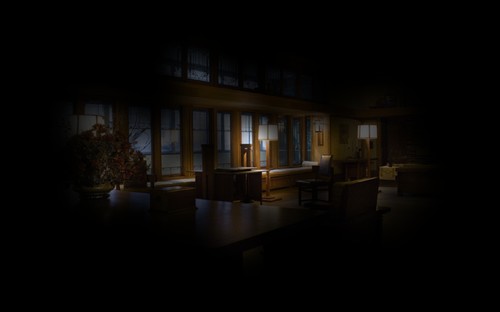}
         & \includegraphics[width=0.14\linewidth]{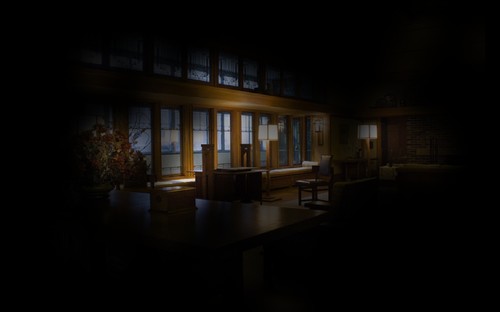} \\
         & & \tiny G-Eymol~\cite{zanca2019gravitational} & \tiny IOR-ROI-LSTM~\cite{chen2018scanpath} & \tiny DeepGazeIII~\cite{kummerer2022deepgaze} \\
         \multirow{3}{*}[1.8em]{\includegraphics[width=0.15\linewidth]{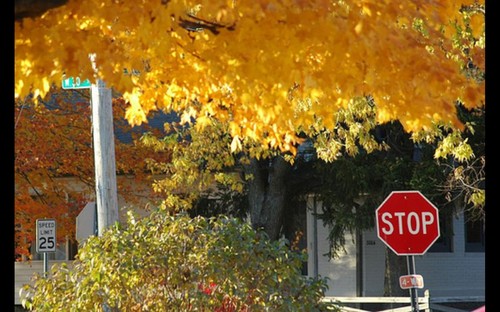}} 
         & & \includegraphics[width=0.14\linewidth]{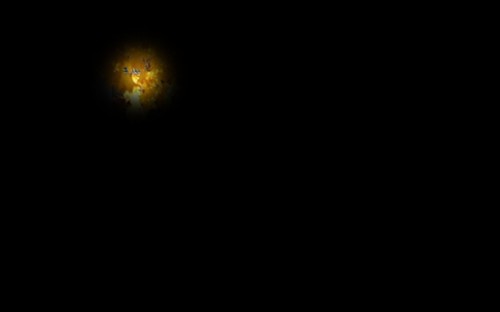}
         & \includegraphics[width=0.14\linewidth]{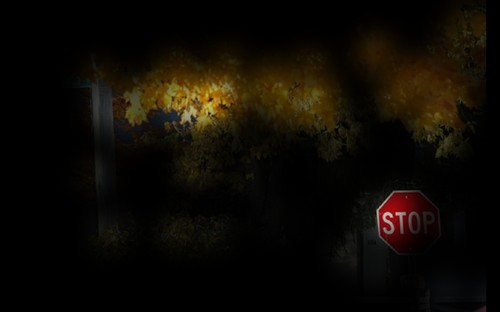}
         & \includegraphics[width=0.14\linewidth]{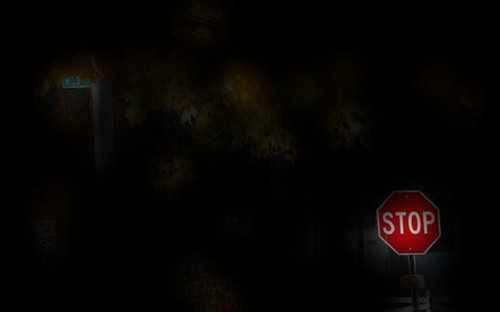} \\ 
         &  & \tiny Scanpath-VQA~\cite{chen2021predicting} & \tiny \textbf{\ours (Ours)}& \tiny Humans \\
         & & \includegraphics[width=0.14\linewidth]{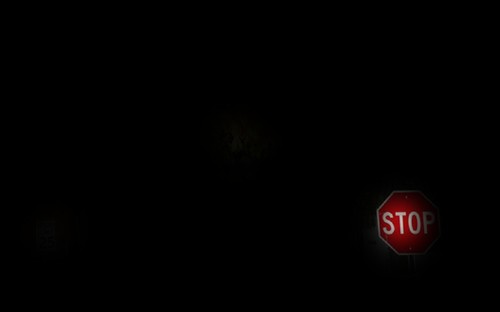}
         & \includegraphics[width=0.14\linewidth]{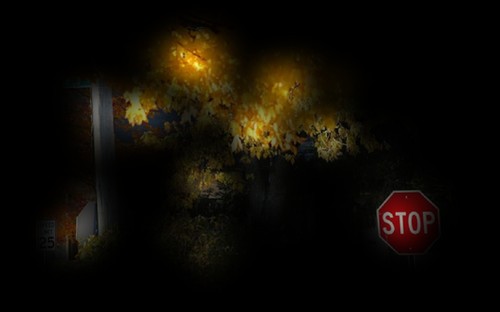}
         & \includegraphics[width=0.14\linewidth]{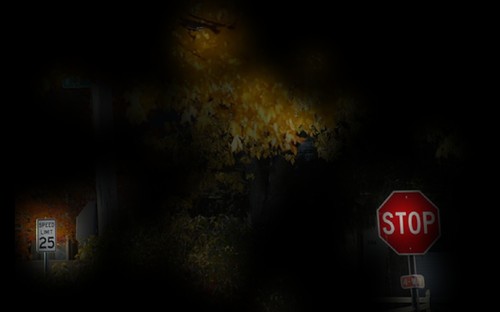} \\
    \end{tabular}
    }
    \vspace{-0.15cm}
    \caption{Saliency maps of sample images from COCO-FreeView dataset computed from the fixations generated by the considered scanpath models. For completeness, we include DeepGazeIII, but note that its training procedure also involves saliency prediction.}
    \label{fig:saliency_cocofreeview}
    \vspace{-0.4cm}
\end{figure*}

\begin{figure*}[t]
\centering
    \footnotesize
    \setlength{\tabcolsep}{.12em}
    \resizebox{0.95\linewidth}{!}{
    \begin{tabular}{cc ccc}
         & & \tiny G-Eymol~\cite{zanca2019gravitational} & \tiny IOR-ROI-LSTM~\cite{chen2018scanpath} & \tiny Scanpath-VQA~\cite{chen2021predicting} \\
         \multirow{3}{*}[1.8em]{\includegraphics[width=0.15\linewidth]{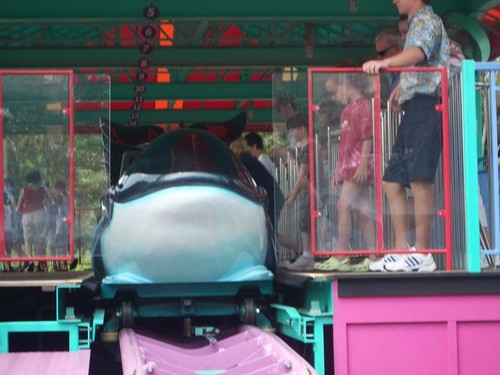}} 
         & & \includegraphics[width=0.14\linewidth]{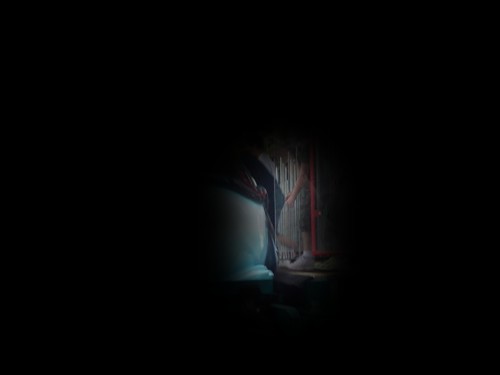}
         & \includegraphics[width=0.14\linewidth]{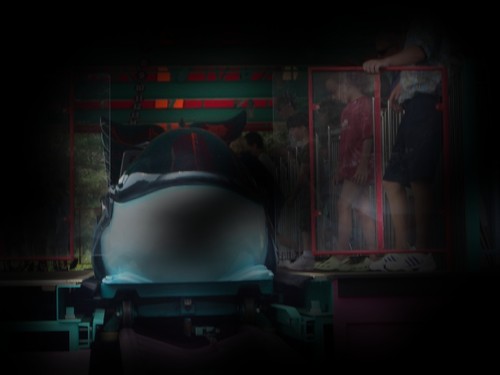}
         & \includegraphics[width=0.14\linewidth]{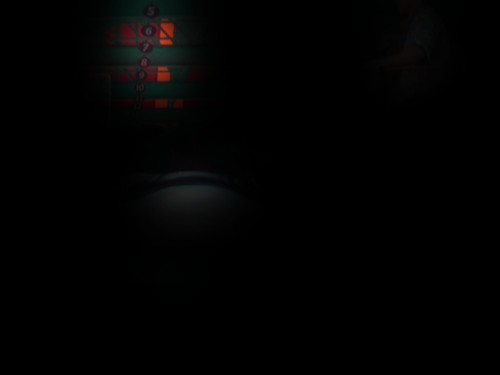} \\ 
         & & \tiny \textbf{\ours (Ours)}& \tiny Humans \\
         & & \includegraphics[width=0.14\linewidth]{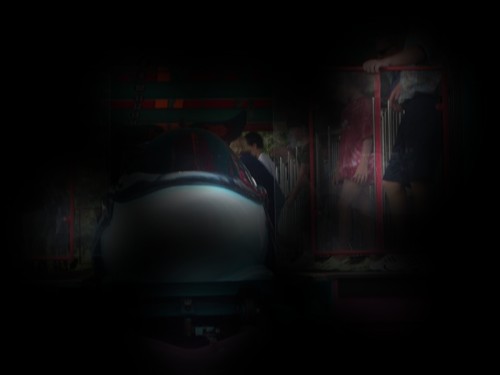}
         & \includegraphics[width=0.14\linewidth]{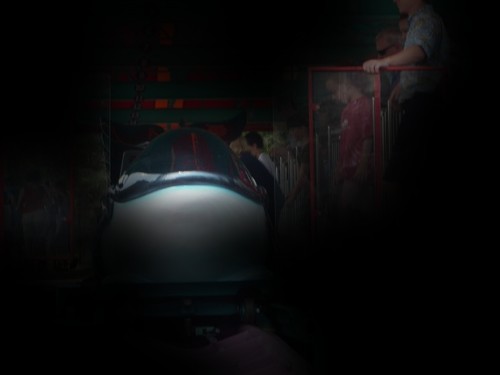}
         & \\
         & & \tiny G-Eymol~\cite{zanca2019gravitational} & \tiny IOR-ROI-LSTM~\cite{chen2018scanpath} & \tiny Scanpath-VQA~\cite{chen2021predicting} \\
         \multirow{3}{*}[1.8em]{\includegraphics[width=0.15\linewidth]{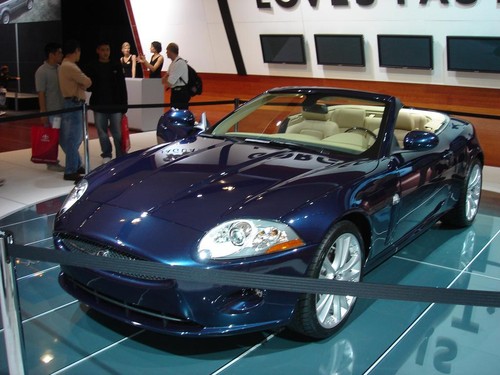}} 
         & & \includegraphics[width=0.14\linewidth]{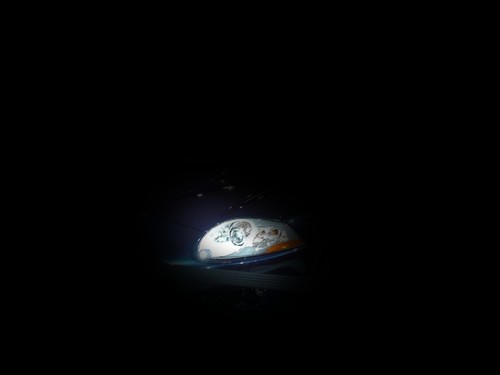}
         & \includegraphics[width=0.14\linewidth]{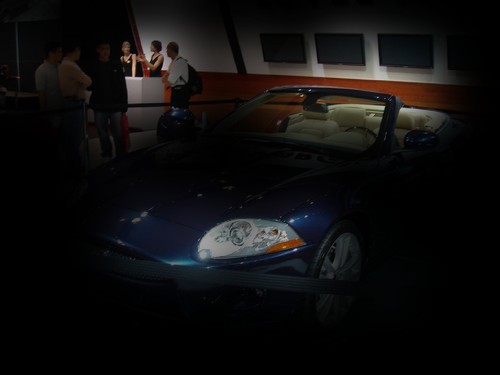}
         & \includegraphics[width=0.14\linewidth]{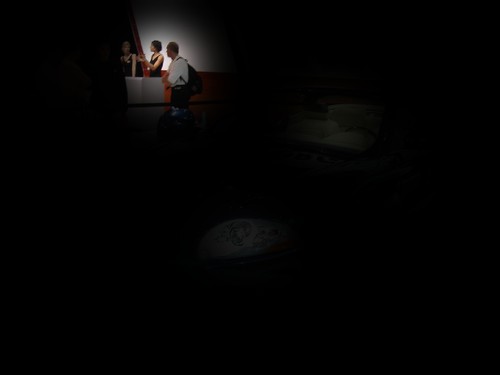} \\ 
         & & \tiny \textbf{\ours (Ours)}& \tiny Humans \\
         & & \includegraphics[width=0.14\linewidth]{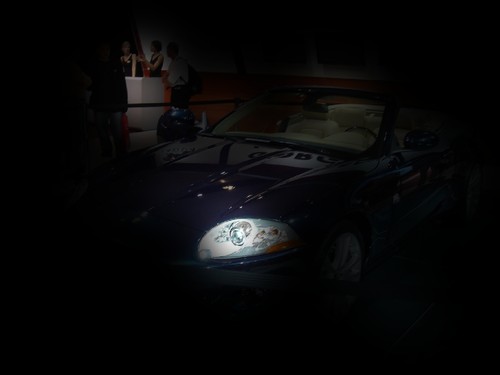}
         & \includegraphics[width=0.14\linewidth]{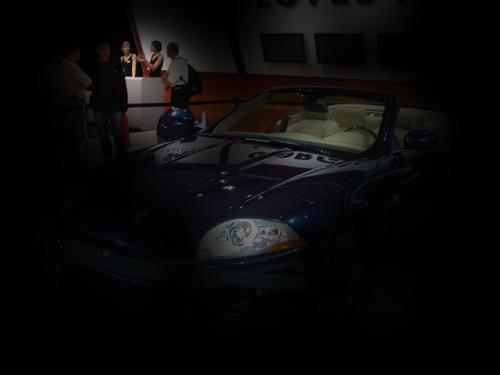} \\
         & & \tiny G-Eymol~\cite{zanca2019gravitational} & \tiny IOR-ROI-LSTM~\cite{chen2018scanpath} & \tiny Scanpath-VQA~\cite{chen2021predicting} \\
         \multirow{3}{*}[1.8em]{\includegraphics[width=0.15\linewidth]{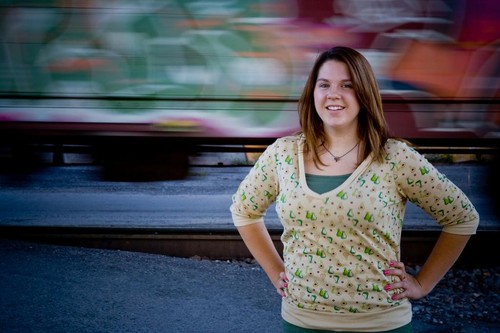}} 
         & & \includegraphics[width=0.14\linewidth]{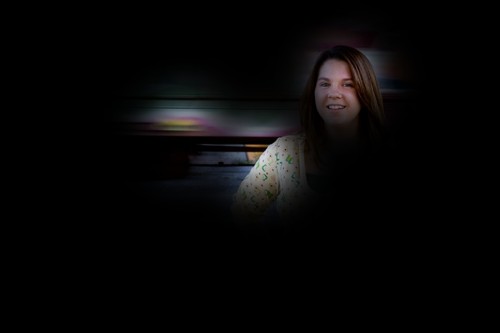}
         & \includegraphics[width=0.14\linewidth]{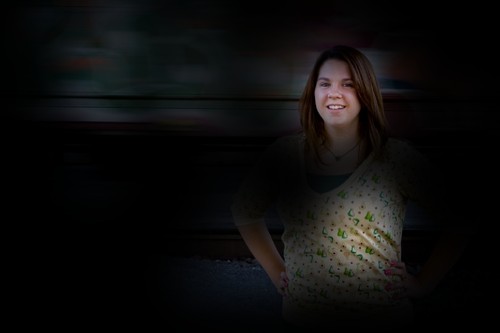}
         & \includegraphics[width=0.14\linewidth]{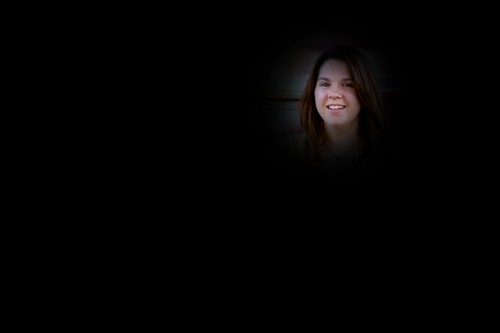} \\ 
         & & \tiny \textbf{\ours (Ours)}& \tiny Humans \\
         & & \includegraphics[width=0.14\linewidth]{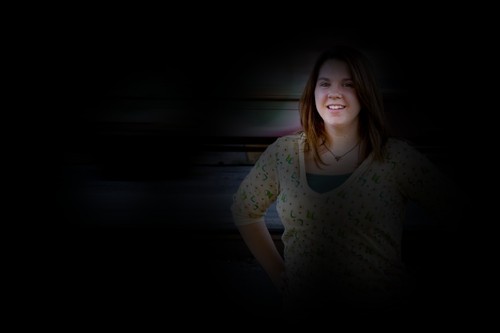}
         & \includegraphics[width=0.14\linewidth]{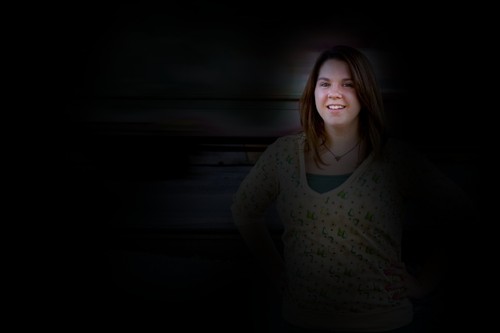}\\
    \end{tabular}
    }
    \vspace{-0.15cm}
    \caption{Saliency maps of sample images from MIT1003 dataset computed from the fixations generated by the considered scanpath models. We omit DeepGazeIII for consistency with the experimental settings described in the main paper.}
    \label{fig:saliency_mit1003}
    \vspace{-0.4cm}
\end{figure*}

\begin{figure*}[t]
\centering
    \footnotesize
    \setlength{\tabcolsep}{.12em}
    \resizebox{0.95\linewidth}{!}{
    \begin{tabular}{cc ccc}
         & & \tiny G-Eymol~\cite{zanca2019gravitational} & \tiny IOR-ROI-LSTM~\cite{chen2018scanpath} & \tiny DeepGazeIII~\cite{kummerer2022deepgaze} \\
         \multirow{3}{*}[1.8em]{\includegraphics[width=0.15\linewidth]{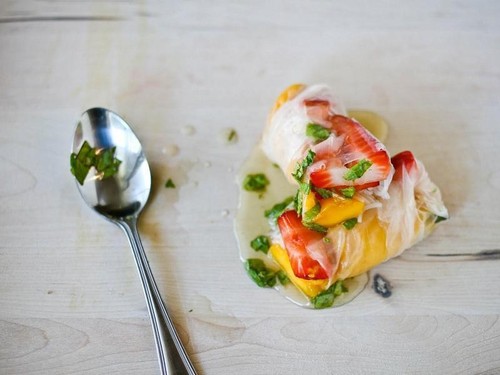}} 
         & & \includegraphics[width=0.14\linewidth]{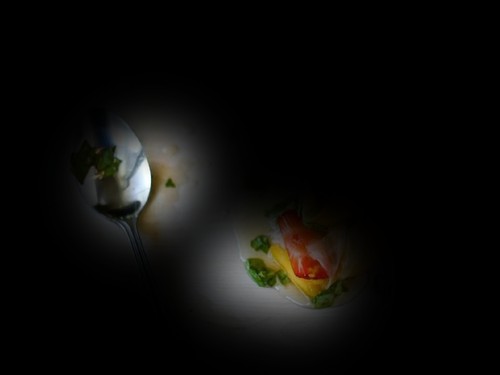}
         & \includegraphics[width=0.14\linewidth]{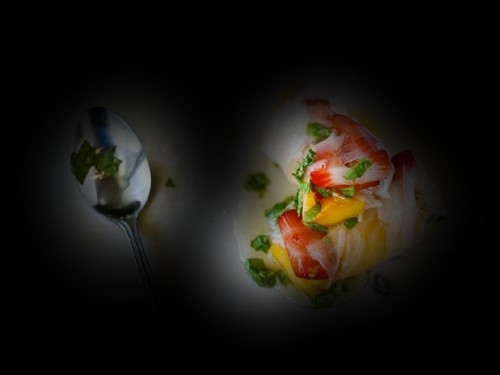}
         & \includegraphics[width=0.14\linewidth]{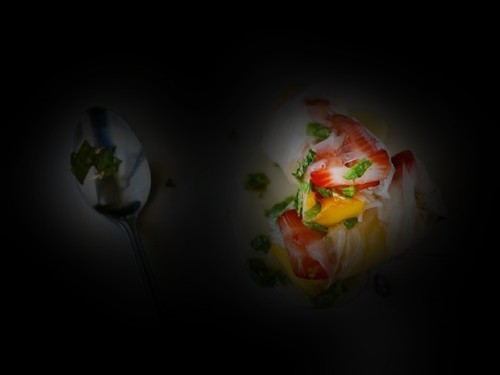} \\ 
         &  & \tiny Scanpath-VQA~\cite{chen2021predicting} & \tiny \textbf{\ours (Ours)}& \tiny Humans \\
         & & \includegraphics[width=0.14\linewidth]{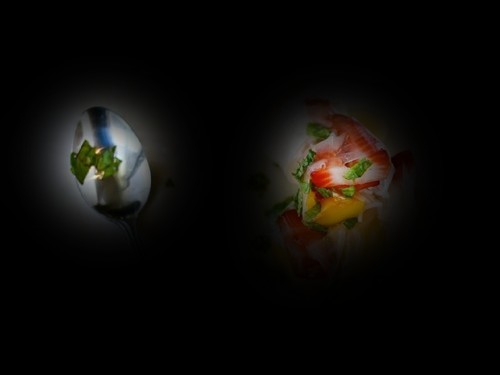}
         & \includegraphics[width=0.14\linewidth]{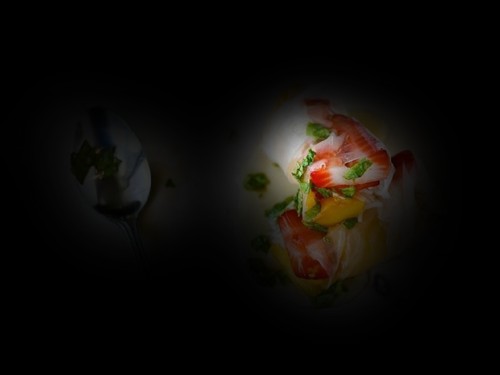}
         & \includegraphics[width=0.14\linewidth]{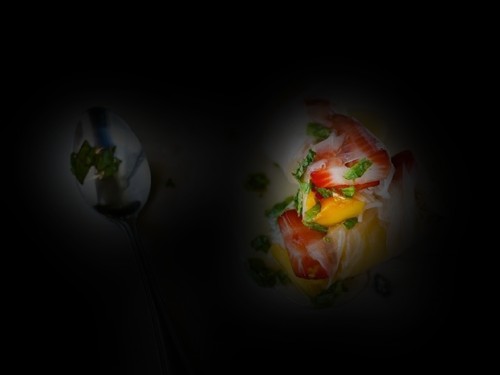} \\
         & & \tiny G-Eymol~\cite{zanca2019gravitational} & \tiny IOR-ROI-LSTM~\cite{chen2018scanpath} & \tiny DeepGazeIII~\cite{kummerer2022deepgaze} \\
         \multirow{3}{*}[1.8em]{\includegraphics[width=0.15\linewidth]{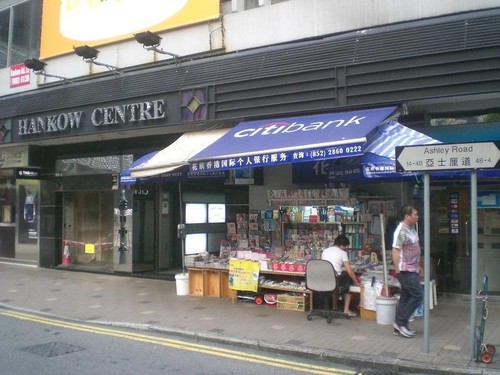}} 
         & & \includegraphics[width=0.14\linewidth]{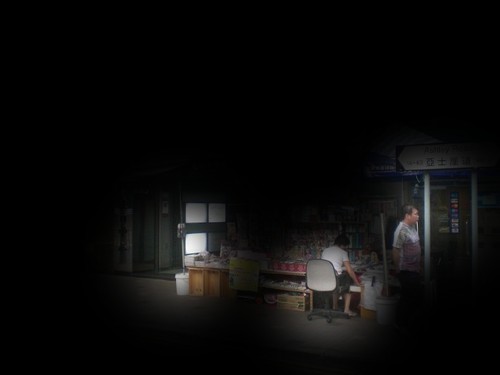}
         & \includegraphics[width=0.14\linewidth]{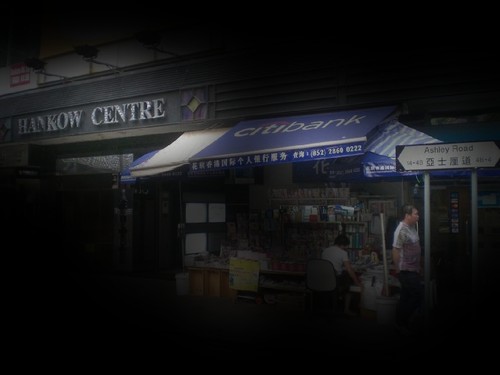}
         & \includegraphics[width=0.14\linewidth]{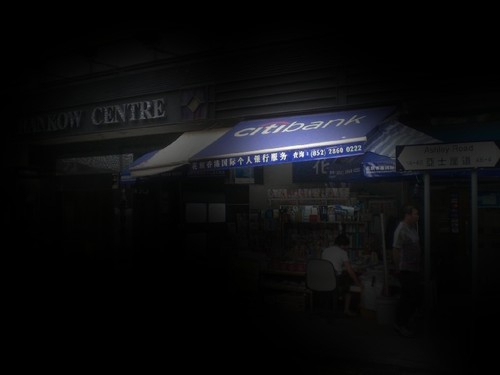} \\ 
         &  & \tiny Scanpath-VQA~\cite{chen2021predicting} & \tiny \textbf{\ours (Ours)}& \tiny Humans \\
         & & \includegraphics[width=0.14\linewidth]{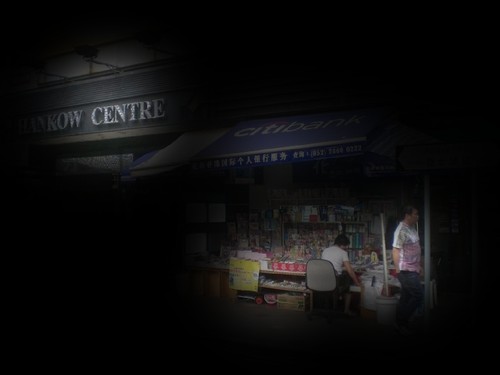}
         & \includegraphics[width=0.14\linewidth]{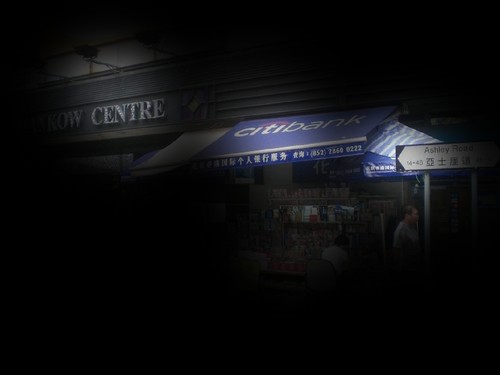}
         & \includegraphics[width=0.14\linewidth]{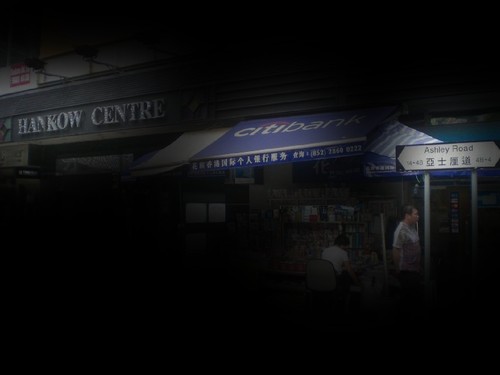} \\
         & & \tiny G-Eymol~\cite{zanca2019gravitational} & \tiny IOR-ROI-LSTM~\cite{chen2018scanpath} & \tiny DeepGazeIII~\cite{kummerer2022deepgaze} \\
         \multirow{3}{*}[1.8em]{\includegraphics[width=0.15\linewidth]{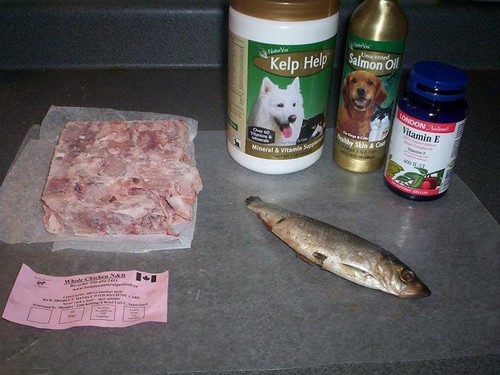}} 
         & & \includegraphics[width=0.14\linewidth]{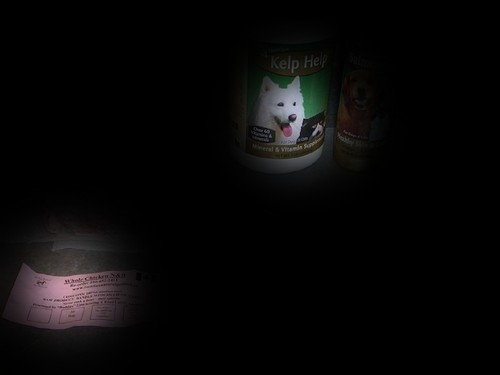}
         & \includegraphics[width=0.14\linewidth]{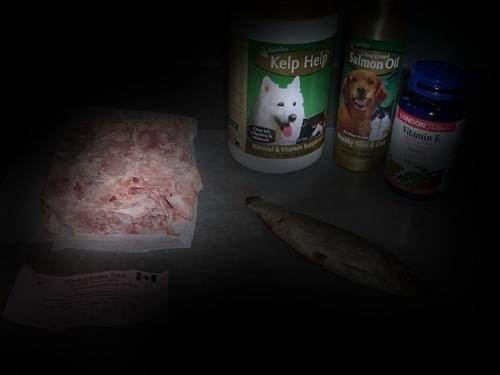}
         & \includegraphics[width=0.14\linewidth]{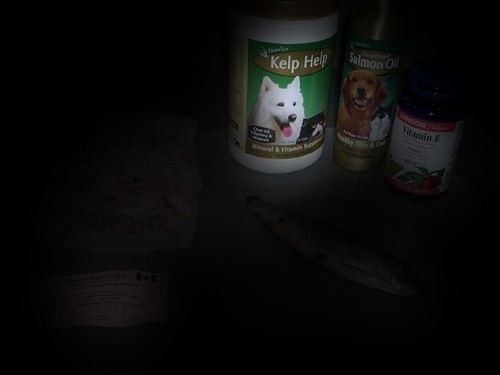} \\ 
         &  & \tiny Scanpath-VQA~\cite{chen2021predicting} & \tiny \textbf{\ours (Ours)}& \tiny Humans \\
         & & \includegraphics[width=0.14\linewidth]{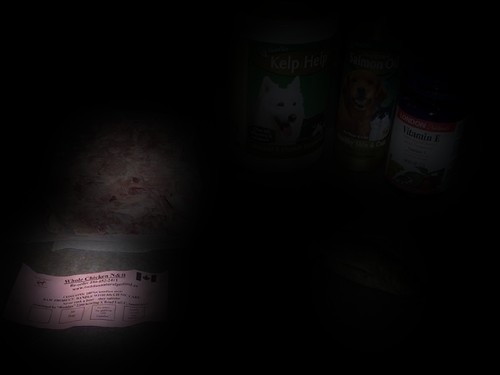}
         & \includegraphics[width=0.14\linewidth]{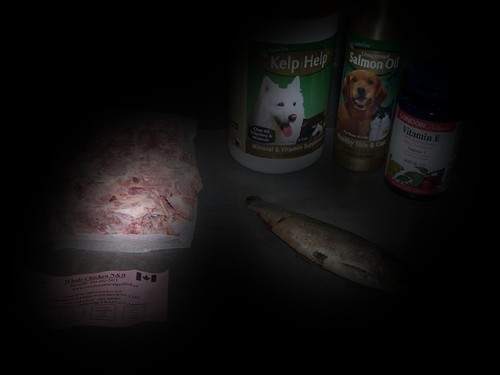}
         & \includegraphics[width=0.14\linewidth]{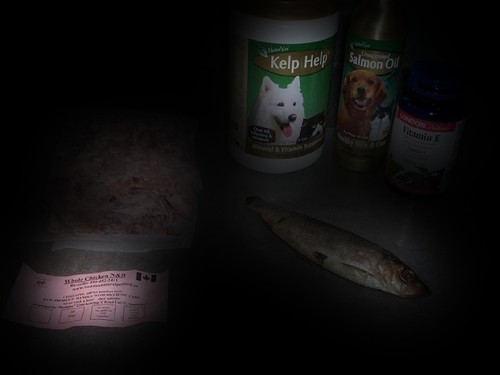} \\
    \end{tabular}
    }
    \vspace{-0.15cm}
    \caption{Saliency maps of sample images from OSIE dataset computed from the fixations generated by the considered scanpath models. For completeness, we include DeepGazeIII, but note that its training procedure also involves saliency prediction.}
    \label{fig:saliency_osie}
    \vspace{-0.4cm}
\end{figure*}

\begin{figure*}[t]
\centering
    \footnotesize
    \setlength{\tabcolsep}{.12em}
    \resizebox{0.95\linewidth}{!}{
    \begin{tabular}{cc ccc}
         & & \tiny G-Eymol~\cite{zanca2019gravitational} & \tiny IOR-ROI-LSTM~\cite{chen2018scanpath} & \tiny DeepGazeIII~\cite{kummerer2022deepgaze} \\
         \multirow{3}{*}[1.8em]{\includegraphics[width=0.15\linewidth]{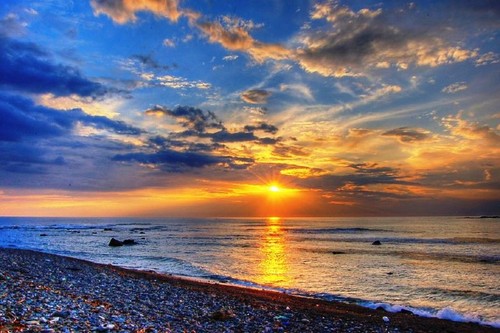}} 
         & & \includegraphics[width=0.14\linewidth]{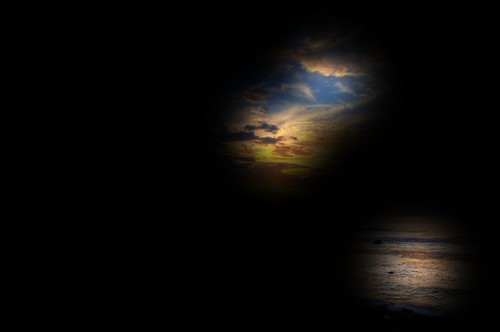}
         & \includegraphics[width=0.14\linewidth]{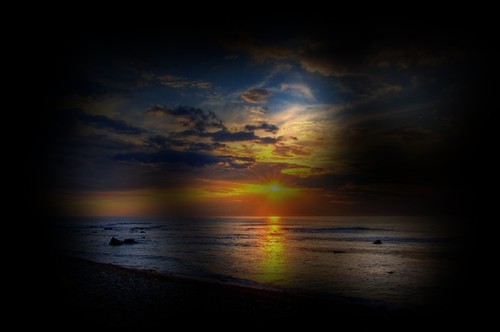}
         & \includegraphics[width=0.14\linewidth]{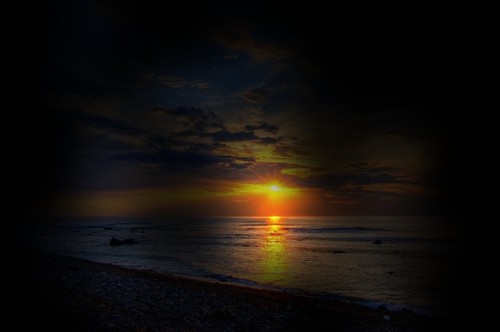} \\ 
         &  & \tiny Scanpath-VQA~\cite{chen2021predicting} & \tiny \textbf{\ours (Ours)}& \tiny Humans \\
         & & \includegraphics[width=0.14\linewidth]{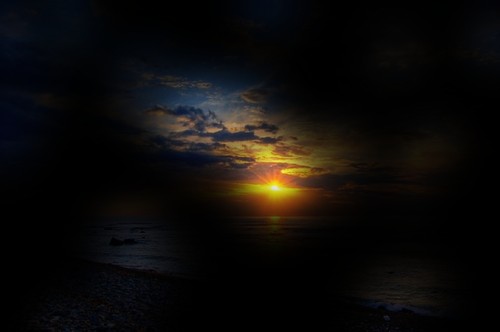}
         & \includegraphics[width=0.14\linewidth]{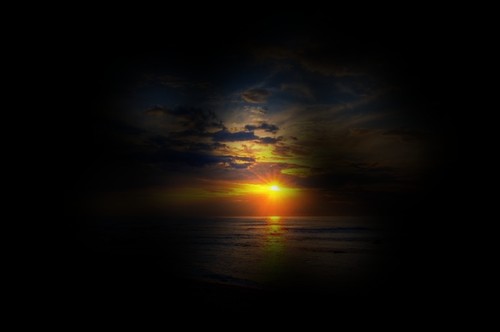}
         & \includegraphics[width=0.14\linewidth]{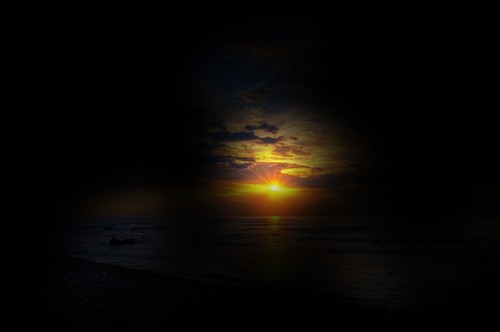} \\
         & & \tiny G-Eymol~\cite{zanca2019gravitational} & \tiny IOR-ROI-LSTM~\cite{chen2018scanpath} & \tiny DeepGazeIII~\cite{kummerer2022deepgaze} \\
         \multirow{3}{*}[1.8em]{\includegraphics[width=0.15\linewidth]{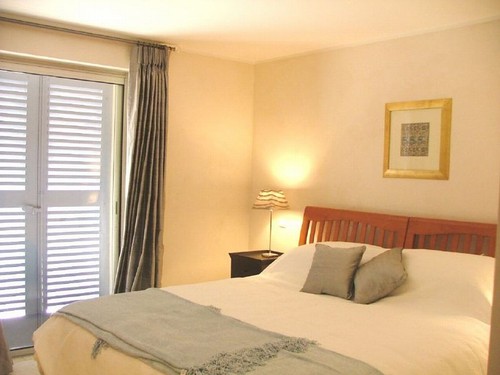}} 
         & & \includegraphics[width=0.14\linewidth]{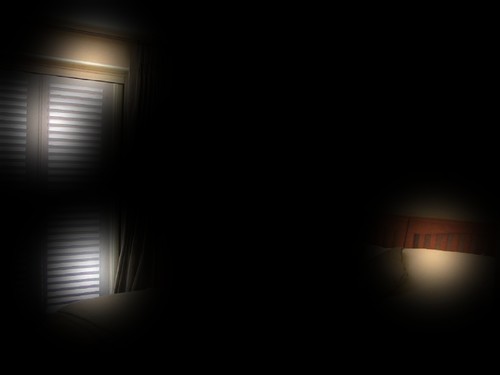}
         & \includegraphics[width=0.14\linewidth]{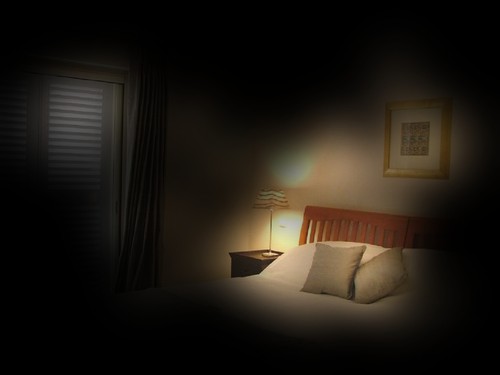}
         & \includegraphics[width=0.14\linewidth]{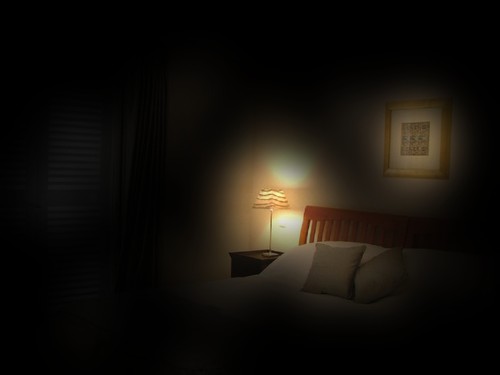} \\ 
         &  & \tiny Scanpath-VQA~\cite{chen2021predicting} & \tiny \textbf{\ours (Ours)}& \tiny Humans \\
         & & \includegraphics[width=0.14\linewidth]{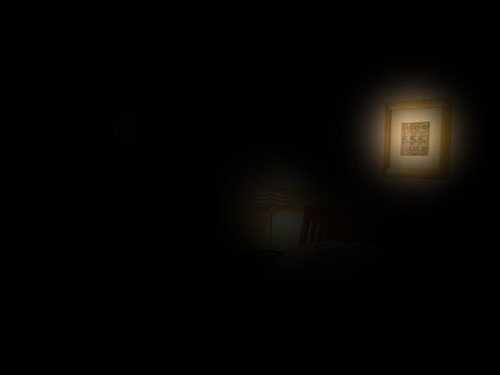}
         & \includegraphics[width=0.14\linewidth]{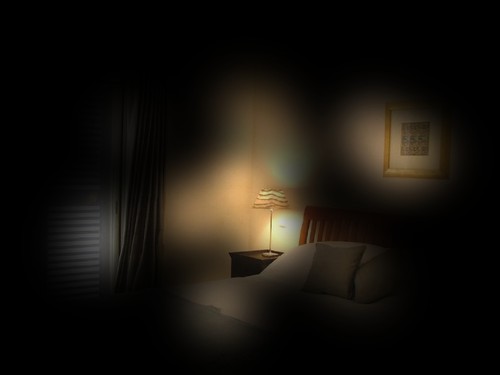}
         & \includegraphics[width=0.14\linewidth]{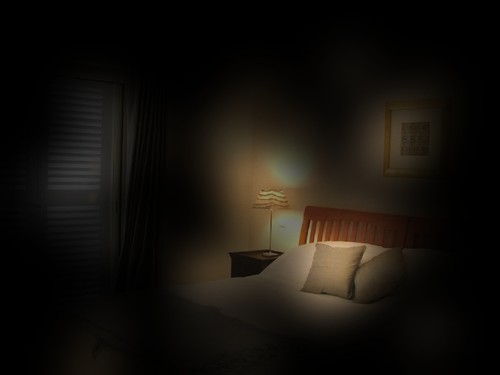} \\
         & & \tiny G-Eymol~\cite{zanca2019gravitational} & \tiny IOR-ROI-LSTM~\cite{chen2018scanpath} & \tiny DeepGazeIII~\cite{kummerer2022deepgaze} \\
         \multirow{3}{*}[1.8em]{\includegraphics[width=0.15\linewidth]{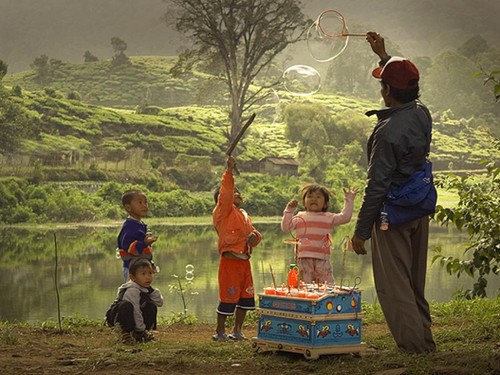}} 
         & & \includegraphics[width=0.14\linewidth]{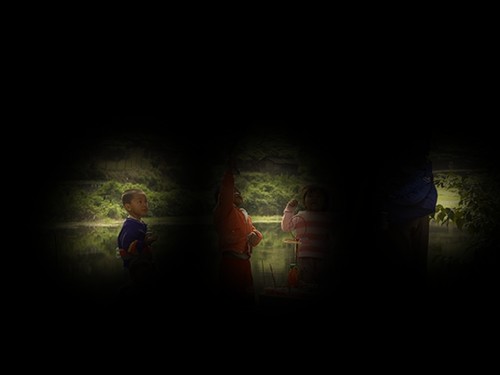}
         & \includegraphics[width=0.14\linewidth]{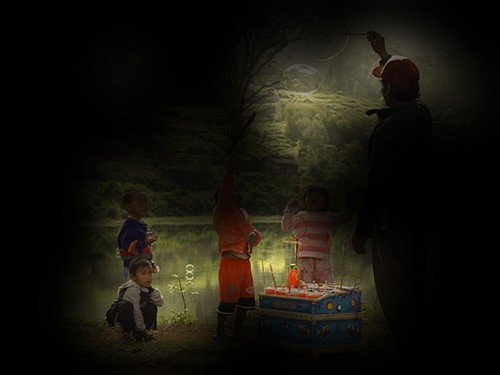}
         & \includegraphics[width=0.14\linewidth]{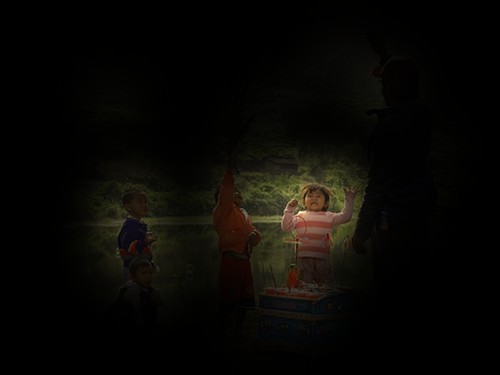} \\ 
         &  & \tiny Scanpath-VQA~\cite{chen2021predicting} & \tiny \textbf{\ours (Ours)}& \tiny Humans \\
         & & \includegraphics[width=0.14\linewidth]{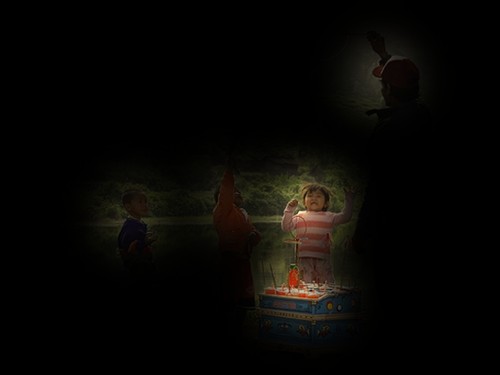}
         & \includegraphics[width=0.14\linewidth]{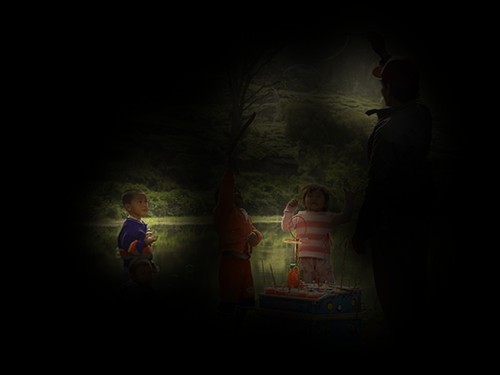}
         & \includegraphics[width=0.14\linewidth]{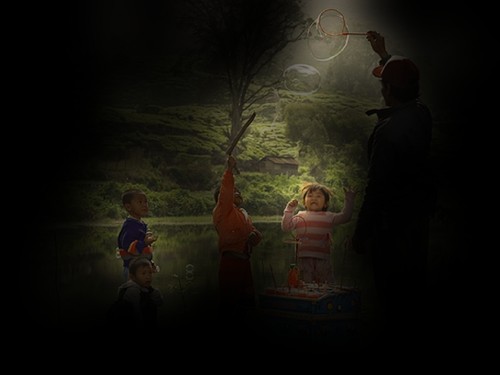} \\
    \end{tabular}
    }
    \vspace{-0.15cm}
    \caption{Saliency maps of sample images from NUSEF dataset computed from the fixations generated by the considered scanpath models. For completeness, we include DeepGazeIII, but note that its training procedure also involves saliency prediction.}
    \label{fig:saliency_nusef}
    \vspace{-0.4cm}
\end{figure*}

\begin{figure*}[t]
\centering
    \footnotesize
    \setlength{\tabcolsep}{.12em}
    \resizebox{0.95\linewidth}{!}{
    \begin{tabular}{cc ccc}
         & & \tiny G-Eymol~\cite{zanca2019gravitational} & \tiny IOR-ROI-LSTM~\cite{chen2018scanpath} & \tiny DeepGazeIII~\cite{kummerer2022deepgaze} \\
         \multirow{3}{*}[1.8em]{\includegraphics[width=0.15\linewidth]{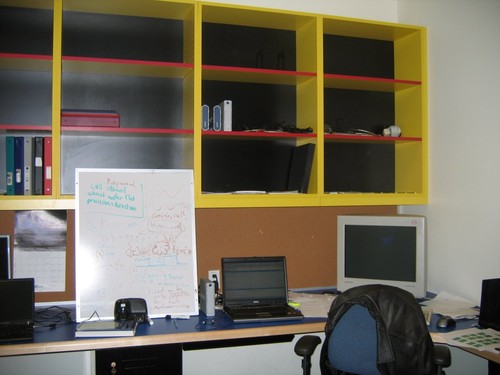}} 
         & & \includegraphics[width=0.14\linewidth]{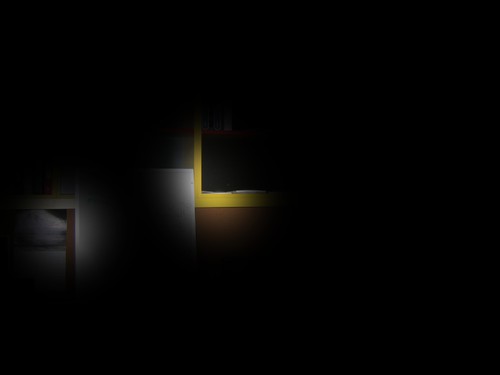}
         & \includegraphics[width=0.14\linewidth]{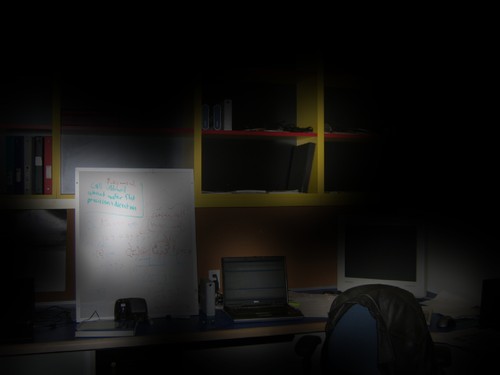}
         & \includegraphics[width=0.14\linewidth]{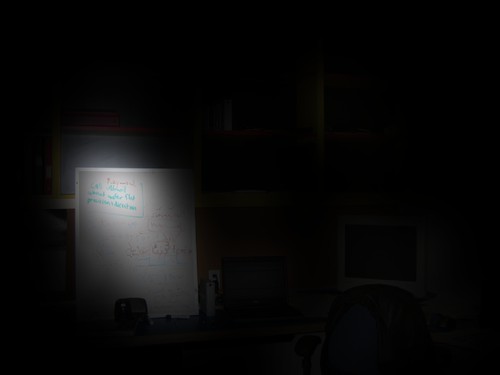} \\ 
         &  & \tiny Scanpath-VQA~\cite{chen2021predicting} & \tiny \textbf{\ours (Ours)}& \tiny Humans \\
         & & \includegraphics[width=0.14\linewidth]{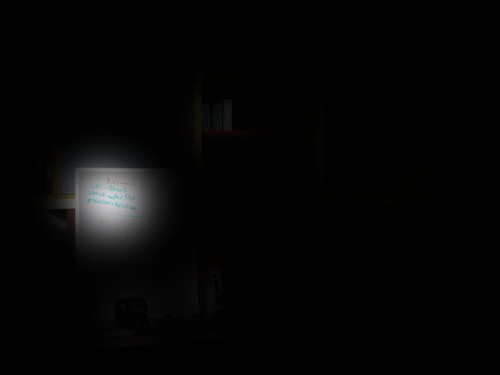}
         & \includegraphics[width=0.14\linewidth]{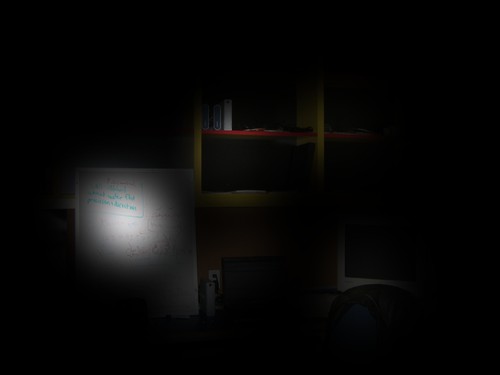}
         & \includegraphics[width=0.14\linewidth]{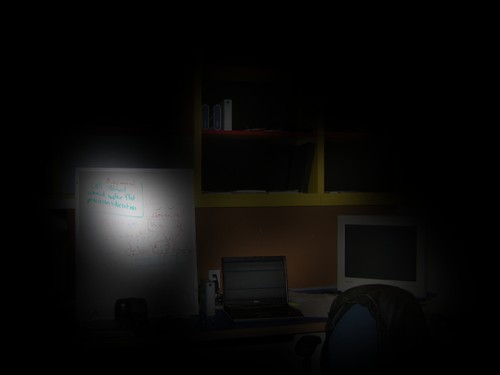} \\
         & & \tiny G-Eymol~\cite{zanca2019gravitational} & \tiny IOR-ROI-LSTM~\cite{chen2018scanpath} & \tiny DeepGazeIII~\cite{kummerer2022deepgaze} \\
         \multirow{3}{*}[1.8em]{\includegraphics[width=0.15\linewidth]{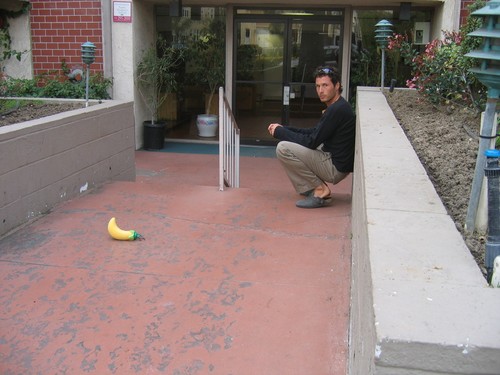}} 
         & & \includegraphics[width=0.14\linewidth]{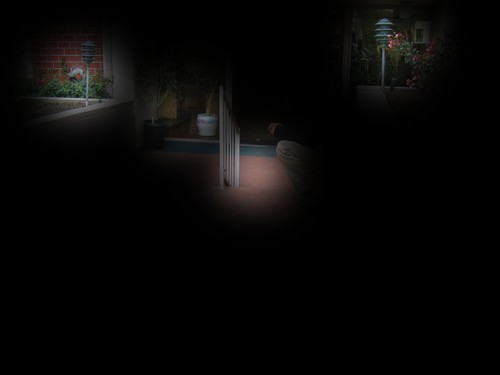}
         & \includegraphics[width=0.14\linewidth]{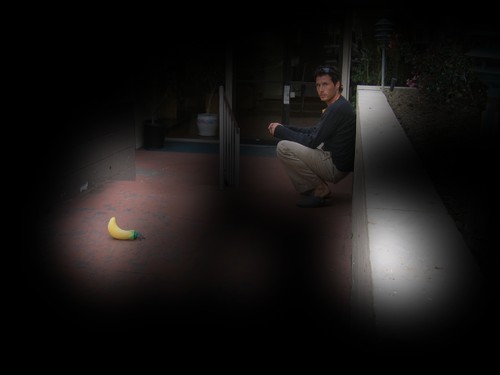}
         & \includegraphics[width=0.14\linewidth]{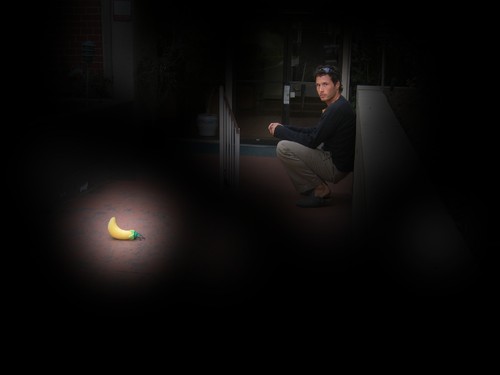} \\ 
         &  & \tiny Scanpath-VQA~\cite{chen2021predicting} & \tiny \textbf{\ours (Ours)}& \tiny Humans \\
         & & \includegraphics[width=0.14\linewidth]{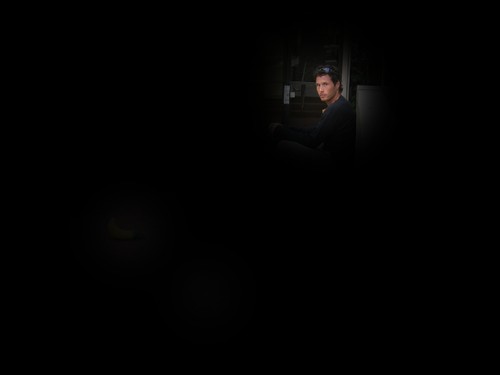}
         & \includegraphics[width=0.14\linewidth]{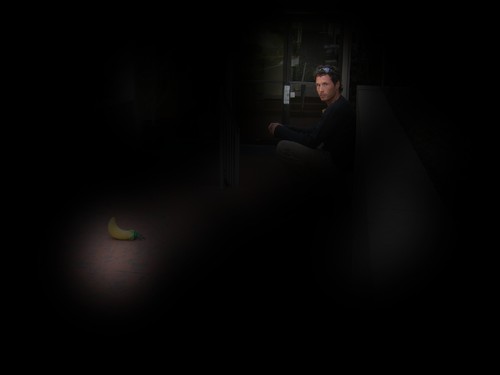}
         & \includegraphics[width=0.14\linewidth]{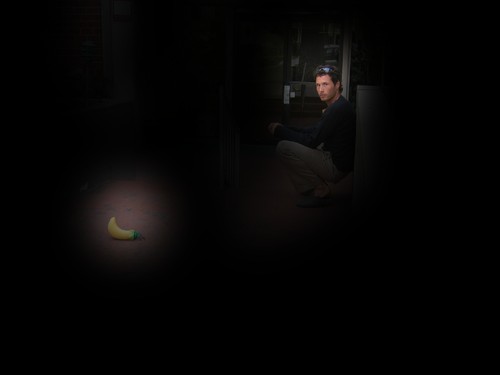} \\
         & & \tiny G-Eymol~\cite{zanca2019gravitational} & \tiny IOR-ROI-LSTM~\cite{chen2018scanpath} & \tiny DeepGazeIII~\cite{kummerer2022deepgaze} \\
         \multirow{3}{*}[1.8em]{\includegraphics[width=0.15\linewidth]{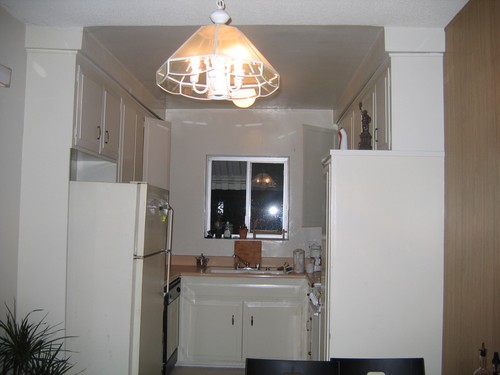}} 
         & & \includegraphics[width=0.14\linewidth]{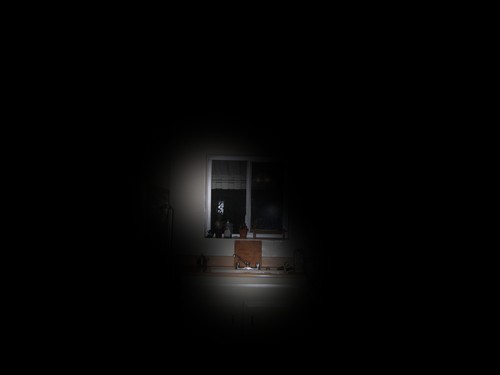}
         & \includegraphics[width=0.14\linewidth]{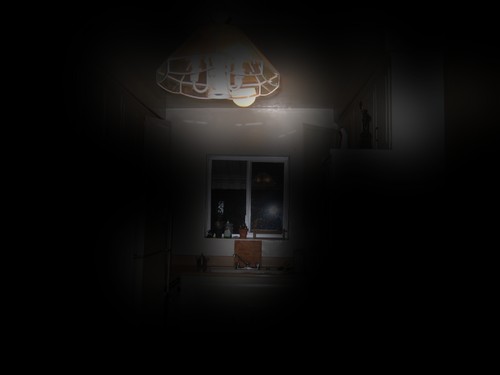}
         & \includegraphics[width=0.14\linewidth]{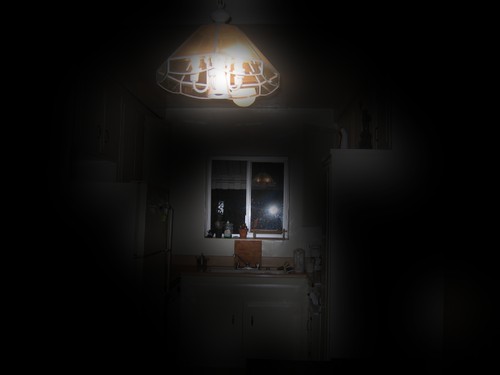} \\ 
         &  & \tiny Scanpath-VQA~\cite{chen2021predicting} & \tiny \textbf{\ours (Ours)}& \tiny Humans \\
         & & \includegraphics[width=0.14\linewidth]{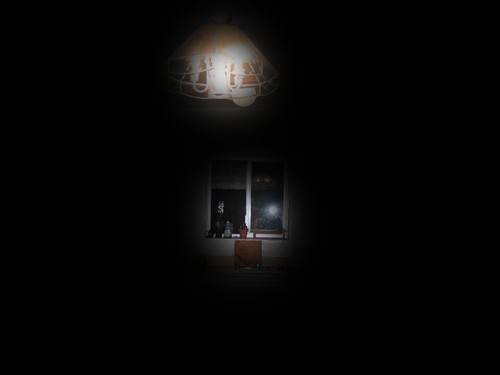}
         & \includegraphics[width=0.14\linewidth]{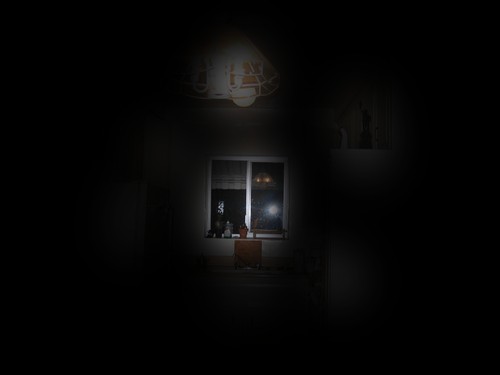}
         & \includegraphics[width=0.14\linewidth]{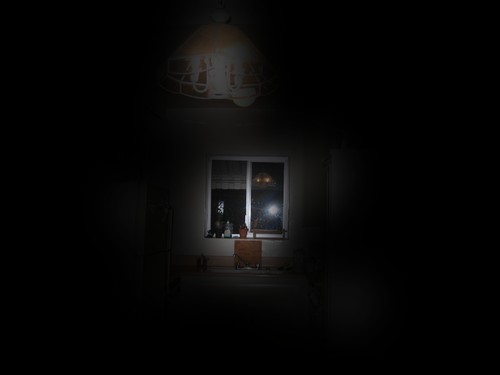} \\
    \end{tabular}
    }
    \vspace{-0.15cm}
    \caption{Saliency maps of sample images from FiFa dataset computed from the fixations generated by the considered scanpath models. For completeness, we include DeepGazeIII, but note that its training procedure also involves saliency prediction.}
    \label{fig:saliency_fifa}
    \vspace{-0.4cm}
\end{figure*}

In the main paper, we also quantitatively assess the performance of the scanpath models on the saliency prediction task. In particular, given a sample image, we construct the aggregated saliency map by convolving a Gaussian kernel over all the locations of predicted fixations~\cite{judd2009learning}.
To support our quantitative analysis, we present the saliency prediction of our model against the competitors from \cref{fig:saliency_cocofreeview} to \cref{fig:saliency_fifa} on COCO-FreeView, MIT1003, OSIE, NUSEF, and FiFa datasets, respectively. Note that we include DeepGazeIII in the comparison for reference even though its results are not directly comparable. Indeed, DeepGazeIII was specifically trained on a large scale dataset~\cite{jiang2015salicon} to predict saliency maps along with scanpaths. Nevertheless, \tppgaze~outperforms DeepGazeIII and the other models in many cases, demonstrating better alignment with humans. 

\begin{figure*}[t]
    \footnotesize
    \setlength{\tabcolsep}{.1em}
    \resizebox{\linewidth}{!}{
    \begin{tabular}{cc ccc}
         & & \tiny Original Image & \tiny \textbf{\ours (Ours)} & \tiny Humans \\
         \rotatebox{90}{\parbox[t]{0.12\linewidth}{\hspace*{\fill}\tiny \textbf{Target:} \texttt{Bowl}\hspace*{\fill}}} 
         & & \includegraphics[height=0.12\linewidth]{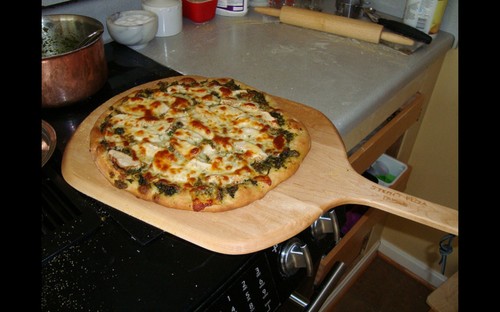} & 
         \includegraphics[height=0.12\linewidth]{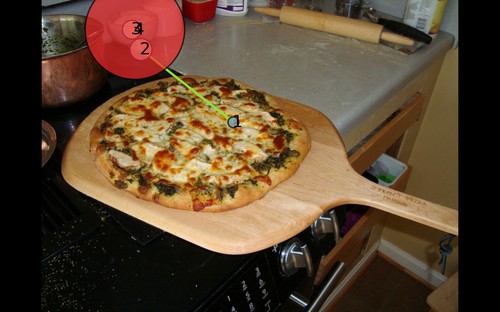} & 
         \includegraphics[height=0.12\linewidth]{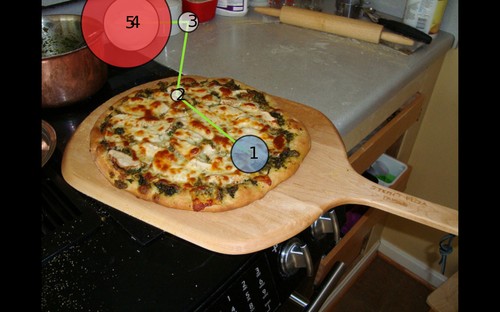} \\
         
         \addlinespace[0.3cm]
         
         & & \tiny Original image & \tiny \textbf{\ours (Ours)} & \tiny Humans \\
         \rotatebox{90}{\parbox[t]{0.12\linewidth}{\hspace*{\fill}\tiny \textbf{Target:} \texttt{Car}\hspace*{\fill}}} 
         & & \includegraphics[height=0.12\linewidth]{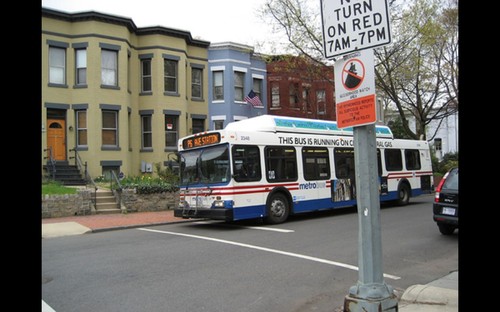} & 
         \includegraphics[height=0.12\linewidth]{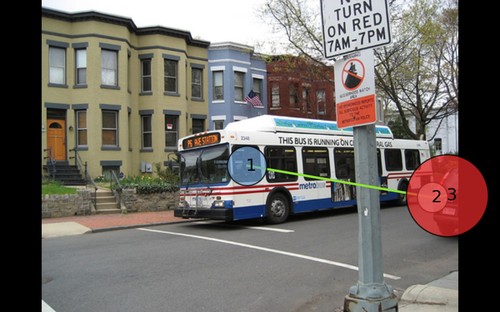} & 
         \includegraphics[height=0.12\linewidth]{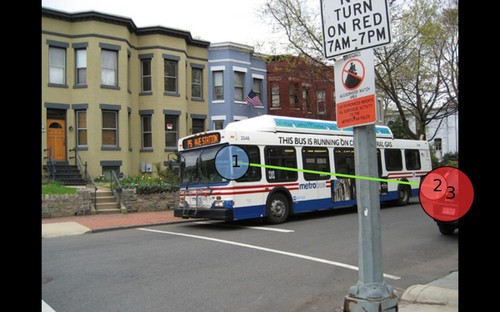} \\

         \addlinespace[0.3cm]
         
         & & \tiny Original Image & \tiny \textbf{\ours (Ours)} & \tiny Humans \\
         \rotatebox{90}{\parbox[t]{0.12\linewidth}{\hspace*{\fill}\tiny \textbf{Target:} \texttt{Chair}\hspace*{\fill}}} 
         & & \includegraphics[height=0.12\linewidth]{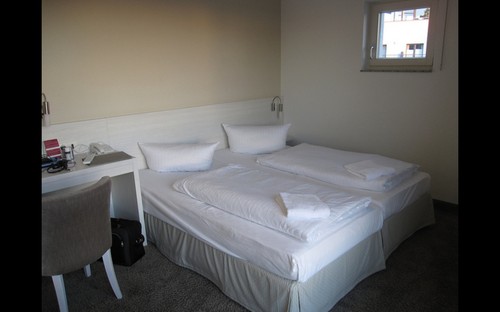} & 
         \includegraphics[height=0.12\linewidth]{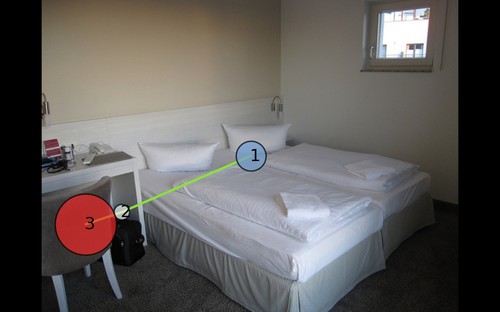} & 
         \includegraphics[height=0.12\linewidth]{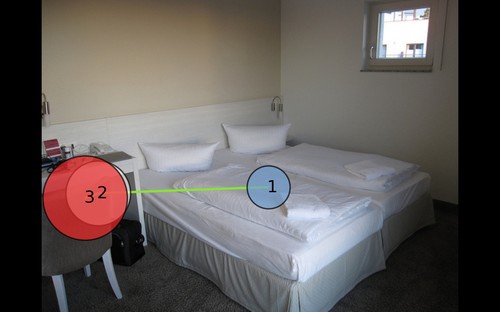} \\

         \addlinespace[0.3cm]
         
         & & \tiny Original Image & \tiny \textbf{\ours (Ours)} & \tiny Humans \\
         \rotatebox{90}{\parbox[t]{0.12\linewidth}{\hspace*{\fill}\tiny \textbf{Target:} \texttt{Knife}\hspace*{\fill}}} 
         & & \includegraphics[height=0.12\linewidth]{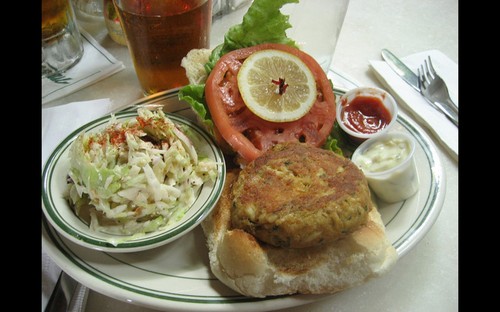} & 
         \includegraphics[height=0.12\linewidth]{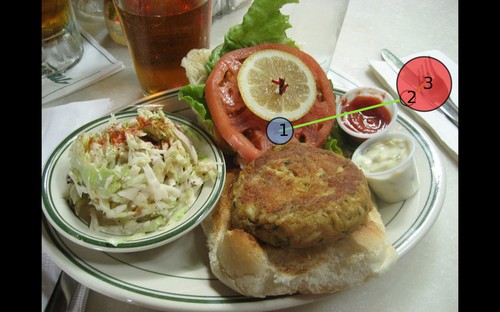} & 
         \includegraphics[height=0.12\linewidth]{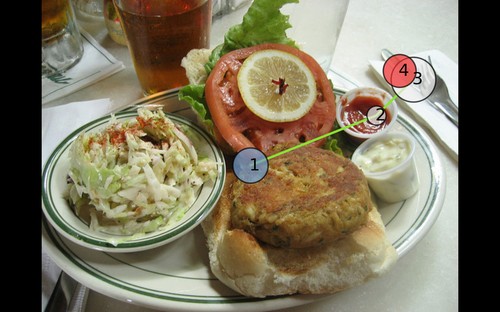} \\
         
    \end{tabular}
    }
    \vspace{-0.15cm}
    \caption{Qualitative comparison of simulated and human scanpaths on the COCO-Search18 dataset for the visual search task.}
    \label{fig:visual_search_supp}
    \vspace{-0.4cm}
\end{figure*}

Finally, in \cref{fig:visual_search_supp} we show additional qualitative results on sample images from COCO-Search18 for the visual search task. As can be observed, \tppgaze~can effectively simulate human-like goal-directed visual attention patterns for various target objects. The model demonstrates its ability to adapt, with a simple architectural variation, from a free-viewing setting to a task-specific visual search scenario. The results illustrate how TPP-Gaze generates plausible attention trajectories that focus on regions likely to contain the target object, mimicking the efficient search strategies employed by humans when looking for specific items in complex scenes.

\end{document}